%% file: paper.tex
\documentclass{new_tlp}
\usepackage[utf8]{inputenc}
\usepackage{amsmath,amssymb}
\usepackage{times,helvet,courier}
\usepackage{url}
\usepackage{stmaryrd}

\usepackage{microtype}
\usepackage{tikz}
\usetikzlibrary{shapes}
\usetikzlibrary{positioning}
\usetikzlibrary{arrows,automata}
\usepackage{pdflscape}
\usepackage{afterpage}
\usepackage{algorithm2e}
\newlength{\commentWidth}
\newcommand{\atcp}[1]{\tcp*[r]{\makebox[\commentWidth]{#1\hfill}}} 
\usepackage{listings}
\lstdefinelanguage{clingo}{
  keywordstyle=[1]\usefont{OT1}{cmtt}{m}{n},%
  keywordstyle=[2]\textbf,%
  keywordstyle=[3]\usefont{OT1}{cmtt}{m}{n},
  alsoletter={\#,\&},%
  keywords=[1]{not,from,import,def,if,else,return,while,break,and,or,for,in,del,and,class},%
  keywords=[2]{\#const,\#show,\#minimize,\#base,\#theory,\#count,\#external,\#program,\#script,\#end,\#heuristic,\#edge,\#project,\#show},%
  keywords=[3]{&,&dom,&sum,&diff,&show,&minimize},%
  morecomment=[l]{\#\ },%
  morecomment=[l]{\%\ },%
  commentstyle={\color{darkgray}}%
}

\usepackage[format=plain, textfont=it]{caption}

\input{macro}

\title{\textit{Clingcon}: The Next Generation}
\author[Mutsunori Banbara, Benjamin Kaufmann, Max Ostrowski and Torsten Schaub]{%
  Mutsunori Banbara\\
  Kobe University, Japan
  \and
  Benjamin Kaufmann and Max Ostrowski\\
  University of Potsdam, Germany
  \and
  Torsten Schaub\thanks{Affiliated with the
    Simon Fraser University,
    Canada,
    and
    Griffith University,
    Australia.} \\
  University of Potsdam, Germany and
  INRIA Rennes, France}

\begin{document}

\maketitle

\input{abstract}

\input{introduction}

\input{background}
\input{approach}
\input{architecture}
\input{language}

\input{algorithms}
\input{features}

\input{multishot}

\input{experiments}

\input{discussion}

\bibliographystyle{acmtrans}
\bibliography{lit,akku,own,procs} 

\end{document}

%% file: macro.tex
\newtheorem{theorem}{Theorem}[section]

\newtheorem{corollary}{Corollary}[section]

\newcommand{\Tsign}{\ensuremath{\mathbf{T}}}
\newcommand{\Fsign}{\ensuremath{\mathbf{F}}} 

\newcommand{\naf}[1]{\ensuremath{{\sim\!{#1}}}}

\newcommand{\head}[1]{\ensuremath{\mathit{head}(#1)}} 
\newcommand{\atom}[1]{\ensuremath{\mathit{atom}(#1)}}
\newcommand{\body}[1]{\ensuremath{\mathit{body}(#1)}} 

\newcommand{\Bss}{\ensuremath{\mathbf{B}}}
\newcommand{\Css}{\ensuremath{\mathbf{C}}}
\newcommand{\Tass}[1]{\ensuremath{#1^{\Tsign}}}
\newcommand{\Fass}[1]{\ensuremath{#1^{\Fsign}}}

\newcommand{\dom}[1]{\ensuremath{\mathit{D}(#1)}}
\newcommand{\img}[1]{\ensuremath{\mathit{img}(#1)}}

\newcommand{\atbody}[2]{\ensuremath{\mathit{body}_{\!#1}(#2)}}
\newcommand{\EB}[2]{\ensuremath{\mathit{EB}_{\!#2}(#1)}}

\newcommand{\Tlit}[1]{\ensuremath{\Tsign #1}}
\newcommand{\Flit}[1]{\ensuremath{\Fsign #1}}

\newcommand{\tolit}[1]{\ensuremath{(#1)^{\ddagger}}} 
\newcommand{\tolitorder}[1]{\ensuremath{(#1)^{\dagger}}} 

\newcommand{\conf}[1]{\ensuremath{\mathit{#1}}}
\newcommand{\confb}[1]{\ensuremath{\boldsymbol{\mathit{#1}}}}

\newcommand{\sysfont}{\textit}
\newcommand{\adsolver}{\sysfont{adsolver}}
\newcommand{\aspartame}{\sysfont{aspartame}}
\newcommand{\aspif}{\sysfont{aspif}}
\newcommand{\chuffed}{\sysfont{chuffed}}
\newcommand{\clasp}{\sysfont{clasp}}
\newcommand{\clingcon}{\sysfont{clingcon}}
\newcommand{\clingo}{\sysfont{clingo}}

\newcommand{\dingo}{\sysfont{dingo}}
\newcommand{\dlvhex}{\sysfont{dlvhex}}
\newcommand{\ezcsp}{\sysfont{ezcsp}}
\newcommand{\ezsmt}{\sysfont{ezsmt}}
\newcommand{\flatzinc}{\sysfont{flatzinc}}
\newcommand{\fztoaspif}{\sysfont{fz2aspif}}
\newcommand{\minizinc}{\sysfont{minizinc}}
\newcommand{\gfd}{\sysfont{g12fd}}
\newcommand{\gecode}{\sysfont{gecode}}
\newcommand{\gringo}{\sysfont{gringo}}

\newcommand{\idp}{\sysfont{idp}}
\newcommand{\inca}{\sysfont{inca}}

\newcommand{\minisatid}{\sysfont{minisatid}}
\newcommand{\mznfzn}{\sysfont{mzn2fzn}}

\newcommand{\picatsat}{\sysfont{picatsat}}
\newcommand{\python}{Python}
\newcommand{\sugar}{\sysfont{sugar}}
\newcommand{\zthree}{\sysfont{z3}}

\newcommand{\commandfont}{\texttt}

\RestyleAlgo{algoruled}
\DontPrintSemicolon
\LinesNumbered
\SetAlgoVlined
\SetFuncSty{textsc}

\SetKwInOut{Global}{Global}
\SetKwInOut{Input}{Input}
\SetKwInOut{Output}{Output}
\SetKw{Backtrack}{return}
\SetKw{Print}{print}
\SetKw{Return}{return}
\SetKw{Exit}{exit}
\SetKw{Or}{or}
\SetKw{Continue}{continue}
\SetKwFunction{Select}{Select}
\SetKwFunction{Conflict}{Conflict}
\SetKwFunction{RootConflict}{RootConflict}
\SetKwFunction{Backtrack}{Backtrack} \newcommand{\BacktrackArg}[2]{\ensuremath{\textsc{Backtrack}_{#1}(#2)}}
\SetKwFunction{Complete}{Complete}
\SetKwFunction{Propagation}{Propagation} \newcommand{\PropagationArg}[2]{\ensuremath{\textsc{Propagation}(#2)}}
\SetKwFunction{Translate}{Translate} \newcommand{\TranslateArg}[2]{\ensuremath{\textsc{Translate}(#2)}}
\SetKwFunction{Print}{Print} 
\SetKwFunction{Split}{Split} \newcommand{\SplitArg}[2]{\ensuremath{\textsc{Split}_{#1}(#2)}}
\SetKwFunction{Select}{Select}
\SetKwFunction{UnitPropagation}{UnitPropagation}  \newcommand{\UnitPropagationArg}[2]{\ensuremath{\textsc{UnitPropagation}_{#1}(#2)}}
\SetKwFunction{UfsPropagation}{UfsPropagation}  \newcommand{\UfsPropagationArg}[2]{\ensuremath{\textsc{UfsPropagation}_{#1}(#2)}}
\SetKwFunction{CspPropagation}{CspPropagation}  \newcommand{\CspPropagationArg}[2]{\ensuremath{\textsc{CspPropagation}(#2)}}
\SetKwFunction{ConflictAnalysis}{Con\-flict\-Ana\-ly\-sis}
\SetKwFunction{NonTight}{NonTight}
\SetKwFunction{UnFoundedSet}{UnfoundedSet}
\SetKwFunction{LoopFormula}{LoopNogood}
\SetKwFunction{CDNL}{CDNL-ASP}
\SetKwFunction{Enum}{CDNL-ENUM-ASP} 
\SetKwFunction{Translate}{Translate}
\SetKwFunction{Create}{Create}
\SetKwFor{Let}{let}{in}{tel}
\SetKw{SuchThat}{such that}
\SetKw{ForSome}{for some}
\SetKwFor{Such}{let}{\SuchThat}{tel}
\SetKwFor{IN}{}{in}{tel}
\SetKwFor{Loop}{loop}{}{}
\SetKwFor{While}{while}{}{}
\SetCommentSty{textit}

%% file: abstract.tex
\begin{abstract}
We present the third generation of the constraint answer set system \clingcon,
combining Answer Set Programming (ASP) with finite domain constraint processing (CP).
While its predecessors rely on a black-box approach to hybrid solving by integrating the CP solver \gecode,
the new \clingcon\ system pursues a lazy approach using dedicated constraint propagators 
to extend propagation in the underlying ASP solver \clasp.
No extension is needed for parsing and grounding \clingcon's hybrid modeling language 
since both can be accommodated by the new generic theory handling capabilities of the ASP grounder \gringo.
As a whole,
\clingcon~3 is thus an extension of the ASP system \clingo~5, which itself relies on the grounder \gringo\ and the solver \clasp.
The new approach of \clingcon\ offers a seamless integration of CP propagation into ASP solving that benefits from the whole spectrum of \clasp's reasoning modes,
including for instance multi-shot solving and advanced optimization techniques.
This is accomplished by a lazy approach that unfolds the representation of constraints and 
adds it to that of the logic program only when needed.
Although the unfolding is usually dictated by the constraint propagators during solving,
it can already be partially (or even totally) done during preprocessing.
Moreover, \clingcon's constraint preprocessing and propagation incorporate several well established CP techniques 
that greatly improve its performance.
We demonstrate this via an extensive empirical evaluation 
contrasting, first, the various techniques in the context of CSP solving
and, second, the new \clingcon\ system with other hybrid ASP systems.
\end{abstract}
%

%% file: introduction.tex
\section{Introduction}\label{sec:introduction}

The shortcoming of Answer Set Programming (ASP;~\cite{lifschitz08b}) to succinctly represent variables over large numeric domains
has led to the development of several systems enhancing ASP with capabilities for finite domain Constraint Processing (CP;~\cite{CPHandbook}).
Starting from the seminal work in~\cite{baboge05a} and the consecutive development of traditional DPLL%
\footnote{Tracing back to the Davis-Putman-Logemann-Loveland procedure~\cite{davput60a,dalolo62a}}-style hybrid ASP solvers
like \adsolver~\cite{megezh08a},
modern hybrid ASP solvers take advantage of CDCL%
\footnote{Standing for: Conflict-Driven Constraint Learning}-based solving technology~\cite{marsak99a,zamamoma01a,gekanesc07a} in different ways.
Let us illustrate this by describing the approach of three representative
Constraint Answer Set Programming (CASP;~\cite{ballie13a}) systems.

A black-box approach is pursued in the two previous \clingcon{} series where
the ASP solver \clasp{} is combined with the CP solver \gecode~\cite{gecode}
by following the lazy approach to SMT%
\footnote{Standing for: Satisfiability Modulo Theories}%
solving~\cite{baseseti09a}.
In the \clingcon{} setting,
this means that \clasp{} only generates truth assignments for abstracted constraint expressions,
while \gecode{} checks whether the actual constraints can be made true or false accordingly.
On the one hand,
this black-box approach benefits from the vast spectrum of constraints available in \gecode\
and seamlessly keeps up with advanced CP technology,
among others regarding preprocessing and propagation.
Moreover, this approach avoids an explicit representation of integer variables in ASP
and thus can deal with very large domains.
On the other hand,
the usage of an external CP solver restricts information exchange which impedes the CDCL approach of \clasp.
First, neither conflict nor propagation information is provided by \gecode{} and thus must be approximated within the interface 
to sustain conflict analysis in CDCL\@.
Second, the granularity induced by constraint abstraction leads to weaker propagation than what is obtainable when encoding integer variables.

A translation-based approach is pursued by the \aspartame{} system~\cite{bageinospescsotawe15a} where
a CSP\footnote{Standing for: Constraint Satisfaction Problem} is fully translated into ASP and then solved by an ASP solver.
This approach follows the one of the CP solver \sugar~\cite{tatakiba09a} 
translating CSPs to SAT\footnote{Standing for: Satisfiability Testing}~\cite{SATHandbook}.
This is done by representing each integer variable along with its domain according to the order encoding scheme~\cite{crabak94a}.
Such an approach is called eager in SMT solving.
On the one hand,
this approach benefits from the full power of CDCL-based search.
Also, the granularity induced by an explicit representation of integer variables provides more accurate conflict and propagation information,
and approximations for reasons and conflicts as used in the former \clingcon{} system~\cite{ostsch12a} are made obsolete.
On the other hand,
such an explicit representation limits scalability:
\aspartame{} (just as \sugar) can only deal with medium sized domains up to a few thousand integers.
Also, when dealing with larger domains, CDCL search may suffer from congestion due to too much conflict information.
Finally,
\aspartame{} cannot make use of readily available CP techniques for preprocessing and propagation;
all this must be captured in the underlying ASP encoding.

A lazy approach is pursued by the \inca{} system~\cite{drewal12a} where
the ASP solver \clasp{} is augmented with dedicated propagators for linear and selected global constraints
by following the approach of lazy clause generation~\cite{ohstco09a}.
The idea is to make parts of the encoding explicit whenever they reflect a conflict or propagation signaled by a propagator.
In this way, the explicit representation of constraints is only unfolded when needed and its extent is controlled by the deletion scheme of the ASP solver.
This approach also benefits from the full power of CDCL-based search but outsources constraint oriented inferences.
In this way, the overall size of the hybrid problem is under control of the ASP solver.
As a consequence, \inca{} can deal with large domains.
But it has its limits because the vocabulary and basic inference schemes of the order encoding must be provided at the outset
by introducing auxiliary variables and nogoods.
The propagators rely on this for making parts of the constraint encoding explicit.
Moreover,
this lazy approach cannot harness implemented CP techniques for preprocessing and propagation;
\inca{} provides advanced means for propagation but uses no sophisticated preprocessing techniques.

The third generation of \clingcon{} also follows a lazy approach to hybrid ASP solving
but largely extends the lazy one of \inca{} 
while drawing on experience with \aspartame{} and the previous \clingcon{} series.
The current version of \clingcon~3 features propagators for linear constraints and can translate distinct constraints.
The ultimate design goal was to conceive a hybrid solver architecture that integrates seamlessly with the infrastructure of the ASP system \clingo{} in
order to take advantage of its full spectrum of grounding and solving capabilities.
For the latter, it is essential to give the solver access to the representation of constraint variables and their domains,
otherwise hybrid forms of multi-objective optimization or operations on models like intersection or union cannot reuse existing capacities.
The lazy approach lets us accomplish this while controlling space demands.
However, we take the approach of \inca{} one step further by permitting lazy variable generation~\cite{thistu09a}
to unfold the vocabulary and the basic inference schemes of the order encoding only when needed.
This enables \clingcon~3 to represent very large (and possibly non-contiguous) domains of integer variables.
Furthermore,
\clingcon~3 features a variety of established CP preprocessing techniques to enhance its lazy approach.
This also includes an initial eager translation that allows for unfolding up front parts or even the entire CSP.

What is more, \clingcon{} is not restricted to single-shot solving but fully blends in with \clingo's multi-shot solving capabilities~\cite{gekaobsc15a}.
This does not only allow for incremental hybrid solving but moreover equips \clingcon{} with powerful APIs. 
For instance, the latter allow for conceiving reactive procedures to loop on solving while acquiring changes in the problem specification.
In fact, due to our design, most of \clingo's elaborate features carry over to \clingcon.
Among others, this includes multi-threaded solving
as well as
unsatisfiable core and model-driven multi-criteria optimization.
Exceptions to this are signature-based forms of reasoning, like projective enumeration or heuristic modifications
that must be dealt with indirectly by associating constraint atoms with auxiliary regular atoms with which such operations can be performed.

Our paper is structured as follows.
The next section provides the formal foundations of Constraint Answer Set Programming (CASP) and 
presents the basics of CDCL-based ASP solving along with their extension to CASP solving.
Section~\ref{sec:approach} details relevant features of \clingcon~3.
We start with an architectural overview in Section~\ref{sec:architecture} and introduce the input language of \clingcon~3 in Section~\ref{sec:language}.
We then explain \clingcon's extended solving algorithms in Section~\ref{sec:algorithms} and detail distinguished features in Section~\ref{sec:features}.
The final subsection of Section~\ref{sec:approach} is dedicated to multi-shot CASP solving.
Section~\ref{sec:experiments} provides a detailed empirical analysis of \clingcon's features and performance in contrast to competing CP and CASP systems.
We summarize the salient features of the new \clingcon{} series in Section~\ref{sec:discussion} and discuss related work.


%% file: background.tex
\section{Formal Preliminaries}\label{sec:background}

We begin in Section~\ref{sec:casp} with a gentle introduction to CASP along with some auxiliary concepts.
We then provide the basics of CDCL-based ASP solving and show how they extend to CASP solving in Section~\ref{sec:bcs}.

\subsection{Constraint Answer Set Programming}\label{sec:casp}

Constraint logic programs consist of 
a logic program $P$ 
over disjoint sets $\mathcal{A},\mathcal{C}$ of propositional variables, and 
an associated constraint satisfaction problem (CSP) $(\mathcal{V},D,C)$.
Elements of $\mathcal{A}$ and $\mathcal{C}$ are referred to as regular and constraint atoms, respectively.
We consider linear CSPs,
where 
$\mathcal{V}$ is a set of integer variables, 
$D$ is a set of corresponding variable domains, and 
$C$ is a set of linear constraints.
 
\paragraph{Logic programs.}

A logic program $P$ consists of rules of the form%
\footnote{We present our approach in the context of normal logic programs, though it readily applies to disjunctive logic programs --- as does \clingcon~3.}
\begin{equation}\label{eqn:rule}
a_0\leftarrow a_1,\dots,a_m,\naf{a_{m+1}},\dots,\naf{a_n}
\end{equation}
where $0\leq m\leq n$ and $a_0\in\mathcal{A}$ and each $a_i\in{\mathcal{A}\cup\mathcal{C}}$ is an atom for $1\leq i\leq n$.

As an example, consider the logic program $P_1$:
\begin{align}
\label{rule:one}a &\leftarrow \naf{b}\\
\label{rule:two}b &\leftarrow \naf{a}\\
\label{rule:tri}c &\leftarrow a, x < 7
\end{align}
This program contains regular atoms  $a$, $b$, and $c$ from $\mathcal{A}$
along with the constraint atom $x < 7$ from $\mathcal{C}$.
Accordingly, $x$ is an integer variable in $\mathcal{V}$.

We need the following auxiliary definitions.
We define $\head{r}=a_0$ as the head of rule $r$ in \eqref{eqn:rule},
\(
\body{r}=\{a_1,\dots,a_m,\naf{a_{m+1}},\dots,\naf{a_n}\}
\)
as its body,
and
\(
\atom{r}=\{a_0,a_1,\dots,a_m,a_{m+1},\dots,a_n\}
\).
Moreover,
we let
\(
\head{P}=\{\head{r}\mid r\in P\}
\),
\(
\body{P}=\{\body{r}\mid r\in P\}
\),
\(
\atbody{P}{a}=\{\body{r}\mid r\in P,\head{r}=a\}
\),
and
\(
\atom{P}=\{\atom{r}\mid r\in P\}
\).
If $\body{r}=\emptyset$, $r$ is called a fact.
If $\head{r}$ is missing, $r$ is called an integrity constraint and $r$ stands for $x\leftarrow\body{r},\naf{x}$ where $x$ is a new atom.%
\footnote{As syntactic sugar, a rule $c \leftarrow a_1,\dots,a_m,\naf{a_{m+1}},\dots,\naf{a_n}$ with a constraint atom $c\in\mathcal{C}$ in the head
stands for $\leftarrow a_1,\dots,a_m,\naf{a_{m+1}},\dots,\naf{a_n}, \naf{c}$.}

In ASP, the semantics of a logic program is given by its \emph{(constraint) stable models}~\cite{gellif88b,geossc09a}.
However,
in view of our focus on computational aspects,
we rather deal with Boolean assignments and constraints and give a corresponding characterization of a program's stable models below.

\paragraph{Constraint Satisfaction Problems.}

A linear CSP $(\mathcal{V},D,C)$ deals with linear constraints in $C$ of the form
\begin{equation}\label{eq:linear:constraint}
  {a_1v_1}+\dots+{a_nv_n} \leq b
\end{equation}
where $a_i$ and $b$ are integers and $v_i\in\mathcal{V}$ for $1\leq i\leq n$.
The domain of a variable $v\in\mathcal{V}$ is given by $\dom{v}$.
The complement of a constraint $c\in C$ is denoted as $\overline{c}$.
We require that $C$ is closed under complements.
Constraint atoms in $\mathcal{C}$ are identified with constraints in $C$ via
a function
\(
\gamma: \mathcal{C}\to C
\).

In our example,
we have $x\in\mathcal{V}$ and 
let $\dom{x}=\{1,\dots,10\}$.
Moreover,
we associate the constraint atom $x < 7$ with the linear constraint $x \leq 6$,
or formally,
\(
\gamma(x < 7)=x \leq 6
\).
Since we require $C$ to be closed under complements,
it contains both $x \leq 6$ and its complement $-x \leq -7$.

An assignment
\(
\Css: v\in\mathcal{V}\mapsto d\in\dom{v}
\)
\emph{satisfies} a linear constraint,
if \eqref{eq:linear:constraint} holds after replacing each $v_i$ by $\Css(v_i)$.
We let $\mathit{sat}_{\Css}(C)$ denote the set of all constraints in $C$ satisfied by \Css.
Following~\cite{drescher15a},
we call $(\Css,\mathit{sat}_{\Css}(C))$ a \emph{configuration} of $(\mathcal{V},D,C)$.
For instance,
the assignment $\Css=\{x\mapsto 5\}$ satisfies the linear constraint $x \leq 6$.
Accordingly,
\(
(\{x\mapsto 5\},\{x \leq 6\})
\)
is a configuration of
\(
(\{x\},\{D(x)\},\{x \leq 6,-x \leq -7\}\})
\).

Moreover, we rely on the CP concept of a \emph{view}.
Following~\cite{schtac06a},
a \emph{view} on a variable $x$ is an expression $ax+b$ for integers $a,b$;
its \emph{image} is defined as $\img{ax+b}=\{ax+b \mid x\in \dom{x}\}$.%
\footnote{Any linear expression with only one variable can be converted to an expression of the form $ax+b$.}
Since a view $ax+b$ can always be replaced with a fresh variable $y$ along with a constraint $y=ax+b$,
we may use them nearly everywhere where we would otherwise use variables.
%
For a view $v$,
we define $\mathit{lb}(v)$ and $\mathit{ub}(v)$ as the smallest/largest value in $\img{v}$.%
\footnote{Note that for a view of the form $1x+0$ we have $\dom{x}=\img{x}$.}
Then,
$\mathit{prev}(d,v)$ ($\mathit{next}(d,v)$) is a function mapping a value $d$
to the largest (smallest) element $d'$ in $\img{v}$ which is smaller (larger) than $d$
if $d > \mathit{lb}(v)$ ($d < \mathit{ub}(v)$), otherwise it is $-\infty$ ($\infty$).
In our example, we have $\mathit{lb}(x)=1$ and $\mathit{ub}(x)=10$,
and for instance $\mathit{prev}(17,2x+3)=15$, $\mathit{prev}(5,x)=4$, and $\mathit{prev}(0,x)=-\infty$, respectively.

\subsection{Basics of ASP and CASP Solving}\label{sec:bcs}

The basic idea of CDCL-based ASP solving is to map inferences from rules as in \eqref{eqn:rule} to unit propagation on Boolean constraints.
Our description of this approach follows the one given in~\cite{gekakasc12a}.

Accordingly,
we represent Boolean assignments, \Bss, over a set of atoms $\mathcal{A}\cup\mathcal{C}$ by sets of \emph{signed literals} $\Tlit{a}$ or $\Flit{a}$ 
standing for $a\mapsto\Tsign$ and $a\mapsto\Fsign$, respectively, where $a\in\mathcal{A}\cup\mathcal{C}$.
The complement of a signed literal $\sigma$ is denoted by $\overline{\sigma}$.
We define
\(
\Tass{\Bss} = \{a \in\mathcal{A}\cup\mathcal{C} \mid \Tlit{a} \in \Bss\}
\)
and
\(
\Fass{\Bss} = \{a \in\mathcal{A}\cup\mathcal{C} \mid \Flit{a} \in \Bss\}
\).
Then,
an assignment $\Bss$ is \emph{complete},
if $\Tass{\Bss}\cap\Fass{\Bss}=\emptyset$ and $\Tass{\Bss}\cup\Fass{\Bss}=\mathcal{A}\cup\mathcal{C}$.
For instance,
the assignment
\(
\{\Tlit{a},\Flit{b},\Flit{c},\Flit{(x<7)}\}
\)
is complete wrt the atoms in our example.

Boolean constraints are represented as \emph{nogoods}.
A nogood is a set of signed literals representing an invalid partial assignment.
A nogood $\delta$ is \emph{violated} by a Boolean assignment \Bss\ whenever $\delta\subseteq\Bss$.
A complete Boolean assignment is a \emph{solution} of a set of nogoods,
if it violates none of them.
Given a Boolean assignment \Bss\ and a nogood $\delta$ such that
$\delta\setminus\Bss=\{\sigma\}$ and $\overline{\sigma}\notin\Bss$,
we say that $\delta$ is \emph{unit} wrt \Bss\ and \emph{asserts} the \emph{unit-resulting literal} $\overline{\sigma}$.
For a set $\Delta$ of nogoods and an assignment~\Bss, \emph{unit propagation}
is the iterated process of extending \Bss\ with unit-resulting literals until
no further literal is unit-resulting for any nogood in $\Delta$.

With these concepts in hand,
the Boolean constraints induced by a logic program $P$ can be captured as follows:
%
\begin{eqnarray}
  \nonumber 
  \Delta_P
  & = & 
  \textstyle{\bigcup}_{\substack{B\in\body{P},\\B=\{a_1,\dots,a_m,\naf{\,a_{m+1}},\dots,\naf{\,a_n}\}}}
  \left\{
    \begin{array}{l}
      \{\Flit{B},\Tlit{a_1},\dots,\Tlit{a_m},\Flit{a_{m+1}},\dots,\Flit{a_n}\},
      \\
      \{\Tlit{B},\Flit{a_1}\},\dots,\{\Tlit{B},\Flit{a_m}\},
      \\
      \{\Tlit{B},\Tlit{a_{m+1}}\},\dots,\{\Tlit{B},\Tlit{a_n}\}
    \end{array}
  \right\}
  \\
  \label{eq:completion:nogoods:two}
  &\cup&
  \textstyle{\bigcup}_{\substack{a\in\atom{P},\\\atbody{P}{a}=\{B_1,\dots,B_k\}}}
  \left\{
    \begin{array}{l}
      \{\Tlit{a},\Flit{B_1},\dots,\Flit{B_k}\},
      \\
      \{\Flit{a},\Tlit{B_1}\},\dots,\{\Flit{a},\Tlit{B_k}\}
    \end{array}
  \right\}
  \\
  \label{eq:loop:nogoods}
  \Lambda_P
  & = &
  \textstyle{\bigcup}_{\substack{U\subseteq\atom{P},\\\EB{U}{P}=\{B_1,\dots,B_k\}}}
  \left\{
    \{\Tlit{a},\Flit{B_1},\dots,\Flit{B_k}\} \mid a\in U
  \right\}
  \\\nonumber
  &&\text{where }
  \EB{U}{P}
  =
  \{\body{r}\in P\mid\head{r}\in U, {\body{r}}\cap U=\emptyset\}\/.
\end{eqnarray}
%
Then, according to~\cite{gekakasc12a},
a set of atoms $X$ is a stable model of a regular logic program $P$
iff
$X=\Tass{\Bss}\cap\atom{P}$ for a (unique) solution \Bss\ of $\Delta_P\cup\Lambda_P$.

For example,
the nogoods obtained in \eqref{eq:completion:nogoods:two} for the atom $a$ in our example are
\(
\{\Tlit{a},\Flit{\{\naf{b}}\}\}
\)
and
\(
\{\Flit{a},\Tlit{\{\naf{b}}\}\}
\).
Similarly, the body $\{\naf{b}\}$ of Rule~\eqref{rule:one} gives rise to nogoods
\(
\{\Flit{\{\naf{b}}\},\Flit{b}\}
\)
and
\(
\{\Tlit{\{\naf{b}}\},\Tlit{b}\}
\).
Hence, once an assignment contains \Tlit{a}, we may derive \Flit{b} via unit propagation
(using both the first and last nogood).

To extend this characterization to programs with constraint atoms,
it is important to realize that the truth value of such atoms is determined external to the program.
In CASP, this is reflected by the requirement that constraint atoms must not occur in the head of rules.%
\footnote{In alternative semantic settings, theory atoms may also occur as rule heads (cf.\ \cite{gekakaosscwa16a}).}
Hence, treating constraint atoms as regular ones leaves them unfounded.
For instance, in our example, we would get from both \eqref{eq:completion:nogoods:two} and \eqref{eq:loop:nogoods} the nogood
\(
\{\Tlit{(x<7)}\}
\),
which would set $(x<7)$ permanently to false.
To address this issue,
\cite{drewal12a} exempt constraint atoms from the respective sets of nogoods and define the variants
$\Delta_P^{\mathcal{C}}$ and $\Lambda_P^{\mathcal{C}}$
by replacing \atom{P} in the qualification of \eqref{eq:completion:nogoods:two} and \eqref{eq:loop:nogoods} with $\atom{P}\setminus\mathcal{C}$.

Then, in~\cite{ostrowski17a} it is shown that
$(X,\Css)$ is a constraint stable model of a program $P$ wrt $(\mathcal{V},D,C)$ as defined in~\cite{geossc09a}
iff
and
$X=\Tass{\Bss}\cap\atom{P}$ for a (unique) solution \Bss\ of 
\(
\Delta_P^{\mathcal{C}}\cup\Lambda_P^{\mathcal{C}}
\cup
\{\{\Flit{c}\}\mid\gamma(c)\in\mathit{sat}_{\Css}(C)\}
\cup
\{\{\Tlit{c}\}\mid\overline{\gamma(c)}\in\mathit{sat}_{\Css}(C)\}
\).

Accordingly,
our example yields the following constraint stable models
\begin{align}\label{eq:models}
\begin{array}{ll}
X & \Css\\
\{a\}  & x\in\{7,\dots,10\} \\
\{b\}  & x\in\{7,\dots,10\} \\
\{b,  x<7\} & x\in\{1,\dots,6\} \\
\{a,c,  x<7\} &  x\in\{1,\dots,6\} \\
\end{array}
\end{align}
where $x\in\{m,\dots,n\}$ means that either $x\mapsto m$, or $x\mapsto {m+1}$, \dots or $x\mapsto n$.
For instance, the very first constraint stable model corresponds to the Boolean assignment
\(
\{\Tlit{a},\Flit{b},\Flit{c},\Flit{(x<7)}\}
\)
paired with the constraint variable assignment
\(
\{x\mapsto 7\}
\).
 
Similar to logic programs,
linear constraints can be represented as sets of nogoods by means of an \emph{order encoding}~\cite{tatakiba09a}.
This amounts to representing the above unit nogoods
\(
\{\{\Flit{c}\}\mid\gamma(c)\in\mathit{sat}_{\Css}(C)\}
\cup
\{\{\Tlit{c}\}\mid\overline{\gamma(c)}\in\mathit{sat}_{\Css}(C)\}
\)
by more elaborate nogoods capturing the semantics provided by $\mathit{sat}_{\Css}(C)$.

To this end,
we let $\mathcal{O}_{\mathcal{V}}$ stand for the set of order atoms associated with variables in $\mathcal{V}$
and require it to be disjoint from $\mathcal{A}\cup\mathcal{C}$.
Whenever the set $\mathcal{V}$ is clear from the context, we drop it and simply write $\mathcal{O}$.
More precisely,
we introduce
an \emph{order atom} $(v \leq d) \in \mathcal{O}$
for each constraint variable $v\in\mathcal{V}$ and value $d \in \dom{v}, d \neq \mathit{ub}(v)$.
We refer to signed literals over $\mathcal{O}$ as \emph{signed order literals}.

Now, we are ready to map a linear CSP $(\mathcal{V},D,C)$ into a set of nogoods.

First,
we need to make sure that each variable in $\mathcal{V}$ has exactly one value from its domain in $D$.
To this end, we define the following set of nogoods.
\begin{equation}
\label{eq:order:nogoods:one}
\begin{split}
\Phi(\mathcal{V},D) = \{\{\Tlit{(v\leq d)},\Flit{(v\leq\mathit{next}(d,v))}\} \mid{} & v\in\mathcal{V},d\in\dom{v},\\ & \mathit{next}(d,v)<\mathit{ub}(v)\}
\end{split}
\end{equation}
Intuitively, each such nogood stands for an implication ``${(v\leq d)} \Rightarrow {(v\leq {d+1})}$''.
In our example,
we get the following nogoods.
\begin{align}\label{ex:order:nogoods:one}
\Phi(\{x\},\{D(x)\}) &=
\{
\{\Tlit{(x\leq 1)},\Flit{(x\leq 2)}\}
,\dots,
\{\Tlit{(x\leq 8)},\Flit{(x\leq 9)}\}
\}.
\end{align}

Second,
we need to establish the relation between constraint atoms $\mathcal{C}$ and their associated linear constraints in $C$.
Following~\cite{fesost11a},
a \textit{reified constraint} is an equivalence ``$\Tlit{c} \Leftrightarrow \gamma(c)$'' where $c\in\mathcal{C}$;
it is decomposable into two \emph{half-reified} constraints ``$\Tlit{c}\Rightarrow\gamma(c)$'' and ``$\Flit{c}\Rightarrow\overline{\gamma(c)}$''.
To proceed analogously, we extend $\gamma$ to signed literals over $\mathcal{C}$ as follows:
\[
\gamma(\sigma)
=
\left\{
\begin{array}{ll}
\overline{\gamma(a)} & \text{ if } \sigma=\Fsign{a}, a\in\mathcal{C}
\\
          \gamma(a)  & \text{ if } \sigma=\Tsign{a}, a\in\mathcal{C}
\end{array}
\right.
\]
For instance, we have
\(
\gamma(\Flit{(x<7)}) = (-x \leq -7)
\).

To translate constraints into nogoods,
we need to translate expressions of the form $av + b\leq 0$
for $v\in \mathcal{V}$ and integers $a,b$ into signed ordered literals.%
\footnote{Any linear inequality using $<,>,\leq,\geq$ and one variable can be converted into this form.}
Following~\cite{tatakiba09a},
we then define $\tolit{av + b \leq 0}$ as
\[
\tolit{av + b \leq 0} =
\left\{
\begin{array}{ll}
\tolitorder{v \leq \lfloor \frac{-b}{a}\rfloor} &\text{if } a > 0\vspace{2mm}
\\
\overline{\tolitorder{v \leq \lceil \frac{-b}{a}\rceil-1}} &\text{if } a < 0
\end{array}
\right.
\]
where $\tolitorder{v\leq d}$ is defined for $\mathit{lb}(v) \leq d < \mathit{ub}(v)$ as
\[
\tolitorder{v\leq d} =
\left\{
\begin{array}{ll}
\Tlit{(v \leq d)} &\text{if }d\in\dom{v}
\\
\Tlit{(v \leq \mathit{prev}(d,v))}&\text{if } d\notin\dom{v}
\end{array}
\right.
\]
If $d\geq\mathit{ub}(v)$ then $\tolitorder{v\leq d}=\Tlit{\emptyset}$;
if $d <\mathit{lb}(v)$ then $\tolitorder{v\leq d}=\Flit{\emptyset}$,
where $\emptyset$ stands for the empty body.%
\footnote{We use \Tlit{\emptyset} and \Flit{\emptyset} as representatives for tautological and unsatisfiable signed literals;
  they are removed in \eqref{eq:order:nogoods:two:a} and \eqref{eq:order:nogoods:two:b} below.}
Expressing our example constraint $x\leq 6$ in terms of signed order literals results in
\(
\tolit{1\cdot {x+(-6)}\leq 0}=\Tlit{(x\leq 6)}
\).
The signed literal \Tlit{(x\leq 6)} indicates that $6$ is the largest integer satisfying the constraint.
Also,
we get the signed literals
$\tolitorder{x \leq 0} = \Flit\emptyset$ and $\tolitorder{x \leq 10} = \Tlit\emptyset$.

We sometimes use $<$,$>$, or $\geq$ as operators in these expressions and 
implicitly convert them to the normal form $av + b \leq 0$ to be used in this translation.
Accordingly, the complementary constraint yields
\(
\tolit{x>6} = \tolit{(-1)\cdot {x + 7}\leq 0} = \overline{\tolitorder{x \leq \lceil \frac{-7}{-1}\rceil -1 }} =\Flit{(x\leq 6)}
\).

The actual relation between the constraint atoms in $\mathcal{C}$ and their associated linear constraints in $C$ is established via the following nogoods.
\begin{align}\label{eq:order:nogoods:two}
\Psi(\mathcal{C})=\textstyle \bigcup_{c\in\mathcal{C}}{\psi(\Tlit{c},\gamma(c)) \cup \psi(\Flit{c},\overline{\gamma(c)})} 
\ .  
\end{align}
For all constraint atoms $c\in\mathcal{C}$ associated with the linear constraint $\gamma(c)=\sum_{i=1}^{n}{a_iv_i} \leq b$ in $C$,
we define for both of its half-reified constraints the set of nogoods
\begin{align}\label{eq:order:nogoods:two:a}
\textstyle \psi(\Tlit{c},          \sum_{i=1}^{n}{a_iv_i} \leq b) &=\textstyle \{ \{\Tlit{c}\} \cup \delta \setminus \{\Tlit{\emptyset}\} \mid \delta \in \phi(          \sum_{i=1}^{n}{a_iv_i} \leq b ), \Flit{\emptyset}\notin\delta \}
\\\label{eq:order:nogoods:two:b}
\textstyle \psi(\Flit{c},\overline{\sum_{i=1}^{n}{a_iv_i} \leq b})&=\textstyle \{ \{\Flit{c}\} \cup \delta \setminus \{\Tlit{\emptyset}\} \mid \delta \in \phi(\overline{\sum_{i=1}^{n}{a_iv_i} \leq b}), \Flit{\emptyset}\notin\delta \}
\end{align}
where
\begin{align*}\textstyle
\phi(\sum_{i=1}^{n}{a_iv_i} \leq b)
&=
\left\{
\begin{array}{lr}
\{\tolit{a_1v_1> b}\}&\text{if } n=1\\
\{\tolit{a_1v_1\geq d}\} \cup \delta &\text{if } n>1\\
\qquad \delta \in \phi(\sum_{i=2}^{n}{a_iv_i} \leq b-d),d\in\img{a_1v_1}
\end{array}
\right\}
\end{align*}
Note that nogoods with \Tlit{\emptyset} and \Flit{\emptyset} are simplified in \eqref{eq:order:nogoods:two:a} and \eqref{eq:order:nogoods:two:b}.
Also, observe that the definition of $\phi$ is recursive although this does not show with our simple examples.

In our example, we obtain
\begin{align}
\label{ex:order:nogoods:two}
\psi(\Tlit{(x<7)},x \leq 6)
&=
\{\{\Tlit{(x<7)},\Flit{(x\leq 6)}\}\}
\\
\label{ex:order:nogoods:tri}
\psi(\Flit{(x<7)},-x \leq -7)
&=
\{\{\Flit{(x<7)},\Tlit{(x\leq 6)}\}\}
\end{align}
Taken together,
both nogoods realize the aforementioned equivalence between the constraint atom $(x<7)$ and its associated constraint.
Note that $(x<7)$ is a constraint atom in $\mathcal{C}$, while $(x\leq 6)$ is an order atom in $\mathcal{O}$
and thus belongs to the encoding of the constraint associated with $(x<7)$.
For further illustration, reconsider the Boolean assignment
\(
\{\Tlit{a},\Flit{b},\Flit{c},\Flit{(x<7)}\}
\)
inducing the first constraint stable models in~\eqref{eq:models}.
Applying unit propagation, we get \Flit{(x\leq 6)} via~\eqref{ex:order:nogoods:tri} and in turn
\Flit{(x\leq 5)} to \Flit{(x\leq 1)} via the nogoods in $\Phi(\{x\},\{D(x)\})$ in~\eqref{ex:order:nogoods:one}.
Similarly,
making \Tlit{(x\leq 7)} true yields \Tlit{(x\leq 8)} and \Tlit{(x\leq 9)} also via the nogoods in \eqref{ex:order:nogoods:one}.

All in all,
a CSP $(\mathcal{V},D,C)$ is characterized by the nogoods in $\Phi(\mathcal{V},D)$ and $\Psi(\mathcal{C})$.

While in~\eqref{eq:models} the corresponding constraint variable assignment \Css\ is determined externally,
it can be directly extracted from a solution $\Bss$ for $\Phi(\mathcal{V},D)$
by means of the following functions:
The upper bound for a view $v$ relative to a Boolean assignment \Bss\ is given by
\(
\mathit{ub}_{\Bss}(v)=\min(\{\mathit{ub}(v)\} \cup \{d\mid d\in\img{v}, \tolit{v\leq d}\in\Bss\})
\)
and its lower bound by 
\(
\mathit{lb}_{\Bss}(v)=\max(\{\mathit{lb}(v)\} \cup \{d\mid d\in\img{v}, \tolit{v\geq d}\in\Bss\})
\).
Then,
$\Css(v)=\mathit{lb}_{\Bss}(v)=\mathit{ub}_{\Bss}(v)$ for all $v\in\mathcal{V}$.
Accordingly, the above Boolean assignment corresponds to $\Css=\{x\mapsto 7\}$.

Combining the nogoods stemming from the logic program and its associated CSP,
we obtain the following characterization of constraint logic programs.
\begin{theorem}\label{thm:casp:nogoods}
Let $P$ be a constraint logic program over $\mathcal{A}\cup\mathcal{C}$
associated with the CSP $(\mathcal{V},D,C)$
and let $X\subseteq{\mathcal{A}\cup\mathcal{C}}$ and \Css\ a total assignment over $\mathcal{V}$.

Then,
$(X,\Css)$ is a constraint stable model of $P$ wrt $(\mathcal{V},D,C)$ as defined in~\cite{geossc09a}
iff
$(\Css,\mathit{sat}_{\Css}(C))$ is a configuration for $(\mathcal{V},D,C)$,
$X=\Tass{\Bss}\cap\atom{P}$ for a (unique) solution \Bss\ of 
\(
\Delta_P^{\mathcal{C}}\cup\Lambda_P^{\mathcal{C}}\cup\Psi(\mathcal{C})\cup\Phi(\mathcal{V},D)
\),
and
$\Css = \{v\mapsto \mathit{lb}_{\Bss}(v)\mid v\in\mathcal{V}\}$.
\end{theorem}
%
The proof of this theorem is obtained by combining existing characterizations of logic programs in terms of nogoods and similar ones for CSPs in terms
of clauses in CNF~\cite{ostrowski17a}.

\paragraph{Nogood propagators.}
The basic idea of lazy constraint propagation is to make the nogoods in $\Psi(\mathcal{C})$ and $\Phi(\mathcal{V},D)$ only explicit when needed.
This is done by propagators corresponding to the respective set of nogoods.
A popular example of this is the unfounded-set algorithm in ASP solvers that only makes the nogoods in $\Lambda_P$ in \eqref{eq:loop:nogoods} 
explicit when needed. 
 
Following~\cite{drewal12a},
a \emph{propagator} for a set $\Theta$ of nogoods is
a function $\Pi_\Theta$ mapping a Boolean assignment $\Bss$ to a subset of $\Theta$
such that for each total assignment $\Bss$:
if $\delta\subseteq\Bss$ for some $\delta\in\Theta$,
then $\delta'\subseteq\Bss$ for some $\delta'\in\Pi_\Theta(\Bss)$.
That is, whenever there is a nogood in $\Theta$ violated by an assignment \Bss,
then $\Pi_\Theta(\Bss)$ yields a violated nogood, too.
A propagator $\Pi_\Theta$ is \emph{conflict optimal},
if for all partial assignments \Bss,
the violation of a nogood in $\Theta$ by \Bss\
implies that some nogood in $\Pi_\Theta(\Bss)$ is violated by \Bss.
$\Pi_\Theta$ is \emph{inference optimal},
if it is conflict optimal and
$\Pi_\Theta(\Bss)$ contains all unit nogoods of $\Theta$ wrt $\Bss$.

We obtain the following extension of Theorem~\ref{thm:casp:nogoods}.
\begin{theorem}\label{thm:casp:propagator}
Let $P$ be a constraint logic program over $\mathcal{A}\cup\mathcal{C}$
associated with the CSP $(\mathcal{V},D,C)$
and let $\Pi_\Theta$ be a propagator for $\Theta$~=~$\Lambda_P^{\mathcal{C}}$, $\Psi(\mathcal{C})$, and $\Phi(\mathcal{V},D)$, respectively.

Then,
\Bss\ is a solution of
\(
\Delta_P^{\mathcal{C}}\cup\Lambda_P^{\mathcal{C}}\cup\Psi(\mathcal{C})\cup\Phi(\mathcal{V},D)
\)
iff
\Bss\ is a solution of
\[
\Delta_P^{\mathcal{C}}\cup\Pi_{\Lambda_P^{\mathcal{C}}}(\Bss)\cup\Pi_{\Psi(\mathcal{C})}(\Bss)\cup\Pi_{\Phi(\mathcal{V},D)}(\Bss)
\ .
\]
\end{theorem}
%
This theorem tells us that the nogoods in $\Psi(\mathcal{C})$, $\Phi(\mathcal{V},D)$, and $\Lambda_P^{\mathcal{C}}$
must not be explicitly represented but can be computed by corresponding propagators $\Pi_\Theta$ that add them lazily when needed.

To relax the restrictions imposed by this theorem,
the idea is to compile out a subset of constraints and variables of the CSP 
while leaving the others subject to lazy constraint propagation.
This is captured by the following corollary to Theorem~\ref{thm:casp:propagator}.
\begin{corollary}\label{cor:partial}
Let $P$ be a constraint logic program over $\mathcal{A}\cup\mathcal{C}$
associated with the CSP $(\mathcal{V},D,C)$
and let $\Pi_\Theta$ be a propagator for $\Theta$~=~$\Lambda_P^{\mathcal{C}}$, $\Psi(\mathcal{C}\setminus \mathcal{C}')$, and $\Phi(\mathcal{V}\setminus\mathcal{V}',D\setminus D')$, respectively,
for subsets $\mathcal{C}' \subseteq \mathcal{C}$, $\mathcal{V}' \subseteq \mathcal{V}$, and $\mathcal{D}' \subseteq \mathcal{D}$.

Then,
\Bss\ is a solution of
\(
\Delta_P^{\mathcal{C}}\cup\Lambda_P^{\mathcal{C}}\cup\Psi(\mathcal{C})\cup\Phi(\mathcal{V},D)
\)
iff
\Bss\ is a solution of
\[
\Delta_P^{\mathcal{C}}\cup\Psi(\mathcal{C}')\cup\Phi(\mathcal{V}',D')
\cup
\Pi_{\Lambda_P^{\mathcal{C}}}(\Bss)\cup\Pi_{\Psi(\mathcal{C}\setminus \mathcal{C}')}(\Bss)\cup\Pi_{\Phi(\mathcal{V}\setminus\mathcal{V}',D\setminus D')}(\Bss)
\ .
\]
\end{corollary}
%
This correspondence nicely reflects upon the basic idea of our approach.
While the entire set of loop nogoods $\Lambda_P^{\mathcal{C}}$ is handled by the unfounded set propagator $\Pi_{\Lambda_P^{\mathcal{C}}}(\Bss)$ as usual,
the ones capturing the CSP is divided among the explicated nogoods in 
$\Psi(\mathcal{C}')\cup\Phi(\mathcal{V}',D')$ 
and the implicit ones handled by the propagators
$\Pi_{\Psi(\mathcal{C}\setminus \mathcal{C}')}(\Bss)$ and $\Pi_{\Phi(\mathcal{V}\setminus\mathcal{V}',D\setminus D')}(\Bss)$.
Note that variables and domain elements are often only dealt with implicitly through their induced order atoms in $\mathcal{O}$.


%% file: approach.tex
\section{The \clingcon\ system}\label{sec:approach}

We now detail various aspects of the new \clingcon~3 system.
We begin with an overview of its architecture along with its salient components.
The next sections detail its input language and major algorithms.
The subsequent section is dedicated to distinguished \clingcon\ features,
which are experimentally evaluated in Section~\ref{sec:experiments}.
Finally, we illustrate in the last section \clingcon's multi-shot solving capabilities by discussing several incremental solutions to the $n$-queens puzzle.


%% file: architecture.tex
\subsection{Architecture}\label{sec:architecture}

\clingcon~3 is an extension of the ASP system \clingo~5,
which itself relies on the grounder \gringo\ and the solver \clasp.
The architecture of \clingcon~3 is given in Figure~\ref{fig:architecture}.
\begin{figure}[h]
\centering
\input{clingconarchi}
\caption{Architecture of \clingcon~3}.
\label{fig:architecture}
\end{figure}
More precisely, \clingcon\ uses \gringo's capabilities to specify and process customized theory languages.
For this, it is sufficient to supply a grammar fixing the syntax of constraint-related expressions.
As detailed in Section~\ref{sec:language}, 
this allows us to express linear constraints similar to standard ASP aggregates by using first-order variables.
Unlike this, \clingcon\ extends \clasp\ in several ways to accommodate its lazy approach to constraint solving.
First,
\clasp's preprocessing capabilities are extended to integrate linear constraints. 
Second,
dedicated propagators are added to account for lazy constraint propagation.
Both extensions are detailed in Section~\ref{sec:algorithms} and~\ref{sec:features}.
And finally,
a special output module was created to integrate CSP solutions.
Notably,
\clingcon\ pursues a lazy yet two-fold approach to constraint solving that 
allows for making a part of the nogoods in $\Psi(\mathcal{C})$ explicit during preprocessing,
while leaving the remaining constraints implicit and the creation of corresponding nogoods subject to the constraint propagator.
In this way, a part of the CSP can be put right up front under the influence of CDCL-based search.
All other constraints are only turned into nogoods when needed.
Accordingly, only a limited subset of order atoms from $\mathcal{O}$ must be introduced at the beginning;
further ones are only created if they are needed upon the addition of new nogoods.
This is also called lazy variable generation.

It is worth mentioning that both the grounding and the solving component of \clingcon\ can also be used separately via \clingo's option `\texttt{--mode}'.
That is, the same result as with \clingcon\ is obtained by passing the output of `\texttt{clingcon --mode=gringo}' to `\texttt{clingcon --mode=clasp}'.
The intermediate result of grounding a CASP program is expressed in the \aspif\ format~\cite{gekakaosscwa16b} that accommodates both the regular ASP
part of the program as well as its constraint-based extension.
This modular design allows others to take advantage of \clingcon's infrastructure for their own CASP solvers.
Also, other front ends can be used for generating ground CASP programs; eg.\ the \flatzinc\ translator used in Section~\ref{sec:experiments}.

Finally, extra effort was taken to transfer \clasp\ specific features to \clingcon's solving component.
This includes
multi-threading~\cite{gekasc12b},
unsatisfiable core techniques~\cite{ankamasc12a},
multi-criteria optimization~\cite{gekakasc11c},
domain-specific heuristics~\cite{gekaotroscwa13a},
multi-shot solving~\cite{gekaobsc15a},
and \clasp's reasoning modes like enumeration, intersection and union of models.
Vocabulary-sensitive reasoning modes like projective enumeration and domain-specific heuristics can be used via auxiliary atoms.


%% file: clingconarchi.tex
\setlength{\unitlength}{1.25pt}
\begin{picture}(270,90)
  \put(  2, 20){\dashbox(33,30){\small\shortstack{CASP\\Program}}}
  \put( 50, 10){\framebox(170,60)[t]{\shortstack{\\~\\\clingcon}}}
  \put( 60, 20){\framebox(60,30){\gringo}}
  \put(150, 20){\framebox(60,30){\clasp\qquad}}      
  \put(185, 25){\framebox(20,10){\small CSP}}      
  \put(235, 20){\dashbox(33,30){\small\shortstack{CASP\\Solution}}}
  \put(120, 35){\vector(1,0){30}}
  \put( 35, 35){\vector(1,0){25}}
  \put(210, 35){\vector(1,0){25}}
  \put(  2, 55){\dashbox(33,30){\small\shortstack{CSP\\Grammar}}
    \put(0, 15){\line(1,0){5}}
    \put(5, 15){\line(0,-1){35}}
  }
\end{picture}%
%

%% file: language.tex
\subsection{Language}\label{sec:language}

As mentioned,
the treatment of the extended input language of CASP programs can be mapped onto \gringo's theory language capabilities~\cite{gekakaosscwa16a}.
For this, it is sufficient to supply a corresponding grammar fixing the syntax of the language extension.
The one used for \clingcon\ is given in Listing~\ref{encoding:csp}.
\begin{lstlisting}[float,caption={Language Syntax},label=encoding:csp]
#theory csp {
    dom_term {
    + : 5, unary;
    - : 5, unary;
    .. : 1, binary, left;
    * : 4, binary, left;
    + : 3, binary, left;
    - : 3, binary, left
    };
    linear_term {
    + : 5, unary;
    - : 5, unary;
    * : 4, binary, left;
    + : 3, binary, left;
    - : 3, binary, left
    };
    show_term {
    / : 1, binary, left
    };
    minimize_term {
    + : 5, unary;
    - : 5, unary;
    * : 4, binary, left;
    + : 3, binary, left;
    - : 3, binary, left;
    @ : 0, binary, left
    };

    &dom/0 : dom_term, {=}, linear_term, any;
    &sum/0 : linear_term, {<=,=,>=,<,>,!=}, linear_term, any;
    &distinct/0 : linear_term, any;
    &show/0 : show_term, directive;
    &minimize/0 : minimize_term, directive
}.

\end{lstlisting}
The grammar is named \texttt{csp} and consists of two parts,
one defining theory terms in lines~2-27 and another defining theory atoms in lines~29-33.
All regular terms are implicitly included in the respective theory terms.
Theory terms are then used to represent constraint-related expressions that are turned by grounding into 
linear constraint atoms using predicate \texttt{\&sum},
domain restrictions using predicate \texttt{\&dom},
directives \texttt{\&show} and \texttt{\&minimize},
and the predefined global constraint \texttt{\&distinct}.

Before delving into further details,
let us illustrate the resulting syntax by the CASP program for two dimensional strip packing given in Listing~\ref{encoding:2sp},
originally due to~\cite{sointabana10a}.
\begin{lstlisting}[float,caption={Two Dimensional Strip Packing},label=encoding:2sp]
&dom{0..w-W}  = x(I) :- r(I,W,H).
&dom{0..ub-H} = y(I) :- r(I,W,H).

1 { le(x(I),WI,x(J));
    le(x(J),WJ,x(I));
    le(y(I),HI,y(J));
    le(y(J),HJ,y(I)) } :- r(I,WI,HI), r(J,WJ,HJ), I < J.

&sum{VI; C} <= VJ :- le(VI,C,VJ).

&dom{0..ub} = height.
&sum{y(I); H} <= height :- r(I,W,H).
&minimize {height}.
&show {height}.

\end{lstlisting}
Given a set of rectangles, each represented by a fact \texttt{r(I,W,H)}
where \texttt{I} identifies a rectangle with width \texttt{W} and height \texttt{H},
the task is to fit all into a container of width \texttt{w} and height \texttt{ub}
while minimizing the needed height of the container.
The first two lines of Listing~\ref{encoding:2sp} restrict the domain of the left lower corner of each rectangle \texttt{I}.
The respective instantiations of \texttt{x(I)} and \texttt{y(I)} yield constraint variables denoting the \texttt{x} and \texttt{y} coordinate of \texttt{I}, respectively.
Note that in both lines the consecutive dots `\texttt{..}' construct a theory term `\texttt{0..w-W}' and `\texttt{0..ub-H}' 
once \texttt{w} and \texttt{ub} are replaced, respectively.
The choice rule in Line~4-7 lets us choose among all combinations of two rectangles, that is, which one is left, right, below or above.
At least one of these relations must hold so that no two rectangles overlap.
Atoms of form \texttt{le(VI,C,VJ)} indicate that coordinate \texttt{VI}+\texttt{C} must be less than or equal to \texttt{VJ}.
This property is enforced by the linear constraint in Line~9.
Finally, to minimize the overall height of (stacked) rectangles, we introduce the variable \texttt{height}.
This variable's value has to be greater than or equal to the \texttt{y} coordinate of any rectangle \texttt{I} plus the rectangle's height \texttt{H}.
This ensures that \texttt{height} is greater or equal to the height of the highest rectangle.
Finally, \texttt{height} is minimized in Line~13.

Now, if we take the three rectangles \texttt{r(a,5,2)}, \texttt{r(b,2,3)}, \texttt{r(c,2,2)} along with \texttt{ub=10} and \texttt{w=6},
we obtain the ground program in Listing~\ref{encoding:2sp:ground}.
\begin{lstlisting}[float,caption={Two Dimensional Strip Packing Example},label=encoding:2sp:ground]
r(a,5,2). r(b,2,3). r(c,2,2).

&dom{0..(6-5)}  = x(a). &dom{0..(6-2)}  = x(b). &dom{0..(6-2)}  = x(c).
&dom{0..(10-2)} = y(a). &dom{0..(10-3)} = y(b). &dom{0..(10-2)} = y(c).

1 <= { le(x(a),5,x(b)); le(x(b),2,x(a));
       le(y(a),2,y(b)); le(y(b),3,y(a)) }.
1 <= { le(x(a),5,x(c)); le(x(c),2,x(a));
       le(y(a),2,y(c)); le(y(c),2,y(a)) }.
1 <= { le(x(b),2,x(c)); le(x(c),2,x(b));
       le(y(b),3,y(c)); le(y(c),2,y(b)) }.

&sum{ x(a); 5 } <= x(b) :- le(x(a),5,x(b)).
&sum{ x(b); 2 } <= x(a) :- le(x(b),2,x(a)).
&sum{ y(a); 2 } <= y(b) :- le(y(a),2,y(b)).
&sum{ y(b); 3 } <= y(a) :- le(y(b),3,y(a)).
&sum{ x(a); 5 } <= x(c) :- le(x(a),5,x(c)).
&sum{ x(c); 2 } <= x(a) :- le(x(c),2,x(a)).
&sum{ y(a); 2 } <= y(c) :- le(y(a),2,y(c)).
&sum{ y(c); 2 } <= y(a) :- le(y(c),2,y(a)).
&sum{ x(b); 2 } <= x(c) :- le(x(b),2,x(c)).
&sum{ x(c); 2 } <= x(b) :- le(x(c),2,x(b)).
&sum{ y(b); 3 } <= y(c) :- le(y(b),3,y(c)).
&sum{ y(c); 2 } <= y(b) :- le(y(c),2,y(b)).

&dom{ 0..10 } = height.

&sum{ y(a); 2 } <= height.
&sum{ y(b); 3 } <= height.
&sum{ y(c); 2 } <= height.

&minimize{ height }.
&show{ height }.
\end{lstlisting}
The domains of the constraint variables giving the x- and y-coordinates are delineated in Line~3 and~4.
Note that in contrast to regular ASP the grounder leaves terms with the theory symbol \texttt{..} intact.
The orientation of each pair of rectangles is chosen in Lines~6-11.
If for example \texttt{le(x(c),2,x(b))} becomes true, that is, rectangle $c$ is left of $b$,
then the constraint $x(c) + 2 \leq x(b)$ is enforced in Line~22.
After setting the domain for the \texttt{height} variable in Line~26,
we restrict it to be greater or equal to the top y-coordinate of all rectangles in Lines~28-30.
Line~32 enforces the minimization of this variable.
A solution with minimal \texttt{height} consists of the regular atoms
\texttt{le(y(b),3,y(a))}, \texttt{le(y(c),2,y(a))}, and \texttt{le(x(c),2,x(b))}
and the constraint variable assignment
$\{\texttt{height}\mapsto 5, \texttt{y(c)}\mapsto 1, \texttt{x(c)}\mapsto 2, \texttt{x(a)}\mapsto 1, \texttt{x(b)}\mapsto 4, \texttt{y(a)}\mapsto 3, \texttt{y(b)}\mapsto 0\}$.
Of course other minimal configurations exist.

We have seen above how seamlessly theory atoms capturing constraint-related expressions can be used in logic programs.
We detail below the five distinct atoms featured by \clingcon\ and refer the interested reader for a general introduction to theory terms and atoms
to~\cite{gekakaosscwa16a}.

Actual constraints are represented by the theory atoms \texttt{\&dom}, \texttt{\&sum}, and \texttt{\&distinct}.
All three can occur in the head and body of rules, as indicated by \texttt{any} in Line~29-31 in Listing~\ref{encoding:csp}.
We discuss below their admissible format after grounding.
In the following, a linear expression is a sum of integers, products of integers, or products of an integer and a constraint variable.
\begin{description}
\item [Domain constraints] are of form $\texttt{\&dom}\{d_1;\dots;d_n\}=t$ where 
  \begin{itemize}
  \item each $d_i$ is a domain term of form
    \begin{itemize}
    \item $u$ or
    \item $v \mathtt{..} w$
    \end{itemize}
    where $u,v,w$ are constraint variable free linear expressions
    and 
  \item $t$ is a linear expression containing exactly one constraint variable.
  \end{itemize}
  Then, the previous expression represents the constraint
  $t \in \bigcup_{i=1}^{n} \llbracket d_i \rrbracket$,
  where
  \(
  \llbracket d \rrbracket = \{u\}
  \)
  if $d=u$,
  \(
  \llbracket d \rrbracket = \{v,\dots,w\}
  \)
  if $d={v \mathtt{..} w}$,
  and undefined otherwise.

  This constraint can be used to set the domain of variables
  where even non-contiguous domains can be used by having $n > 1$.
        For example $\texttt{\&dom}\{1..3;5\}=x$ represents the constraint $x\in\{1,\dots,3\} \cup \{5\}$.
\item [Linear constraints] are of form ${\texttt{\&sum}\{t_1;\dots;t_n\}}\circ t_{n+1}$ where
  \begin{itemize}
  \item each $t_i$ is a linear expression containing at most one constraint variable, and
  \item  $\circ$ is one of the operators \texttt{<=,=,>=,<,>,!=}
  \end{itemize}
This expression represents the linear constraint
$(t_1 + \dots + t_n) \circ t_{n+1}$,
which can be translated into one or two linear constraints as in~\eqref{eq:linear:constraint}.
\item [Distinct constraints] are of form $\texttt{\&distinct}\{t_1;\dots;t_n\}$ where each $t_i$
is a linear expression containing at most one constraint variable.
  Such an expression stands for the constraints $t_i \neq t_j$ for $0 \leq i < j \leq n$.

  The distinct constraint is one of the most common global constraints in CP\@.
  We use it to show how global constraints can be incorporated into the language.
\end{description}
The two remaining theory atoms provide directives, similar to their regular counterparts.
\begin{description}
\item [Output directives] are of form $\texttt{\&show}\{s_1;\dots;s_n\}$ where each $s_i$ is a show term of form
  \begin{itemize}
  \item $f/m$ where $f$ is a function symbol and $m$ a positive integer
  \item $t$, where $t$ is a constraint variable.
  \end{itemize}
While the latter adds variable $t$ to the list of output variables,
the first one adds all variables of the form $f(t_1,\dots,t_m)$ (where $t_i$ is a term) as output variables.
For all constraint stable models,
the value of the output variables is shown in a solution.

\item [Minimize directives] are of form $\texttt{\&minimize}\{m_1;\dots;m_n\}$ where each $m_i$ is a minimize term of form
$t_i@l_i$ and $t_i$ being a linear expression with at most one constraint variable.
Since we support multi-objective optimization, $l_i$ is an integer stating the
priority level.
Whenever $@l_i$ is omitted, it is assumed to be zero.
Priorities allow for representing lexicographically ordered minimization objectives.
As in regular ASP,
higher levels are more significant than lower ones.

Let us make precise how minimize statements induce optimal constraint stable models.
Let $P$ be a constraint logic program 
associated with $(\mathcal{V},D,C)$.
For a variable assignment \Css\ and an integer $l$,
define $\sum_l^\Css$ as the sum of all values $a\cdot \Css(v) + c$
for all occurrences of minimize terms $av+c@l$ in all minimize statements in $P$.
A constraint stable model $(X,\Css)$ of $P$ wrt $(\mathcal{V},D,C)$ is dominated if there is a
constraint stable model $(X',\Css')$ such that $\sum_l^{\Css'} < \sum_l^\Css$
and $\sum_{l'}^{\Css'} = \sum_{l'}^\Css$ for all $l' > l$,
and optimal otherwise.
Maximization can be achieved by multiplying each minimize term by $-1$.
\end{description}

Note that the set of constraints supported by \clingcon\ is only a subset of the constraints expressible with the syntax fixed in
Listing~\ref{encoding:csp}.
While for example expressions with more than one constraint variable are well-formed according to the syntax,
they are not supported by \clingcon.


%% file: algorithms.tex
\subsection{Algorithms}\label{sec:algorithms}

As mentioned,
\clingcon\ pursues a lazy approach to constraint solving that distinguishes two phases.
During preprocessing, any part of the nogoods representing a CSP can be made explicit and thus put right away under the influence of CDCL-based solving.
Unlike this, the remaining constraints are at first kept implicit and their corresponding nogoods are only added via constraint propagators to CDCL solving when needed.
This partitioning of constraints constitutes a trade-off.
On the one hand,
constraint propagators are usually slower than unit propagation,
in particular, when dealing with sets of nogoods of moderate size
because of modern SAT techniques such as the two-watched-literals scheme~\cite{zamamoma01a}.
On the other hand,
translating all constraints is often impracticable, in particular,
when dealing with very large domains.
Hence,
a good trade-off is to restrict the translation to ``small constraints''
in order to benefit from the high performance of CDCL solving
and to unfold ``larger constraints'' only by need.

In what follows,
we make \clingcon's two-fold approach precise 
by presenting algorithms for translation and propagation of constraints
before discussing implementation details in Section~\ref{sec:features}.

\paragraph{Partial Translation.}

Following Corollary~\ref{cor:partial},
a subset $\mathcal{C'}\subseteq\mathcal{C}$ of the constraint atoms
is used to create the set of nogoods $\Psi(\mathcal{C}')$.
%
Therefore,
Algorithm~\ref{algo:trans} creates a set of nogoods that is
equivalent to $\psi(\sigma, a_1v_1 + \dots + a_nv_n \leq b)$,
as defined in \eqref{eq:order:nogoods:two:a} and \eqref{eq:order:nogoods:two:b};
in turn, they are used to create $\Psi(\mathcal{C}')$ as shown in~\eqref{eq:order:nogoods:two}.
To this end, it is initially engaged by \TranslateArg{}{\{\sigma\}, a_1v_1 + \dots + a_nv_n \leq b}.
\input{translate}
We start the algorithm by
having $\sigma$ in our set of literals $\delta$, and
setting $d$ to the smallest value greater than $b-\sum_{j=2}^{n}{ub(a_jv_j)}$
in the image of $a_1v_1$.
This is the smallest value needed to violate the constraint.
If $d$ and the least sum $\sum_{j=2}^{n}{lb(a_jv_j)}$ added by all other views
is still less than $b$ in Line~4,
we have to recursively translate the rest of the constraint,
while subtracting $d$ from the right-hand side in Line~5.
Otherwise the constraint is already violated and we return
all nogoods created so far in Line~7.
We iteratively increase $d$ in Line~8 and repeat this process (Line~3)
for all values in $\img{a_1v_1}$.
Note that this also involves adding all order atoms
$
\mathcal{O}_{\Psi(\mathcal{C}')} = \bigcup_{\delta\in\Psi(\mathcal{C}')}{\Tass{\delta} \cup \Fass{\delta}}
$
included in the created nogoods $\Psi(\mathcal{C}')$ to the solver.

Which constraints to translate is subject to heuristics and command line options,
as explained in Section~\ref{sec:features}.

\paragraph{Extended Conflict Driven Constraint Learning.}
After translating a part of the problem into a set of nogoods $\Psi(\mathcal{C}')$,
using the order atoms $\mathcal{O}_{\Psi(\mathcal{C}')}\subseteq\mathcal{O}$,
we explain how to solve the remaining constraint logic program $P$ over $\mathcal{A}, \mathcal{C}$
associated with $(\mathcal{V},D,C)$.
Our algorithmic approach follows the one in \cite{drewal12a},
where a modified CDCL algorithm supporting external propagators is presented.
We extend this algorithm with lazy nogood and variable generation in Algorithm~\ref{cdnl}.
\input{cdcl}
The algorithm relies upon a growing set of Boolean variables $\mathcal{B}$,
which is initiated with all atoms (regular, constraint, and a subset of the order atoms in $\mathcal{O}_{\Psi(\mathcal{C}')}$),
and subsequently expanded by further order atoms.
Accordingly, the Boolean assignment \Bss\ is restricted to atoms in $\mathcal{B}$,
and recorded nogoods are accumulated in $\nabla$. 
%
Starting with an empty assignment,
the \Propagation\ method (Line~5),
extends the assignment $\Bss$ with propagated literals,
adds new nogoods to $\nabla$ and extends the set of atoms $\mathcal{B}$.
This method is detailed below in Algorithm~\ref{algo:propagate}.
When encountering a conflicting assignment (Line~6),
we either backtrack (Line~8) or,
if we cannot recover from the conflict,
return \emph{unsatisfiable}.
Whenever all atoms in $\mathcal{B}$ are assigned (Line~9),
we check whether a complete assignment for the variables in $\mathcal{V}$ is obtained from \Bss\ in Line~10.
If this is the case,
we return the constraint stable model $(\Tass{\Bss}\cap \atom{P},\{v \mapsto \mathit{lb}_{\Bss}(v) \mid v\in\mathcal{V}\})$.
Otherwise,
$\SplitArg{\mathcal{V},D}{\mathcal{B},\Bss}$ creates a new order atom
for the constraint variable with the currently largest domain
that splits the domain in half.
If we face an incomplete assignment,
we extend it using the \textsc{Select} function.

\input{propagation}
Algorithm~\ref{algo:propagate} reflects the proceeding of our propagators.
At first, \UnitPropagation\ is run on the completion nogoods $\Delta_P^{\mathcal{C}}$,
the nogoods from the partial translation $\Psi(\mathcal{C}')$,
and finally the already learned nogoods $\nabla$.
Then, propagator $\Pi_{\Lambda_P^{\mathcal{C}}}$ is engaged via \UfsPropagation.
If it does not add any new nogoods to $\nabla$,
\CspPropagation\ is called.
This method acts as a propagator, returning a set of nogoods $\nabla'$.
Since some of these nogoods may use new order atoms not introduced so far,
we dynamically extend the set of atoms $\mathcal{B}$ by
the atoms in $\Tass{\delta}\cup\Fass{\delta}$ stemming from the added nogoods $\delta\in\nabla'$.

New nogoods produced by any propagator are added to the set $\nabla$ of recorded nogoods
and propagation resumes afterwards (lines 6 and 11).
Notably, \CspPropagation\ is not run until a fixpoint is obtained.
However, its set of returned nogoods remains non empty until a fixpoint is reached.
In this way,
unit propagation interleaves with constraint propagation while delaying more complex propagation.
In all,
since unit propagation is much faster, it always precedes unfounded set propagation,
which again precedes constraint propagation.
This order reflects the complexity of the respective propagators,
so that the faster the propagation, the sooner it is engaged.

\paragraph{Lazy Variable Generation.}
Realizing \CspPropagation\ as a propagator for $\Pi_{\Psi(\mathcal{C})}$ and $\Pi_{\Phi(\mathcal{V},D)}$
allows for lazy nogood generation and for capturing inferences of the order encoding.
However, to be effective, lazy variable generation requires a different set of constraints to be propagated.
For illustration, suppose \CspPropagation\ is a propagator for $\Psi(\mathcal{C}) \cup \Phi(\mathcal{V},D)$.
Considering our example program $P_1$ along with $\Tlit{(x<7)} \in \Bss$ results in
\(
\CspPropagationArg{}{\emptyset, \emptyset, {\{\Tlit{(x<7)}\}}}=\{\{\Tlit{(x<7)},\Flit{(x\leq 6)}\}\},
\)
which is a subset of $\Psi(\mathcal{C})$ according to~\eqref{ex:order:nogoods:two}.
This nogood comprises the order atom $(x\leq 6)$ which is added to $\mathcal{B}$ in Line~10 of Algorithm~\ref{algo:propagate}.
Having this nogood, unit propagation adds in turn $\Tlit{(x\leq 6)}$ to the assignment in Line~2.
Then,
\(
\CspPropagationArg{(\mathcal{V},D,C)}{{\{(x\leq6)\}}, \emptyset, \{\Tlit{(x\leq6)}\}}
\)
yields the nogoods
\(
\{
\{{\Tlit{(x\leq 6)}}, {\Flit{(x\leq 7)}}\},
\dots,
\{{\Tlit{(x\leq 8)}}, {\Flit{(x\leq 9)}}\}
\}
\)
belonging to $\Phi(\mathcal{V},D)$
and produces the corresponding order atoms $(x\leq7),\dots,(x\leq 9)$.
We see that
once a certain upper bound $\Tlit{(v\leq x)}\in\Bss$ is found,
all order atoms in $\{(v\leq x') \mid x' > x, x'\in\dom{v}, x'<\mathit{ub}(v)\}$
are added to $\mathcal{B}$.
Similarly,
if a lower bound $\Flit{(v\leq x)} \in \Bss$ is fixed,
all order atoms $\{(v\leq x') \mid x' \leq x, x'\in\dom{v}\}$
are added to $\mathcal{B}$.
To avoid adding superfluous order atoms,
we let \CspPropagation\ be a propagator for $\Psi(\mathcal{C}) \cup \Phi'(\mathcal{V},D)$
where 
\begin{align*}
\Phi'(\mathcal{V},D)= \{\{\Tlit{(v\leq d)},\Flit{(v\leq e)}\} \mid v\in\mathcal{V},d\in\dom{v},e\in\dom{v}, d<e<\mathit{ub}(v)\}.
\end{align*}
Although $\Phi'(\mathcal{V},D)$ is a superset of $\Phi(\mathcal{V},D)$,
\CspPropagation\ only adds nogoods from $\Phi'(\mathcal{V},D)$ whose order atoms have already been introduced,
that is, $\{(v\leq d),(v\leq e)\} \subseteq \mathcal{B}$.
While $\Phi(\mathcal{V},D)$ contains for each variable $v$ a linear number of nogoods of form  $\{\Tlit{(v \leq d)}, \Flit{(v\leq \mathit{next}(d,v))}\}$,
$\Phi'(\mathcal{V},D)$ contains a quadratic number of nogoods for each variable.
The nogoods in $\Phi(\mathcal{V},D)$ allow for propagating the truth value of one order literal to its adjacent one.
Unlike this,
$\Phi'(\mathcal{V},D)$ contains redundant nogoods that allow for propagating the truth value
of one order literal to all greater ones
by means of nogoods of form $\{\Tlit{(v\leq d)},\Flit{(v\leq e)}\}$ for all values $e\in\dom{v}$ such that $d<e<\mathit{ub}(v)$.
Instead of ``chaining'' all values together,
the latter nogoods allow us to directly jump to any value.
As we restrict our propagator for $\Phi'(\mathcal{V},D)$ to only return nogoods
where all order atoms are included in $\mathcal{B}$,
no new order atoms are created.
In our example, this optimized \CspPropagation\ function does not return any nogoods,
viz.\
\(
\CspPropagationArg{(\mathcal{V},D,C)}{\{(x\leq6)\}, \emptyset, \{\Tlit{(x\leq6)}\}}=\emptyset
\),
as none of the order atoms $(x\leq7),\dots,(x\leq 9)$ are included in $\mathcal{B}$
and no propagation needs to be done.

\CspPropagation\ is depicted in Algorithm~\ref{algo:cspprop} and consists of two parts (lines 1-10 and 11-21).
\input{cspprop}
The first part starts with selecting the unit nogoods from $\Phi'(\mathcal{V},D)$.
For every variable $v\in\mathcal{V}$,
we check if it already has an upper bound $ub$ (lines 3-4)
given by $\Tlit{(v\leq ub)} \in \Bss$.
If this is the case,
we add the nogoods
\[
\{\{\Tlit{(v\leq ub)}, \Flit{(v\leq x)} \} \mid x>ub, (v\leq x)\in \mathcal{B}, \Tlit{(v\leq x)}\notin \Bss\}
\]
to $\Sigma$ to ensure consistency of all order atoms $(v\leq x) \in \mathcal{B}$ where $x>ub$ that are not already true.
Lines 6-8 do the same for current lower bound of the variable.
If any nogoods are found, they are immediately returned in Line~10.
The \Propagation\ function continues with unit propagation
on the new nogoods.
The second part of the constraint propagation (lines 11-21),
generating the nogoods in $\Psi(\mathcal{C}\setminus\mathcal{C}')$ lazily,
is only done if all order atoms are properly propagated,
i.e.\ no new nogoods have been generated in the first part (lines 1-10).
This is detailed in the next paragraph.

\paragraph{Constraint Propagation.}
To generate the nogoods in $\Psi(\mathcal{C}\setminus\mathcal{C}')$ lazily,
Algorithm~\ref{algo:cspprop} uses functions
\textsc{PropagateBounds} and \textsc{PropagateReification}
for half-reified constraint $\Tlit{c}\Rightarrow \gamma(c)$ and $\Flit{c}\Rightarrow \overline{\gamma(c)}$,
respectively,
for each $c \in \mathcal{C}\setminus\mathcal{C}'$.
In the respective algorithms~\ref{proptrue} and~\ref{propimpl},
we consider four different strengths of propagation,
denoted by $\mathit{ps}$.
A strength of 1 means that our propagator only produces
conflicting nogoods. 
A strength of 2 means that it additionally checks if yet
undecided constraints became true.
Strength 3 furthermore adds unit nogoods that also
propagate the bounds of the variables in a constraint if it is already
decided to be true,
whereas strength 4 also computes optimized nogoods for yet undecided constraints.
The propagators are conflict optimal
and for strength 4 even inference optimal.
We divided our propagator into two algorithms,
handling reified constraints of form $\sigma \Rightarrow a_1v_1 + \dots + a_nv_n \leq b$.
Algorithm~\ref{proptrue} is only called if $\sigma\in\Bss$.
\input{propbounds}
Whenever $\sigma$ is true, we check whether the constraint $a_1v_1 + \dots + a_nv_n \leq b$
can be falsified.
If it can never be falsified,
e.g.\ the sum of the current upper bounds already satisfies the constraint in Line~1, we are done.
If we only have propagation strength 1 or 2, we check in Line~3
whether the sum of
the current lower bounds is already above the bound $b$.
In this case, we simply return the current lower bounds of the views
as a nogood,
since the constraint is already violated.
For example,
take the constraint $\sigma \Rightarrow x + y \leq 9$ with $\dom{x}=\dom{y}=\{1,\dots,15\}$
and the current lower and upper bounds
$\mathit{lb}_{\Bss}(x)=7$, 
$\mathit{ub}_{\Bss}(x)=10$, 
$\mathit{lb}_{\Bss}(y)=5$, and
$\mathit{ub}_{\Bss}(y)=12$.
The sum of the lower bounds $7+5$ is greater than $9$,
and so the constraint is violated.
Therefore, we add the nogood $\{\sigma, \tolit{x \geq 7}, \tolit{y \geq 5}\}$.
If the propagation strength is greater than 2 (Lines 6-10),
we try to find new upper bounds for the views of the constraint.
For this purpose,
$cur$ represents the maximal value that $a_iv_i$ can take
without violating the constraint.
All other views $a_jv_j$ ($j\neq i$) contribute at least their
current lower bound to the sum. 
In our example, this means that $cur = 9 - 5 = 4$.
If this value is less than the current upper bound of $a_iv_i$ (Line~8),
we create a nogood that allows us to propagate the new upper bound.
In the example, this is $\{\sigma, \tolit{x > 4}, \tolit{y \geq 5}\}$.
Compared to the nogood that was created in Line~4, this nogood is
stronger as the required minimum for $x$ is lower.
If $cur$ is even below the current lower bound of $a_iv_i$,
we have a conflict and stop eagerly (Line~11).
Since $cur=4$ and $\mathit{lb}_{\Bss}(x)=7$, this is the case in our example.
This algorithm has linear complexity $O(n)$,
but since we consider domains/images with holes, finding the literal $\tolit{a_iv_i > cur}$ is actually
$O(\mathit{log}(|\dom{v_i}|))$ which raises the overall complexity for propagation strength greater than~2.

Algorithm~\ref{propimpl} is only called
if neither $\sigma\in \Bss$ nor $\overline{\sigma}\in \Bss$, e.g.\ whenever $\sigma$ is unknown,
and propagation strength is at least 2 (Line~1).
\input{propreif}
If the sum of all current lower bounds of the left hand side is greater than $b$ (Lines 2 and 3),
the constraint can never become satisfied.
Given a propagation strength below~4,
we simply create a nogood based on the current lower bounds.
In our example, this is the same nogood $\{\sigma, \tolit{x \geq 7}, \tolit{y \geq 5}\}$
generated in Algorithm~\ref{proptrue}.
If the propagation strength is~4 (Lines~8-13),
we try to find a sum of the views that is minimally greater than $b$.
In our example,
we start with a lower bound $low=12$.
By subtracting $\mathit{lb}_{\Bss}(x)$, we get $low'=5$.
This leaves us with $cur=\mathit{next}(9-5,x)=5$ adding $\tolit{x \geq 5}$
to our nogood.
In the second iteration, we now have to find a sufficient lower bound for $y$
that violates the constraint.
We see that this value is 5, adding $\tolit{y \geq 5}$ to $\delta$ in Line~11
resulting in the nogood $\{\sigma, \tolit{x \geq 5}, \tolit{y \geq 5}\}$.
Again,
the complexity of the refined search is higher but also the produced nogoods
are stronger.
Note that as an optimization, \textsc{PropagateBound} and \textsc{PropagateReification}
are only called if the bounds of the variables of the constraints have changed.
The propagation strength is set using the option \commandfont{-{}-prop-strength}.


%% file: translate.tex
\begin{algorithm}[ht]
\Input{A set of signed literals $\delta$ and a linear constraint $a_1v_1 + \dots + a_nv_n \leq b$}
\Output{A set of nogoods}
   $\Sigma \gets \emptyset$\;
   $d \gets next(b-\sum_{j=2}^{n}{\mathit{ub}(a_jv_j)},a_1v_1)$\;
   \While{$d \leq \mathit{ub}(a_1v_1)$}
   {
     \uIf{$d + \sum_{j=2}^{n}{\mathit{lb}(a_jv_j)} \leq b$}
     {
       $\Sigma \gets \Sigma \cup \Translate{$\delta \cup \{\tolit{a_1v_1 \geq d}\}$, $a_2v_2 + \dots + a_nv_n \leq b-d$}$\;
     }
     \uElse
     {
       \Return $\Sigma \cup \{\delta \cup \{\tolit{a_1v_1 \geq d}\}\}$\;
     }
     $d \gets next(d,a_1v_1)$\;
   }
   \Return $\Sigma$
\caption{\textsc{Translate}\label{algo:trans}}
\end{algorithm}%

%% file: cdcl.tex
\begin{algorithm}[ht]
\BlankLine
\Input{A constraint logic program $P$ over $\mathcal{A},\mathcal{C}$ associated with $(\mathcal{V},D,C)$,
a set of constraints atoms $\mathcal{C'}\subseteq\mathcal{C}$, and a set of order atoms $\mathcal{O}_{\Psi(\mathcal{C'})}\subseteq \mathcal{O}$}
\Output{A constraint stable model or \emph{unsatisfiable}}
$\mathcal{B} \gets \mathcal{A}\cup\mathcal{C}\cup\mathcal{O}_{\Psi(\mathcal{C'})}$ \atcp{\textit{set of atoms}}
$\Bss \gets \emptyset$                         \atcp{\textit{assignment over } $\mathcal{A}\cup\mathcal{C}\cup\mathcal{O}$}
$\nabla \gets \emptyset$                       \atcp{\textit{set of (dynamic) nogoods}}
\Loop{}{
  $(\mathcal{B},\Bss,\nabla)\gets \PropagationArg{P,\mathcal{V},D,C}{\mathcal{B},\Bss,\mathcal{C'},\nabla}$ \;
  \uIf{\Conflict{$\Bss$}}
  {
   \lIf{\RootConflict{$\Bss$}}{\Return \emph{unsatisfiable}}
   $(\Bss,\nabla) \gets \BacktrackArg{P}{\Bss,\nabla}$ \;
  }
  \uElseIf{\Complete{$\Bss$}}
  {
    \uIf{$\mathit{lb}_{\Bss}(v)=\mathit{ub}_{\Bss}(v)$ \ for all \ $v\in \mathcal{V}$}
    { \Return $(\Tass{\Bss}\cap \atom{P},\{v \mapsto \mathit{lb}_{\Bss}(v) \mid v\in\mathcal{V}\})$\; 
    }
    \lElse{$\mathcal{B} \gets \mathcal{B} \cup \{\SplitArg{\mathcal{V},D}{\mathcal{B},\Bss}\} $} 
  }
  \lElse{$\Bss \gets \Bss \cup \{\Select{$\mathcal{B}$}\}$}
}
\caption{\textsc{Extended CDCL}\label{cdnl}}
\end{algorithm}%

%% file: propagation.tex
\begin{algorithm}[ht]
\BlankLine
\Global{A constraint logic program $P$ over $\mathcal{A},\mathcal{C}$ associated with $(\mathcal{V},D,C)$}
\Input{A set of atoms $\mathcal{B}$, a Boolean assignment $\Bss$,
 a set of constraint atoms $\mathcal{C'}$, and a set of learned nogoods $\nabla$}
\Output{A set of atoms, a Boolean assignment, and a set of learned nogoods}
\Loop{}{
  $\Bss \gets \UnitPropagationArg{P}{\Bss,\Psi(\mathcal{C'}) \cup \nabla}$\;
  \lIf{$\Conflict{\Bss}$}{\Return $(\mathcal{B},\Bss,\nabla)$}
  $\nabla' \gets \UfsPropagationArg{P}{\Bss}$\;
  \uIf{$\nabla'\neq \emptyset$}
  {
    $\nabla \gets \nabla \cup \nabla'$\;
  }
	\Else{
    $\nabla' \gets \CspPropagationArg{\mathcal{V},D,C}{\mathcal{B},\mathcal{C'},\Bss}$\;
    \uIf{$\nabla'\neq \emptyset$}
    {
		  \lFor{$\delta\in\nabla'$}
			{
		    $\mathcal{B} \gets \mathcal{B} \cup \Tass{\delta}\cup\Fass{\delta}$
			}
      $\nabla \gets \nabla \cup \nabla'$\;
    }
		\lElse
		{
      \Return $(\mathcal{B},\Bss,\nabla)$
		}
	}
}
\caption{\textsc{Propagation}\label{algo:propagate}}
\end{algorithm}%

%% file: cspprop.tex
\begin{algorithm}[ht]
\BlankLine
\Global{A constraint logic program $P$ over $\mathcal{A},\mathcal{C}$ associated with $(\mathcal{V},D,C)$}
\Input{A set of atoms $\mathcal{B}$, a set of constraint atoms $\mathcal{C'}$, and a Boolean assignment $\Bss$}
\Output{A set of nogoods}
$\Sigma \gets \emptyset$ \atcp{\textit{an empty set of nogoods}}
   \For{$v \in \mathcal{V}$}%
	 {
	    \uIf{$\Tlit{(v\leq d)} \in \Bss$ for some $d\in\dom{v}$}
			{
			$ub \gets \min{\{d\mid d\in \dom{v}, \Tlit{(v\leq d)} \in \Bss\}}$\;
			$\Sigma \gets \Sigma \cup \{\{\Tlit{(v\leq ub)}, \Flit{(v\leq x)} \} \mid x>ub, (v\leq x)\in \mathcal{B}, \Tlit{(v\leq x)}\notin \Bss\}$
			}
      \uIf{$\Flit{(v\leq d)} \in \Bss$ for some $d\in\dom{v}$}
			{
			$lb \gets \max{\{d\mid d\in \dom{v}, \Flit{(v\leq d)} \in \Bss\}}$\;
			$\Sigma \gets \Sigma \cup \{\{\Tlit{(v\leq x)}, \Flit{(v\leq lb)} \} \mid x<lb, (v\leq x)\in \mathcal{B}, \Flit{(v\leq x)}\notin \Bss\}$
			}

	 }
	 \lIf{$\Sigma \neq \emptyset$} {\Return $\Sigma$}
	 \For{$c\in \mathcal{C}\setminus\mathcal{C'}$}%
	 {
	 \uIf{$\Tlit{c}\in\Bss$}
	 {
	 $\Sigma \gets \Sigma \cup \textsc{PropagateBounds}(\Bss,\Tlit{c}\Rightarrow \gamma(c))$
	 }
	 \uElseIf{$\Flit{c}\in\Bss$}
	 {
	 $\Sigma \gets \Sigma \cup \textsc{PropagateBounds}(\Bss,\Flit{c}\Rightarrow \overline{\gamma(c)})$
	 }
	 \uElse
	 {
	 $\Sigma \gets \Sigma \cup \textsc{PropagateReification}(\Bss,\Tlit{c}\Rightarrow \gamma(c))$\;
	 $\Sigma \gets \Sigma \cup \textsc{PropagateReification}(\Bss,\Flit{c}\Rightarrow \overline{\gamma(c)})$\;
	 }
	 \lIf{$\Sigma \neq \emptyset$} {\Return $\Sigma$}
	 }
	 \Return $\emptyset$
\caption{\textsc{CspPropagation}\label{algo:cspprop}}
\end{algorithm}%

%% file: propbounds.tex
\begin{algorithm}[ht]
\BlankLine
\Global{An integer $\mathit{ps}$}
\Input{A Boolean assignment \Bss{} and a half-reified constraint $\sigma \Rightarrow a_1v_1 + \dots + a_nv_n \leq b$}
\Output{A set of nogoods}
$\Sigma \gets \emptyset$ \atcp{An empty set of nogoods}
   \lIf{$\sum_{j=1}^{n}{\mathit{ub}_{\Bss}(a_j v_j)} \leq b$}{\Return $\emptyset$}
   \uIf{$\mathit{ps}\leq2$}
   {
     \uIf{$\sum_{j=1}^{n}{\mathit{lb}_{\Bss}(a_j v_j)} > b$}
     {
       $\Sigma \gets \{\{\sigma\} \cup \{\tolit{a_jv_j \geq lb_{\Bss}(a_jv_j)}\mid 1\leq j \leq n\}\}$\;
     }
     \Return $\Sigma$
   }
   \For{$i = 1..n$}
   {
     $cur \gets b - \sum_{j=1,j\neq i}^{n}{\mathit{lb}_{\Bss}(a_j v_j)}$\;
     \uIf{$cur < \mathit{ub}_{\Bss}(a_iv_i)$}
     {
       $\Sigma \gets \Sigma \cup \{\{\sigma,\tolit{a_iv_i> cur}\} \cup \{\tolit{a_jv_j \geq \mathit{lb}_{\Bss}(a_jv_j)}\mid 1 \leq j \leq n, j\neq i\}\}$\hspace{-1cm}\;
       \lIf{$cur < \mathit{lb}_{\Bss}(a_iv_i)$}{ \Return $\Sigma$}
     }
   }
	 \Return $\Sigma$
\caption{\textsc{PropagateBounds}\label{proptrue}}
\end{algorithm}%

%% file: propreif.tex
\begin{algorithm}[ht]
\BlankLine
\Global{An integer $\mathit{ps}$}
\Input{A Boolean assignment \Bss\ and a half-reified constraint $\sigma\Rightarrow a_1v_1 + \dots + a_nv_n \leq b$}
\Output{A set of nogoods}
   \lIf{$\mathit{ps}=1$}{ \Return $\emptyset$}
   $low \gets \sum_{j=1}^{n}{lb_{\Bss}(a_j v_j)}$\;
   \uIf{$low > b$}
   {
     $\delta \gets \{\sigma\}$\;
     \uIf{$\mathit{ps}<4$}
     {
       $\delta \gets \delta \cup \{\tolit{a_jv_j \geq lb_{\Bss}(a_jv_j)}\mid 1 \leq j \leq n\})$\;
     }
     \uElse
     {
       \For{$j \in 1..n$}
       {
         $low' \gets low - lb_{\Bss}(a_jv_j)$\;
         $cur \gets next(b - low',a_jv_j)$\;
         $\delta \gets \delta \cup \{\tolit{a_jv_j \geq cur}\}$\;
         $low \gets low' + cur$\;
       }
     }
     \Return $\{\delta\}$
   }
\caption{\textsc{PropagateReification}\label{propimpl}}
\end{algorithm}%

%% file: features.tex
\subsection{Distinguished Features}\label{sec:features}

After presenting the algorithmic framework of \clingcon~3,
we now describe some of its specific features.
Many of them aim at reducing the size of domains and the number of variables,
while others address special functionalities, like global constraints or multi-objective optimization over integer variables, respectively.
%

When we refer in the following to the truth value of atoms,
we consider a partial assignment obtained by propagation and/or preprocessing.

\paragraph{Views.}
A view $av+b$ can be represented with the same set of order atoms
as its variable $v$~\cite{thistu09a}.
Consider the view $-5v + 7$ together with the domain $\dom{v}=\{1,2,3,4,5\}$.
We show how the order atoms of $v$ are used to encode constraints over the view in \clingcon.
The view $-5v + 7$ has the following values in its image: $\img{-5v + 7}=\{-18,-13,-8,-3,2\}$.
The order literals for $\{\tolit{v\leq x} \mid x\in\dom{v}\}$ and
$\{\tolit{-5v+7 \leq x} \mid x\in\img{-5v+7}\}$
are given in Table~\ref{tab:view}.
\begin{table}[ht]
\begin{tabular}{l|r|ccccc}
Expression & Image & \multicolumn{3}{l}{Order Literals} \\\cline{1-7}
$\phantom{-5}v$ & $\{1,2,3,4,5\}$  & $\Tlit{v\leq1}$ & $\Tlit{v\leq2}$ & $\Tlit{v\leq3}$ & $\Tlit{v\leq4}$ & \Tlit{\emptyset}\\
$-5v + 7$ & $\{-18,-13,-8,-3,2\}$ & $\Flit{v\leq4}$& $\Flit{v\leq3}$& $\Flit{v\leq2}$& $\Flit{v\leq1}$ & \Tlit{\emptyset}
\end{tabular}
\caption{Order literals of different views of one variable.}
\label{tab:view}
\end{table}
We see that the set of order atoms used for these literals is the same.
By allowing views instead of variables,
we avoid introducing new variables (for views).
In fact, neither the XCSP~\cite{roulec09a} nor the
\flatzinc\footnote{http://www.minizinc.org/downloads/doc-1.3/flatzinc-spec.pdf}
format allow for using views in global constraints.
For instance,
a distinct constraint over the set the views
$\{1000v_1, 1000v_2, 1000v_3, 1000v_4, 1000v_5\}$
translates into the same nogoods as a
distinct constraint over $\{v_1, v_2, v_3, v_4, v_5\}$.
Due to the restriction to use variables,
according solvers like
\sugar~\cite{tatakiba09a} introduce auxiliary variables
 $v'_i=1000v_i$ for $1 \leq i \leq 5$.
If $\dom{v_i}=\{1,\dots,10\}$,
bound propagation yields the domains $\dom{v'_i} = \{1000\cdot1, \dots, 1000\cdot10\} = \{1000,\dots,10000\}$.%
\footnote{As done in the \sugar{} system.}
Furthermore,
around 220000 nogoods for the equality constraints are created.
By handling views directly,
we avoid introducing these auxiliary variables and constraints in \clingcon.

The same holds for minimize statements.
Views on variables such as $3*v_2$ or $-v_3$
allow for weighting variables during minimization as well as maximization,
without the need of introducing auxiliary variables and additional constraints.

\paragraph{Non-Contiguous Integer Domains.}
We represent domains of variables (and images of views) as sorted lists of 
ranges like $[1..3, 7..12, 39..42]$ instead of single ranges like $[1..42]$.
This has the advantage that we can represent domains with holes directly,
without any additional constraints.
Introducing order atoms for such a non-contiguous domain
produces fewer atoms (12 in this example) than for a domain only represented with two bounds (41).
A drawback of this representation is
that the lookup for a certain value $d$ in the domain becomes logarithmic,
as we rely upon binary search in the list of ranges.
This is frequently done in Algorithms~\ref{algo:trans},~\ref{proptrue} and~\ref{propimpl} 
whenever a calculated value $d$ leads to searching for a literal
$\tolit{v \leq d}$.

\paragraph{Equality Processing.}
To minimize the number of atoms and nogoods that have to be created during
a translation or solving process,
we need to reduce the number of integer variables.
To accomplish this, 
we consider the equalities in a CSP that include only two integer variables,
and replace all occurrences of the first variable with a view
on the second variable in all other constraints.
Consider a constraint logic program $P$ over $\mathcal{A}$, $\mathcal{C}$
associated with $(\mathcal{V},D,C)$.
For each element $\gamma(\sigma) \in C$ of the form $ax+c_1 = by+c_2$ (or $ax+c_1 \neq by+c_2$)
where $\sigma$ is true (false), 
$a,b,c_1,c_2$ are integers, and $x,y \in \mathcal{V}$,
we successively replace constraints in $C$.
For this, we normalize the constraint $\gamma(\sigma)$ to $ax = by+c$ 
where $x$ is lexicographically smaller than $y$ and
multiply all constraints in $C$ containing variable $y$ with $b$
and replace $by+c$ by $ax$ in them.
The domain of $x$ is made domain consistent
such that $ad \in \img{by+c}$ holds for all $d\in\dom{x}$.
Afterwards,
we remove $\gamma(\sigma)$ from $C$ and $y$ from $\mathcal{V}$.
Note that by replacing variables,
new equalities may arise,
which we process until a fixpoint is reached.

For illustration,
consider the following set $C$ of constraints.
\begin{align}
a = 2b\label{eqp:a}\\
b = 2c\label{eqp:b}\\
c = 2d\label{eqp:c}\\
d = 2e\label{eqp:d}\\
e = 2f\label{eqp:e}\\
a + 14d -3f + b \leq -g\label{eqp:g}
\end{align}
And assume that the constraint literals associated with the first 5 constraints are true.
Furthermore, let $D= \{\dom{x} = \{-2^{12},\dots,2^{12} \mid x \in \{a,b,c,d,e,f,g\}\}$.
Without any simplification, we have 7 variables,
all with a domain size of roughly 8000.
By simply translating these constraints,
we would create around 120000 order atoms
and 118 million nogoods.
Let us show how equality processing allows us to significantly reduce these numbers in our example.
To begin with,
we multiply the constraint in~\eqref{eqp:g}, viz.\ $a + 14d -3f + b \leq -g$, with 2
and replace $-6f$ with $-3e$ using the constraint in~\eqref{eqp:e}.
This yields $2a + 28d -3e + 2b \leq -2g$.
Also, \eqref{eqp:e} allows us to restrict the domain of $e$ to $\dom{e}=\{-2^{11},\dots,2^{11}\}$.
We then remove $e = 2f$ from the set of constraints and $f$ from the set of variables.
We repeat this procedure for all other equalities.
To replace $e$,
we again multiply the obtained constraint by 2, yielding $4a + 56d -6e + 4b \leq -4g$,
and replace $6e$ with $3d$ using~\eqref{eqp:d}.
This results in $4a + 53d + 4b \leq -4g$.
Again, we remove $d = 2e$ and variable $e$,
and obtain $\dom{d}=\{-2^{10},\dots,2^{10}\}$.
Using~\eqref{eqp:c}, we multiply by 2 and replace $106d$
with $53c$ which leads to the constraint $8a + 53c + 8b \leq -8g$.
To remove $c$, the constraint in~\eqref{eqp:b} is used to replace
$106c$ with $53b$ resulting in  $16a + 69b \leq -16g$.
In the last step,
we apply~\eqref{eqp:a} to get $32a + 69a \leq -32g$ which simplifies to $101a \leq -32g$.
As a result,
the overall set of constraints is thus reduced to a single constraint $101a \leq -32g$.
This constraint uses only two variables
with domains $\dom{a}=\{-2^{7},\dots,2^{7}\}$ and $\dom{g}=\{-2^{12},\dots,2^{12}\}$.
All other constraints and variables have been removed.
To translate this constraint, we need 9265 order atoms and 268 nogoods.

Our approach to equivalence processing is inspired by Boolean \emph{Equi-propagation}~\cite{mecost13a},
which directly replaces the order atoms of one
variable with the other.
Directly using integer variables,
without considering the order literal representation,
allows us to use this technique also in the context of lazy variable generation.
Here, it reduces the number of variables,
which leads to shorter constraints,
which ultimately reduces the number of nogoods in the translation process.

Equality preprocessing is done once in \clingcon, before the actual solving starts
and can be controlled using the 
command line option \commandfont{-{}-equality-processing}.

\paragraph{Distinct Translation.}
\newcommand{\distinctatom}{\ensuremath{\mathit{c}}}
\newcommand{\ndistinctatom}{\ensuremath{\mathit{c'}}}
\clingcon{} features two alternatives for translating global distinct constraints.
Assume that constraint atom \distinctatom\ represents a distinct
constraint
over a set $\{v_1,\dots,v_n\}$.
Since we represent distinct constraints in terms of rules and other linear constraints,
this constraint atom becomes a regular atom and is used in the head of rules.

The first method to handle this constraint uses a quadratic number of new, regular atoms $neq(v_i,v_j)$
for all $1 \leq i < j \leq n$
together with the rules
\begin{align*}
neq(v_i,v_j) \leftarrow (v_i - v_j \leq 1)\\
neq(v_i,v_j) \leftarrow (v_j - v_i \leq 1)
\end{align*}
to represent that two variables are unequal.
By adding the following rule to the program
\begin{align*}
\distinctatom \leftarrow neq(v_1,v_2), neq(v_1,v_3), \dots, neq(v_1,v_n),\\
                                                          \phantom{neq(v_1,v_2), }neq(v_2,v_3), \dots, neq(v_2,v_n),\\
                                                          \ddots \hspace{0.8cm} \vdots\phantom{v_n,}\hspace{0.2cm}\\
                                                          \phantom{neq(v_1,v_2), neq(v_1,v_3), \dots,} neq(v_{n-1},v_n)\phantom{,}
\end{align*}
\clingcon\ ensures that $c$ is only true if all variables are distinct from each other.
%

The second alternative uses a so-called \emph{direct encoding}~\cite{walsh00a}.
For each value ${d\in\bigcup_{i=1}^{n}{\img{v_i}}}$,
we ensure that at most one variable from $\{v_1,\dots,v_n\}$ takes this value.
Therefore,
we introduce regular atoms of form $eq(v_i,d)$ for all these variables together with the rule
\begin{align}
\label{directencodingatoms}
eq(v_i,d) \leftarrow (v_i \leq d), (-v_i \leq -d)
\end{align}
representing that $v_i=d$.
Furthermore,
we add a cardinality constraint~\cite{siniso02a}
for each value $d$
to the effect that no two or more variables may have the same value,
viz.\
\begin{align*}
\ndistinctatom \leftarrow {2\;\{eq(v_1,d),\dots,eq(v_n,d)\}}
\end{align*}
The new regular atom \ndistinctatom\ is true
if two or more variables have the same value $d$.
If this is not the case, the distinct constraint atom holds via the rule:
\begin{align*}
\distinctatom \leftarrow \naf{\ndistinctatom}
\end{align*}
We reuse the direct encoding atoms $eq(v_i,d)$ for other distinct constraints.
Note that introducing all direct encoding atoms also
involves the creation of corresponding order atoms
before the solving process.
So no variable from a distinct constraint can be created lazily.
This is also the reason why this option is not enabled in \clingcon{} by default
and distinct constraints are translated using inequalities.
The use of the direct encoding along with cardinality constraints
is enabled with the option \commandfont{--distinct-to-card}.

\paragraph{Pigeon Hole Constraints.}
To enhance the propagation strength when translating distinct constraints in \clingcon,
we add rules for the lower and upper bounds.
Consider the constraint atom \distinctatom\  
for a distinct constraint over $\{v_1,\dots,v_n\}$ and let $U = \bigcup_{i=0}^{n}{\img{v_i}}$,
$l$ be the $n$th smallest element in $U$,
and $u$ be the $n$th greatest element in $U$.
We add the rules:
\begin{align*}
&\leftarrow \distinctatom, (v_1 > u), \dots, (v_n > u)\\
&\leftarrow \distinctatom, (v_1 < l), \dots, (v_n < l)
\end{align*}
where as before, \distinctatom\ is treated as regular atom.

So given a distinct constraint over $\{v_1,v_2,v_3\}$
with $\dom{v_i}=\{1,\dots,10\}$ for $1\leq i \leq 3$
we add the rules
\begin{align*}
\leftarrow \distinctatom, (v_1 > 8), (v_2 > 8), (v_3 > 8)\\
\leftarrow \distinctatom, (v_1 < 3), (v_2 < 3), (v_3 < 3)
\end{align*}
This forbids all variables to have a value greater than eight or to have a value less than three.
This feature only causes a constant overhead in the number of rules.
It can be controlled using the option \commandfont{-{}-distinct-pigeon}.

\paragraph{Permutation Constraints.}
A distinct constraint over $\{v_1,\dots,v_n\}$ where $U=\bigcup_{i=1}^{n}{\img{v_i}}$ and $|U|=n$
induces a permutation on the variables.
Let \distinctatom{} be the constraint atom representing this global constraint.
In this special case,
we can add the rules
\begin{align*}
\leftarrow \distinctatom, \naf{eq(v_1,d)}, \dots, \naf{eq(v_n,d)} \qquad \text{ for all } d\in U.
\end{align*}
These rules enforce that each value is taken at least once.

For example,
given a
distinct constraint over $\{v_1,v_2,v_3\}$
with $\dom{v_i}=\{1,\dots,3\}$ for $1\leq i \leq 3$
we add the rules
\begin{align*}
\leftarrow \distinctatom, \naf{eq(v_1,1)}, \naf{eq(v_2,1)}, \naf{eq(v_3,1)}\\
\leftarrow \distinctatom, \naf{eq(v_1,2)}, \naf{eq(v_2,2)}, \naf{eq(v_3,2)}\\
\leftarrow \distinctatom, \naf{eq(v_1,3)}, \naf{eq(v_2,3)}, \naf{eq(v_3,3)}\\
\end{align*}
This feature introduces direct encoding atoms along with the respective rules and order atoms in~\eqref{directencodingatoms}.
Since these atoms cannot be treated lazily,
this feature is disabled by default but can be controlled
using the option \commandfont{-{}-distinct-permutation}.

\paragraph{Sorting.}
Sorting constraints by descending coefficients
is known to avoid redundant nogoods in the translation process~\cite{basota13a}.
Also, systems like \sugar{} sort
constraints by smallest domain first,
and when tied, with largest coefficient.
\clingcon\ can either sort by coefficient or domain size first,
in decreasing or increasing order.
The option \commandfont{-{}-sort-coefficient} 
controls the sorting of the constraints.

\paragraph{Splitting Constraints.}
Considering that directly translating a linear constraint 
$a_1v_1 + \dots + a_nv_n \leq b$ with the order encoding
leads to an exponential number of nogoods,
we split long constraints into shorter ones
by introducing new variables.
Thereby we adapt the heuristics of \sugar.
We only split a constraint if the number of variables is greater than $\alpha$
and if its translation produces more than $\beta$ nogoods.
If both conditions hold,
we recursively split a constraint into $\alpha$ parts.
The new constraints have the form $a_k v_k + \dots + a_l v_l =  v^k_l$
where $1 \leq k \leq l \leq n$.
$\alpha$ and $\beta$ are freely configurable.
By default,
splitting is disabled in \clingcon,
but $\alpha$ and $\beta$ can be changed with options \commandfont{-{}-split-size}
and \commandfont{-{}-max-nogoods-size}.

\paragraph{Symmetry Breaking.}
When splitting a constraint like $a_1v_1 + a_2v_2 + a_3v_3 \leq b$,
we get the constraints $a_1v_1 + a_1v_2 = v^1_2$
and $v^1_2 + a_3v_3 \leq b$.
Equations like $a_1v_1 + a_1v_2 = v^1_2$ are represented as conjunctions of
$a_1v_1 + a_1v_2 \leq v^1_2$ and $a_1v_1 + a_1v_2 \geq v^1_2$.
By dropping the latter inequality,
we obtain an equi-satisfiable set of constraints
being smaller than before but admitting more (symmetric) solutions,
as $v^1_2$ freely varies.
Symmetry breaking should therefore be enabled if one wants to enumerate
all solutions without duplicates.
This form of symmetry breaking is usually skipped in SAT-based CSP solvers like \sugar.
This option is set via \commandfont{-{}-break-symmetries}.

\paragraph{Domain Propagation.}
To create the domain of variables like $v^1_n$ in the aforementioned constraints of form
$a_1v_1 + \dots + a_nv_n = v^1_n$,
we may use bound propagation.
For example, the constraint
$42x + 1337z = y$
where $\dom{x} = \dom{z} = \{0,1\}$
results in the domain
$\dom{y} = \{42\cdot\mathit{lb}(x) + 1137\cdot\mathit{lb}(z), \dots, 42\cdot\mathit{ub}(x) + 1137\cdot\mathit{ub}(z)\} = \{0,\dots,1379\}$.
Using domain propagation instead
leads to the much smaller domain
$\dom{y} = \{42d_x +  1137d_z \mid d_x \in \dom{x}, d_z\in\dom{z}\} = \{0,42,1337,1379\}$.
However, we restrict domain propagation to preprocessing by default,
as it has an exponential runtime.
\clingcon{} allows for controlling domain propagation by setting a threshold on the domain size;
this is set by option \commandfont{-{}-domain-size}.

\paragraph{Translate Constraints.}
Following a two-fold approach,
\clingcon{} is able to translate some constraints
while leaving others to constraint propagators
as shown in Section~\ref{sec:algorithms}.
%
\clingcon{} provides the option \commandfont{-{}-translate-constraints=m} to decide which constraints 
to translate or not.
The translation depends on the estimated number of nogoods $\prod_{i=1}^{n-1}\left|\dom{v_i}\right|$ 
that Algorithm~\ref{algo:trans} produces
for a constraint $a_1v_1 + \dots + a_nv_n \leq b$.
If this number is below the threshold \commandfont{m},
\clingcon{} translates the constraint.
Also all order atoms used in these nogoods are created.

\paragraph{Redundant Nogood Check.}
A nogood $\delta$ is said to be stronger than a nogood $\delta'$,
 iff for all literals $\tolit{v> d}\in\delta$,
there exists a literal $\tolit{v> d'}\in\delta'$ such that $d \leq d'$ and $v$ is a view.
Whenever a nogood is created in Line 7 in Algorithm~\ref{algo:trans},
we compare it to the previously created one.
If one of them is stronger,
we only keep the stronger one,
otherwise, we keep both.
This feature allows \clingcon{} to remove some redundant nogoods
 during the translation process.
It is especially useful if the constraints are not sorted by descending coefficients.
The check just adds constant overhead to the translation process
but avoids creating a significant amount of nogoods.
For instance,
translating the famous \emph{send more money} problem
results in 628 nogoods
among which 327 are redundant,
when using \commandfont{-{}-split-size=3}.
This feature can be triggered using option \commandfont{-{}-redundant-nogood-check}.

\paragraph{Don't Care Propagation.}
Suppose we want to express that  $(x>7)$ should hold whenever $a$ holds;
otherwise we do not care whether $(x>7)$ holds or not.
A corresponding constraint logic program is given in the first row of Table~\ref{table:strict}
together with its constraint stable models.
\begin{table}
\begin{tabular}{p{2.5cm}|ll}
logic program $P$& \multicolumn{2}{c}{constraint stable models of $P$}\\
\hline
$\{a\}$                       & $\{(\{a,(x>7)\}$, & $\{x\mapsto d\}) \mid d\in\{8,\dots,10\}\} \cup $\\
$\leftarrow a, \naf{(x>7)}$   & $\{(\{(x>7)\}$, & $\{x\mapsto d\}) \mid d\in\{8,\dots,10\}\} \cup $\\
                              & $\{(\emptyset$, & $\{x\mapsto d\}) \mid d\in\{1,\dots,\phantom{1}7\}\} $\\
\hline
$\{a\}$                       & $\{(\{a,(x>7)'\}$, & $\{x\mapsto d\}) \mid d\in\{8,\dots,10\}\} \cup$\\
$\leftarrow a, \naf{(x>7)'}$  & $\{(\emptyset$, & $\{x\mapsto d\}) \mid d\in\{1,\dots,10\}\} $\\
$\leftarrow \naf{a}, (x>7)'$    &                              &                                       
\end{tabular}
\caption{Constraint logic programs using reified $\Tlit{(x>7)}\Leftrightarrow x>7$ and half-reified \mbox{$\Tlit{(x>7)'} \Rightarrow x>7$} constraints.}
\label{table:strict}
\end{table}
In the standard case for CASP, the constraint atom is reified with its constraint
via $\Tlit{(x>7) \Leftrightarrow x > 7}$.
In the case that $a$ is true, the constraint atom $(x>7)$ has to be true.
The reification ensures that $x$ is greater than 7,
leading to three different assignments $\{\{x\mapsto d\} \mid d\in\{8,\dots,10\}\}$ for variable $x$.
In the case that $a$ is false,
the constraint atom $(x>7)$ can either be true or false.
The first case results in the same three assignments,
while the latter corresponds to seven others, viz.\ $\{\{x\mapsto d\} \mid d\in\{1,\dots,7\}\}$,
as the reification imposes that the constraint $x>7$ does not hold,
basically enforcing $x\leq7$.
We note that in case $a$ is false,
the constraint imposed on $x$ is either $x>7$ or $x\leq 7$.
This means that there is actually no restriction on the assignment of $x$.
We exploit this observation
by replacing $(x>7)$ with a new constraint atom $(x>7)'$
and adding the rule $\leftarrow \naf{a}, (x>7)'$.
The idea is that atom $(x>7)'$ imposes $(x>7)$ as a half-reified constraint,
meaning that $x$ is enforced to be greater than 7 only if the constraint atom $(x>7)'$ is true,
i.e. $\Tlit{(x>7)'}\Rightarrow x>7$.
We obtain exactly the same stable models in terms of the regular atoms and integer variable assignments,
as depicted in the second row of Table~\ref{table:strict}.
The difference between these two programs lies in the assignment of the constraint atoms.
The additional rule $\leftarrow \naf{a}, (x>7)'$ ensures that the constraint atom $(x>7)'$ is false, whenever $a$ is false.
Since we connect the constraint atom with its constraint using a half-reified constraint,
this constraint has no effect on the assignment of $x$,
resulting in $\{\{x\mapsto d\} \mid d\in\{1,\dots,10\}\}$.
Although the number of constraint stable models stays the same, the number of different Boolean assignments is reduced.

This technique is called \emph{Don't Care Propagation}~\cite{thbawa04a}.
All constraint atoms that only occur in integrity constraints and only positively (negatively)
in the whole program are don't care atoms.
\clingcon{} fixes the truth value of don't care atoms to false (true),
if all integrity constraints containing the atom have at least one literal being false under the current assignment.
Don't care propagation can be useful in SAT,
but it has even more potential to be helpful in CASP/SMT,
since we not only reduce the search space
but also the theory propagator has to handle only one half-reified constraint per don't care atom.
This means only half of the inferences have to be checked.
This technique is not specifically designed for CSP but it can
also be used for other theories.
Don't care propagation is controlled using the option \commandfont{-{}-dont-care-propagation}.

\paragraph{Order Atom Generation.} 
When translating a constraint,
all order atoms
for all its integer variables must be available.
By not translating all constraints,
we also do not need to create all order atoms. 
Some of them can be created on the fly during propagation.
With this in mind, it might still be useful to create a certain number
of order atoms per variable in a preprocessing step.
\clingcon{} can create $n$ atoms evenly spread among the domain values of a variable $v$.
So if we have a domain $\dom{v}=\{1,\dots,10, 90,\dots,100\}$ and create four order atoms
we use $(v\leq 3), (v\leq 8), (v\leq 92)$ and $(v\leq 97)$.
These order atoms allow the solver to split the domain
during the search.
Option \commandfont{-{}-min-lits-per-var=n} adds at least $min(n,|\dom{v}|-1)$ order atoms for each variable $v$.

\paragraph{Explicit Binary Order Nogoods.}
Some order atoms are created before solving.
Therefore,
it can also be beneficial to create a subset of the order nogoods $\Phi'(\mathcal{V},D)$
in advance, as shown in Corollary~\ref{cor:partial}.
Given that we created the set of order atoms $\{(v\leq x_1), \dots, (v\leq x_n)\}$
for a variable $v\in\mathcal{V}$ where $x_i < x_{i+1}$ for $1 \leq i \leq n$,
the explicit order nogoods
\[
\{\{\Tlit{v\leq x_1}, \Flit{v\leq x_2}\}, \dots, \{\Tlit{v\leq x_{n-1}}, \Flit{v\leq x_n}\}\}
\]
can also be created.
To introduce these binary order nogoods for all order atoms that have been created 
before the solving process, the option \commandfont{-{}-explicit-binary-order} can be used.

\paragraph{Objective Functions.}
We support multi-objective optimization
on sets of views.
For all views $av + c$ subject to minimization,
we use the signed order literals $\tolit{av + c \geq d}$
with weight
\[
\begin{cases} 
      d - prev(d,av + c) & \text{if }d > \mathit{lb}(av + c) \\
      d  & \text{if } d = \mathit{lb}(av + c) 
\end{cases}
\]
for all values $d \in \img{av + c}$ in an ASP minimize statement.
This minimizes the total sum of the set of views.
By using native ASP minimize statements, \clingcon{}
reuses \clasp's branch and bound
and unsatisfiable core based techniques~\cite{ankamasc12a}.
For instance,
for minimizing $3x$ where $\dom{x} = \{1,3,7\}$,
we have the following weighted literals in the (internal) ASP minimize statement
$\tolit{3x\geq 3}=3$, $\tolit{3x\geq 9}=6$, and $\tolit{3x\geq21}=12$.
In terms of ASP-pseudo-code this amounts to a minimize statement of form
\(
\#minimize\{6 : \naf{x\leq 1}; 12 : \naf{x\leq 3}\}
\)
although order literals are not part of the input language.
$\tolit{3x\geq 3}$ evaluates to true,
while $\tolit{3x\geq 9}$ and $\tolit{3x\geq21}$ can be expressed via order literals as $\naf{(x\leq 1)}$ and $\naf{(x\leq 3)}$, respectively.
%

\paragraph{Flattening Objective Functions.}
Minimizing the value of an integer variable $y$
that is included in a constraint $\gamma(\sigma)=a_1v_1 + \dots + a_nv_n = y$
where $\sigma$ is true, is equivalent
to minimizing the value of $a_1v_1 +  \dots + a_nv_n$.
Directly using the views $a_iv_i$ strengthens the nogoods used to represent
the minimize statement.
The constraint $a_1v_1 + \dots + a_nv_n = y$
can be removed
if $y$ is not used anywhere else.\footnote{We keep the constraint to be able to correctly print $y$ in a solution.}
In fact, this pattern occurs quite often in our \minizinc{} benchmark set.
Replacing variable $y$ with its constituents $a_1v_1 +  \dots + a_nv_n$ can be controlled with the option \commandfont{-{}-flatten-optimization}.

\paragraph{Reduced Nogood Learning.}
Whenever \CspPropagation\ in Algorithm~\ref{algo:propagate} and~\ref{algo:cspprop} derives a nogood, it is possible
to not add it to the store of learned nogoods $\nabla$ but rather
keep it implicit and only add it if it is really needed
for conflict analysis.
The internal interface of \clasp{} supports such a behavior.
While the learned nogoods $\nabla$ improve the
strength of unit propagation,
too many nogoods decrease its performance.
Therefore, lazily adding these nogoods when they are actually needed
can improve unit propagation.
To disable the storage of nogoods and handle them implicitly,
\clingcon{} provides option \commandfont{-{}-learn-nogoods}.


%% file: multishot.tex
\subsection{Multi-Shot CASP Solving}\label{sec:multi:shot}

As mentioned,
a major design objective of \clingcon~3 is to transfer \clingo's functionalities to CASP solving.
A central role in this is played by multi-shot solving~\cite{gekakasc14b,gekaobsc15a} because it allows for casting manifold reasoning modes.
More precisely,
multi-shot solving is about solving continuously changing logic programs in an operative way.
This can be controlled via reactive procedures that loop on solving while reacting, for instance, to outside changes or previous solving results.
These reactions may entail the addition or retraction of rules that the operative approach can accommodate by leaving the unaffected program parts
intact within the solver.
This avoids re-grounding and benefits from heuristic scores and nogoods learned over time.

To extend multi-shot solving to CASP,
our propagators allow for adding and deleting constraints in order to capture evolving CSPs.
Evolving constraint logic programs can be extremely useful in dynamic applications, for example, to:
\begin{itemize}
\item add new resources in a planning domain, 
\item set the value of an observed variable measured using sensors,
\item add restrictions to reduce the capacity of containers, or
\item increase their capacity depending on other systems like weather forecast etc\@.
\end{itemize}
The presented propagators provide means for all these issues.
New resources can be added using additional constraint variables and domains.
Values can be limited by adding constraints and rules to the constraint logic program.
Due to our monotone treatment of CSPs in CASP,
it is always possible to add new constraint atoms.
Since they are not allowed to occur in rule heads they to not interfere with the completion of the logic program.
Hence, we can combine (and therefore extend) two constraint logic programs under exactly the same restrictions
that apply to normal logic programs (cf.~\cite{gekakasc14b}).

While confining variables is easy, accomplished by adding constraints on those variables,
increasing their capacity is addressed via lazy variable generation.
That is, we start with a virtually maximum domain that is restrained by retractable constraints.
The domain is then increased by relaxing these constraints.
Importantly, the order atoms representing the active domain are only generated when needed.
This avoids introducing a large amount of atoms,
especially in the non-active area of the domain.
As an example,
consider the variable $x$ and its domain $\dom{x}=\{1,\dots,10^9\}$ having
one billion elements.
By adding the constraint $x\leq10$,
only the first 10 values are valid assignments.
After retracting $x\leq10$ and adding $x\leq20$,
only the first 20 values constitute the search space.
Since order atoms are only introduced in the actual search space,
no atoms are introduced for the huge amount ($10^9 - 20$) of other values.
Using this technique,
CASP can deal with increasing domains within reasonable space.

For illustration,
let us consider the well-known $n$-queens puzzle for demonstrating
how to incrementally add new constraints and constraint variables to a constraint logic program
and how to remove constraints from it.
To illustrate how seamlessly \clingcon\ integrates CASP and multi-shot solving,
we apply \clingo's exemplary \python\ script for incremental solving
to model different incremental versions of the $n$-queens puzzle in CASP.
Multi-shot solving in \clingo\ relies on two directives~\cite{gekakasc14b}, 
the \texttt{\#program} directive for regrouping rules
and 
the \texttt{\#external} directive for declaring atoms
as being external to the program at hand.
The truth value of such external atoms is set via \clingo's API\@.
\sysfont{Clingo}'s incremental solving procedure is provided in \python\ and loops over increasing integers until a stop criterion is met.
It presupposes three groups of rules declared via \texttt{\#program} directives.
At step 0 the programs named \texttt{base} and \texttt{check(n)} are ground and solved for $\texttt{n}=0$.
Then, in turn programs \texttt{check(n)} and \texttt{step(n)} are added for $\texttt{n}>0$ and the obtained program is grounded and solved.
Other names and components are definable by appropriate changes to the \python\ program.
Stop criteria can be the satisfiability or unsatisfiability of the respective program at each iteration.
In addition, at each step $\texttt{n}$ an external atom \texttt{query(n)} is introduced;
it is set to true for the current iteration $\texttt{n}$ and false for all previous instances with smaller integers than $\texttt{n}$.
Although we reproduce the exemplary \python\ program from \clingo's example pool in~Listing~\ref{encoding:incmode},
we must refer the reader to~\cite{gekakasc14b} for further details.
\begin{lstlisting}[float=ht,caption={Incremental mode of \emph{Clingo}},label=encoding:incmode,basicstyle=\ttfamily\scriptsize]
#script (python)

import clingo
    
def get(val, default):
    return val if val != None else default
        
def main(prg):
    imin   = get(prg.get_const("imin"), clingo.Number(0))
    imax   = prg.get_const("imax")
    istop  = get(prg.get_const("istop"), clingo.String("SAT"))
    
    step, ret = 0, None
    while ((imax is None or step < imax.number) and
           (step == 0 or step < imin.number or (
              (istop.string == "SAT"     and not ret.satisfiable) or
              (istop.string == "UNSAT"   and not ret.unsatisfiable) or
              (istop.string == "UNKNOWN" and not ret.unknown)))):
        parts = []
        parts.append(("check", [step]))  
        if step > 0:
            prg.release_external(clingo.Function("query", [step-1]))
            parts.append(("step", [step]))
            prg.cleanup()
        else:
            parts.append(("base", []))
        prg.ground(parts)
        prg.assign_external(clingo.Function("query", [step]), True)
        ret, step = prg.solve(), step+1
#end.   
        
#program check(t).
#external query(t). 
\end{lstlisting}

The CASP encoding of the incremental $n$-queens puzzle in Listing~\ref{encoding:multishot1}
demonstrates the addition and removal of constraints
and also shows how variable domains are dynamically increased. 
\begin{lstlisting}[float,caption={Incremental $n$-queens encoding $Q_1$ (\texttt{incqueens.lp})},label=encoding:multishot1]
#include "incmode.lp".
#include "csp.lp".
#show.

#program step(n).

pos(n).

&sum{ q(n) } > 0.
&sum{ q(X) } <= n :- pos(X), query(n).

&distinct{ q(X)     : pos(X) }.

&distinct{ q(X)+X-1 : pos(X) }.
&distinct{ q(X)-X+1 : pos(X) }.

&show{ q(n) }.
\end{lstlisting}
As usual, the goal is to put $n$ queens on an $n\times n$ board such that no two queens threaten each other.
Here, however,
this is done for an increasing sequence of integers $n$ such that the queens puzzle for $n$ is obtained by extending the one for $n-1$.
While the first line of Listing~\ref{encoding:multishot1} includes the \python\ program in Listing~\ref{encoding:incmode},
the next one includes the grammar from Listing~\ref{encoding:csp}.
Line~3 suppresses the output of regular atoms.
The remaining encoding makes use of two features of \clingo's exemplary incremental solving procedure,
viz.\ subsequently grounding and solving rules regrouped under program \texttt{step(n)} and the external atom \texttt{query(n)}.%
\footnote{Strictly speaking, Line 1-3 belong to the program \texttt{base} that is treated once at the beginning
  (cf.~Listing~\ref{encoding:incmode} and~\cite{gekakasc14b} for details).}
In Listing~\ref{encoding:multishot1}, all rules in lines~7-17 are regrouped under subprogram \texttt{step(n)}.
The \python\ program in Listing~\ref{encoding:incmode} makes \clingcon\ in turn solve 
the  empty program,
then program \texttt{step(1)},
then program \texttt{step(1)} and \texttt{step(2)} together,
then both former programs and \texttt{step(3)},
etc.
This is done by keeping the previous programs in the solver and by replacing parameter \texttt{n} in lines 7-17 with the respective integer when
grounding the added subprogram.
Thus, at each step \texttt{n} a fact `\texttt{pos(n).}' is added to the solver (cf.~Line~7).
The heads of Line~9 and~10 represent the linear constraints
\[
q(n) > 0
\qquad\text{ and }\qquad
q(x) \leq n \ \text{ for } x\in\{1,\dots,n\}
\ .
\]
At each step \texttt{n},
the integer variable \texttt{q(n)} is introduced and required to be a positive integer.
Moreover, all integer variables \texttt{q(1)} to \texttt{q(n)} are required to take values less or equal than $n$.
However,
while the former constraint is unconditional,
the latter are subject to the external atom \texttt{query(n)}.
The functioning of Listing~\ref{encoding:incmode} ensures that only \texttt{query(n)} is true while \texttt{query(s)} is false for all
$\texttt{s}<\texttt{n}$.
In this way, the domain of all constraint variables \texttt{q(1)} to \texttt{q(n)} is increased by one at each step.
Lines~12-15 in Listing~\ref{encoding:incmode} add distinct constraints to the effect that 
no two queens can be placed on the same row or diagonal of the board.
Line~17 simply instructs \clingcon\ to add \texttt{q(n)} to the output constraint variables.

In the following, we detail the grounding process for this example.
The base program simply consists of the first 3 lines of the original encoding.
Afterwards, program \texttt{step(1)} is grounded,
adding the first constraints of the problem.
The result is shown in Listing~\ref{encoding:multishot1grounded1}.
\begin{lstlisting}[float,caption={Grounded incremental $n$-queens program \texttt{step(1)}.},label=encoding:multishot1grounded1]
pos(1).

&sum{ q(1) } > 0.
&sum{ q(1) } <= 1 :- query(1).

&distinct{ q(1) }.

&distinct{ q(1) }.
&distinct{ q(1) }.

&show{ q(1) }.
\end{lstlisting}
The first variable \texttt{q(1)} is introduced and its lower bound is fixed to~1 in Line~3.
Its upper bound is also restricted to~1
but here only if \texttt{query(1)} holds.
This is only the case of \texttt{n=1} when solving program \texttt{step(1)} (Line~4).
In all subsequent cases, \texttt{query(1)} is false, and hence $q(1)\leq1$ is not imposed anymore.
Accordingly, the atom \mbox{\texttt{\&sum\{q(1)\} <= 1}} can vary freely
(since it is an external constraint atom).
Don't care propagation, described in Section~\ref{sec:features},
addresses such atoms and removes them from the system.

As solving the $1$-queen problem is uninteresting,
the second solving step adds program \texttt{step(2)} shown in Listing~\ref{encoding:multishot1grounded2}.
\begin{lstlisting}[float,caption={Grounded incremental $n$-queens program \texttt{step(2)}.},label=encoding:multishot1grounded2]
pos(2).

&sum{ q(2) } > 0.
&sum{ q(1) } <= 2 :- query(2).
&sum{ q(2) } <= 2 :- query(2).

&distinct{ q(1), q(2) }.

&distinct{ q(1), q(2)+1 }.
&distinct{ q(1), q(2)-1 }.

&show{ q(2) }.
\end{lstlisting}
We are now solving the second step and \texttt{query(1)} is no longer true,
which amounts to removing the rule from Line~4 in Listing~\ref{encoding:multishot1grounded1}.
The new step adds two rules for this instead (lines 4-5) and
restricts all variables to be less than or equal 2.
Also, additional distinct constraints are added involving \texttt{q(2)}.
The next step again removes the rules in lines 4-5 by making \texttt{query(2)} false
and adds a new restriction (lines 4-6 in Listing~\ref{encoding:multishot1grounded3}).
\begin{lstlisting}[float,caption={Grounded incremental $n$-queens program \texttt{step(3)}.},label=encoding:multishot1grounded3]
pos(3).

&sum{ q(3) } > 0.
&sum{ q(1) } <= 3 :- query(3).
&sum{ q(2) } <= 3 :- query(3).
&sum{ q(3) } <= 3 :- query(3).

&distinct{ q(1),q(2),q(3) }.

&distinct{ q(1), q(2)+1, q(3)+2 }.
&distinct{ q(1), q(2)-1, q(3)-2 }.

&show{ q(3) }.
\end{lstlisting}
In this way,
we not only add new variables at each step,
but also increase the upper bounds of existing ones.
For solving the third step,
the grounded rules of all three steps are taken together,
only \texttt{query(3)} is set to true,
and all previously added instances of \texttt{query/1} are false.

Listing~\ref{lst:run} shows a run of Listing~\ref{encoding:multishot1} up to 10 steps.
Setting the stop criterion to \texttt{UNKNOWN} makes sure that the process neither terminates upon satisfiable nor unsatisfiable result.
\begin{lstlisting}[float,numberblanklines,basicstyle=\ttfamily\scriptsize,caption={Running Listing~\ref{encoding:multishot1} ($Q_1$;~\texttt{incqueens.lp})},label=lst:run]
$ clingcon incqueens.lp -c imax=10 -c istop=\"UNKNOWN\"
clingcon version 3.2.0
Reading from incqueens.lp
Solving...
Answer: 1

Solving...
Answer: 1
q(1)=1
Solving...
Solving...
Solving...
Answer: 1
q(4)=2 q(3)=4 q(2)=1 q(1)=3
Solving...
Answer: 1
q(5)=3 q(1)=1 q(2)=4 q(3)=2 q(4)=5
Solving...
Answer: 1
q(6)=5 q(5)=3 q(1)=2 q(2)=4 q(3)=6 q(4)=1
Solving...
Answer: 1
q(7)=6 q(6)=3 q(5)=5 q(1)=2 q(2)=4 q(3)=1 q(4)=7
Solving...
Answer: 1
q(8)=7 q(7)=3 q(6)=1 q(5)=6 q(1)=4 q(2)=2 q(3)=5 q(4)=8
Solving...
Answer: 1
q(9)=3 q(8)=6 q(7)=8 q(6)=5 q(5)=2 q(1)=1 q(2)=4 q(3)=7 q(4)=9
SATISFIABLE

Models       : 8+
Calls        : 10
Time         : 0.075s (Solving: 0.02s 1st Model: 0.02s Unsat: 0.00s)
CPU Time     : 0.070s
\end{lstlisting}

A closer look at the distinct constraints in lines~12 to~15 of Listing~\ref{encoding:multishot1} reveals quite some redundancy.
This is because the constraints added at each step supersede the ones added previously,
and they all coexist in the system.
For example,
at Step~3 the system contains 3 instances of Line~12, namely
\texttt{\&distinct\{q(1)\}},
\texttt{\&distinct\{q(1),q(2)\}}, and
\texttt{\&distinct\{q(1),q(2),q(3)\}}.
Clearly, the first two constraints are redundant in view of the third but remain in the system.
To avoid this redundancy,
we can make use of the external atom \texttt{query(n)} to remove the redundant distinct constraints at each step
in the same way we tighten the upper bound of variable domains.
This amounts to replacing lines~12-15 in Listing~\ref{encoding:multishot1} with the ones given in Listing~\ref{encoding:multishot2} below.
\begin{minipage}{\linewidth}
\begin{lstlisting}[caption={Retracting Constraints, encoding $Q_2$},label=encoding:multishot2,firstline=12,lastline=15,firstnumber=12]
&distinct{ q(X)     : pos(X)} :- query(n).

&distinct{ q(X)+X-1 : pos(X)} :- query(n).
&distinct{ q(X)-X+1 : pos(X)} :- query(n).
\end{lstlisting}
\end{minipage}

Although the last modification guarantees that the system bears no redundant distinct constraints,%
\footnote{Given that don't care propagation is enabled by default.}
it leads to adding and removing the same restrictions over and over again.
For example, the constraint that \texttt{q(1)} and \texttt{q(2)} must have different values is included in every distinct constraint after step 1.
And this information is retracted and re-added at each step.
This is avoided by the constraints in Listing~\ref{encoding:multishot3}.
This formulation only adds constraints for the new variable \texttt{q(n)} at each step \texttt{n}
and stays clear from retracting any constraints.
\begin{minipage}{\linewidth}
\begin{lstlisting}[caption={Partial Constraints, encoding $Q_3$},label=encoding:multishot3,firstline=12,lastline=15,firstnumber=12]
&sum{ q(X)     } != q(n)     :- X=1..n-1.

&sum{ q(X)+X-1 } != q(n)+n-1 :- X=1..n-1.
&sum{ q(X)-X+1 } != q(n)-n+1 :- X=1..n-1.
\end{lstlisting}
\end{minipage}

Table~\ref{tab:nqueens} gives a comparison of the three different encodings for the incremental $n$-queens problem for 30 steps.
\begin{table}
\begin{tabular}{|l|r|r|r|}
\cline{1-4}
Measure & $Q_1$ & $Q_2$ & $Q_3$ \\
\cline{1-4}
\cline{1-4}
           Time &  138s &  10s & 16s \\
      Variables &   55k &  55k & 32k \\
Static  Nogoods &   24k &   5k & 2k \\ 
Dynamic Nogoods & 1181k & 320k & 301k \\
  \cline{1-4}
\end{tabular}
\caption{Comparison of different incremental $n$-queens programs.}%
\label{tab:nqueens}
\end{table}
The first row gives the respective total running time.
The second one reports the total number of introduced atoms.
The third one gives the sum of static nogoods generated at each step,
and the last one the sum of dynamic nogoods generated by lazy constraint propagation.
We observe that the initial encoding $Q_1$ performs worst in all aspects.
The inherent redundancy of $Q_1$ is reflected by the high number of dynamic nogoods generated by the constraint propagator.
This is the source of its inferior overall performance.
Unlike this,
the two alternative approaches bear less redundancy, as reflected by their much lower number of dynamic nogoods.
In $Q_2$, this is achieved by eliminating duplicate inferences from redundant constraints.
Although $Q_3$ even further reduces the number of atoms as well as static and dynamic nogoods,
its runtime is slightly inferior.
This is arguably due to the usage of elementary linear constraints 
rather then global distinct constraints (and the pigeon hole constraints which are enabled by default).


%% file: experiments.tex
\section{Experiments}\label{sec:experiments}

In this section,
we evaluate the afore-presented features
and compare \clingcon\ with other systems.
We performed all our benchmarks on an
Intel Xeon E5520 2.27GHz processor with Debian GNU/Linux 7.9 (wheezy).
We used a timeout of 1800 seconds and restricted main memory to 6GB\@.
In all tests, we count a memory out as a timeout.
The experiments are split into three sections.
First,
we evaluate the presented features and discuss corresponding configurations of \clingcon.
Second,
we compare \clingcon{} with state of the art CP solvers using the benchmark
classes of the \minizinc\ competition 2015.
And finally, we contrast \clingcon{} with other CASP systems
using different CASP problems.

\begin{table}[ht]
\small
\center
\begin{tabular}{|l|@{\hspace{-1mm}}r|@{\hspace{-0.7mm}}l|}
  \cline{1-3}
Option& Value & Explanation\\
  \cline{1-3}
  \cline{1-3}
\commandfont{-{}-equality-processing}& \commandfont{true} & Enable equality processing\\
\commandfont{-{}-distinct-to-card}& \commandfont{false} &  Translate distinct constraints using inequalities\\
\commandfont{-{}-distinct-pigeon}& \commandfont{true} & Use pigeon hole constraints\\
\commandfont{-{}-distinct-permutation}& \commandfont{false} & Not using permutation constraints\\
\commandfont{-{}-sort-coefficient}& \commandfont{false} & Sort by domain size first\\
\commandfont{-{}-sort-descend-coefficient}& \commandfont{true} & Sort using decreasing coefficients\\
\commandfont{-{}-sort-descend-domain}& \commandfont{false} & Sort using increasing domain sizes\\
\commandfont{-{}-split-size}& \commandfont{-1} & Not splitting constraints\\
\commandfont{-{}-max-nogoods-size}& \commandfont{1024} & Not splitting constraints with less than 1024 nogoods\\
\commandfont{-{}-translate-constraints}& \commandfont{10000} & Translate constraints with less than 10000 nogoods\\
\commandfont{-{}-break-symmetries}& \commandfont{true} & Break symmetries when splitting\\
\commandfont{-{}-domain-size}& \commandfont{10000} & Use 10000 as a threshold for domain propagation\\
\commandfont{-{}-redundant-nogood-check}& \commandfont{true} & Enable redundant nogood check when translating\\
\commandfont{-{}-dont-care-propagation}& \commandfont{true} & Enable don't care propagation\\
\commandfont{-{}-min-lits-per-var}& \commandfont{1000} & Introduce at least 1000 order atoms per variable\\
\commandfont{-{}-flatten-optimization}& \commandfont{true} & Flatten the objective function\\
\commandfont{-{}-prop-strength}& \commandfont{4} & Use highest propagation strength 4\\
\commandfont{-{}-explicit-binary-order}& \commandfont{false} & Not explicitly creating nogoods from $\Phi'(\mathcal{V},D)$\\
\commandfont{-{}-learn-nogoods}& \commandfont{true} & Add all learned nogoods to $\nabla$ immediately\\
  \cline{1-3}
\end{tabular}
\caption{Default configuration \conf{D} of \clingcon~3.2.0.}%
\label{tab:default}
\end{table}

To evaluate the presented techniques,
we give a comprehensive comparison in Table~\ref{tab:internal}.
To concentrate on the CP techniques of \clingcon~3.2.0,
we use the CP benchmarks of the \minizinc\ competition 2015.\footnote{\url{http://www.minizinc.org/challenge2015/challenge.html}}
We removed the benchmark classes \emph{large scheduling} and \emph{project planning}
as they cannot be translated into the \flatzinc\ format without the use of special global constraints.
For all other classes, we used the \mznfzn\footnote{\url{http://www.minizinc.org/software.html}} toolchain to convert all instances to \flatzinc\ while removing
all non-linear and global constraints except for distinct.
This functionality is provided by \mznfzn,
which translates non-supported constraints away.
We use the standard translation provided by \mznfzn{} to be able to handle all benchmark classes.
In this way, even problems using constraints on sets, non-linear equations,
or complex global constraints can be handled by solvers restricted to basic linear constraints.
For making this benchmark suite available to the CASP community,
we build 
a converter from \flatzinc\ to the \aspif\ format~\cite{gekakaosscwa16b} used by \clingcon;
it is called \fztoaspif.\footnote{\url{https://potassco.org/labs/2016/12/02/fz2aspif.html}}
To evaluate the different features,
we modified the scoring system of the \minizinc\ competition,
which is based on the Borda count evaluation technique.
On a per instance basis,
a configuration gets one point for every other configuration being worse.
A configuration is considered worse,
if either the found optimization value is at least $1\%$ lower,
or if it has the same optimization value but is slower.
A configuration is considered slower if it is at least 5 seconds slower.
Classes marked with~* are decision problems (all others are optimization problems);
classes containing the global distinct constraint are marked with \textdagger.
We have exactly five instances per class.

The following discussion refers to the results shown in Table~\ref{tab:internal}.
The columns used for comparison are named in the paragraph heading.
Column $D$ presents the default configuration
of \clingcon{} given in Table~\ref{tab:default}.
All other listed configurations differ only in one or two options from this default in order to test specific techniques.
For instance, for evaluating equality processing,
we compare default configuration \conf{D}, using equality processing, 
with configuration \conf{NE}, disabling equality processing.
Thus, except for \commandfont{-{}-equality-processing},
all other options remain unaltered.
\afterpage{%
\clearpage%
\thispagestyle{empty}%
\begin{landscape}
{\setlength{\tabcolsep}{0.12em}
\small
\center
\hspace{-1.0cm}
    \begin{tabular}{|l|r|r|r|r|r|r|r|r|r|r|r|r|r|r|r|r|r|r|r|r|r|r|r|r|r|r|r|}
      \cline{1-28}
      \input{internal_table.csv}

      \cline{1-28}
    \end{tabular}
}
\captionof{table}{Comparison of different features of \clingcon~3.2.0 on the benchmark set of the \minizinc\ competition 2015.
  Shown are scores of how often a configuration is better than another one.
  Bold numbers indicate the best configuration for the benchmark class.}
\label{tab:internal}
\clearpage%
\end{landscape}%
}

\begin{description}
\item[Equality Processing (\confb{D}, \confb{NE})]
To evaluate the influence of equality processing,
we compare default configuration \conf{D} (with equality processing)
with configuration \conf{NE} (without equality processing).
This feature improves performance on nearly all benchmark classes significantly.
By simply removing constraints and variables the underlying CSP gets easier to solve
(no matter if it is solved by translation or propagators).
\item[Distinct Translation (\confb{D}, \confb{DT})]
Translating global distinct constraints into cardinality rules prevents
order atoms from being created lazily.
The default configuration \conf{D} translates them
into a set of inequalities.
The translation using cardinality constraints in column \conf{DT}
performs better on \emph{cvrp}, \emph{open-stacks}, and \emph{p1f},
while it performs worse on the benchmark class \emph{costas}.
As long as the domain size is small,
this feature can be useful for problems using distinct constraints.
The configuration \conf{DT} performs best of all tested configurations.
\item[Pigeon Hole Constraints (\confb{D}, \confb{NP})]
Since pigeon hole constraints add only
constant overhead in the number of nogoods,
they are enabled in configuration \conf{D}.
Disabling their addition, 
slightly increases performance on 
benchmark classes containing distinct constraints
(marked with $\dagger$),
as witnessed in column \conf{NP}.
Although these constraints have no positive effect on the benchmarks at hand,
we keep this feature enabled by default since it increases propagation strength.
\item[Permutation Constraints (\confb{D}, \confb{PO})]
Unlike pigeon hole constraints,
permutation constraints introduce direct encoding
atoms which prevents lazy variable generation
for some constraint variables.
This is the reason why this feature is disabled by default
in configuration \conf{D}.
We enabled it in column \conf{PO}.
Again, this feature only influences benchmark classes
containing distinct constraints.
It improves performance for the \emph{cvrp} class
but decreases it on the other classes.
The impact of this feature depends upon the respective problem.
\item[Sorting (\confb{D}, \confb{SC})]
As we cannot account for all combinations of sorting mechanisms,
we evaluate this feature only on the cases discussed in~\cite{basota13a}.
Default configuration \conf{D} implements the one in \sugar;
it sorts by smallest domain first and prefers on ties larger coefficient.
The alternative sorting recommended in~\cite{basota13a} first sorts on larger coefficients
and afterwards uses the smaller domain.
This behavior is enforced by setting \commandfont{-{}-sort-coefficient=true}
and reflected in column \conf{SC}.
We see that both sorting methods yield a similar performance when applied to our lazy nogood generating approach.
\item[Splitting Constraints (\confb{D}, \confb{SP}, \confb{T_4}, \confb{ST})]
Splitting constraints into smaller ones is mandatory
for any translation-based approach using the order encoding to
avoid an exponential number of nogoods.
We restricted our evaluation to a splitting size of 3,
as done in \sugar.
The default configuration \conf{D} of \clingcon{} does not split
any constraints.
The effect of splitting constraints into ternary ones (\commandfont{-{}-split-size=3})
is reflected by column \conf{SP}; 
it performs poorly in our lazy nogood generating setting
because it introduces many new constraints and variables.
On the other hand,
when translating all constraints (\commandfont{-{}-translate-constraints=-1})
as shown in column \conf{T_4},
the split into constraints of up to three variables
(\commandfont{-{}-translate-constraints=-1} and \commandfont{-{}-split-size=3})
increases performance significantly,
as witnessed by column \conf{ST}.
We conclude that splitting constraints is not necessary for lazy nogood generating solvers
but essential for translational approaches that use the order encoding.
\item[Symmetry Breaking (\confb{SP}, \confb{NS})]
Splitting constraints introduces auxiliary variables that may lead to redundant solutions.
Symmetry breaking eliminates such redundancies and has only an effect when splitting constraints.
This is why it is interesting to compare column \conf{SP} (\commandfont{-{}-split-size=3})
where symmetry breaking is enabled
with column \conf{NS} (\commandfont{-{}-split-size=3} and \commandfont{-{}-break-symmetries=false})
where it is disabled.
In both cases,
all constraints are split into ternary ones.
The additional constraints
remove symmetric solutions from the search space
and therefore seem to be beneficial,
especially on classes \emph{tdtsp}, \emph{radiation}, and \emph{mapping}.
\item[Domain Propagation (\confb{D_1}, \confb{D_2}, \confb{SP}, \confb{D_3})]
To investigate the impact of domain propagation during preprocessing,
we tested four different configurations
that all split constraints into ternary ones (\commandfont{-{}-split-size=3}).
They only differ in using the options
\commandfont{-{}-domain-size=0} (no domain propagation) in column \conf{D_1},
\commandfont{-{}-domain-size=1000} in column \conf{D_2},
\commandfont{-{}-domain-size=10000} in column \conf{SP},
and 
\commandfont{-{}-domain-size=-1} (unlimited domain propagation) in column \conf{D_3}.
We observe that unlimited domain propagation
reduces performance in benchmark class \emph{triangular} but has no significant influence otherwise.
The other tested configurations have no influence on the runtime of the benchmarks.
We assume that domain propagation prunes the domains not
enough to make a considerable difference.
For the default configuration of \clingcon,
we decided to restrict it to a reasonable number (10000)
which leaves it enabled for mid-sized domains.
\item[Translate Constraints (\confb{T_1}, \confb{T_2}, \confb{D}, \confb{T_3}, \confb{T_4})]
We have already seen that translating all constraints as shown in column \conf{T_4} is
not very beneficial.
Now, we evaluate whether the translation of ``small'' constraints
improves performance through a mixture of
``translating small constraints'' and ``handling larger ones lazily''.
Therefore, we compare the results obtained with option \commandfont{-{}-translate-constraints=0} (no constraints are translated) in column \conf{T_1},
with \conf{T_2} where \commandfont{-{}-translate-constraints=1000} (translate constraints that produce up to 1000 nogoods) is used,
with \conf{D} using \commandfont{-{}-translate-constraints=10000} (up to 10000 nogoods),
with \conf{T_3} using \commandfont{-{}-translate-constraints=50000} (up to 50000 nogoods),
and
\conf{T_4} using \commandfont{-{}-translate-constraints=-1} (all constraints are translated).
There is a trade-off on the size of constraints to translate.
While translating small constraints (constraints that produce up to 1000 nogoods)
improves performance, the translation of larger constraints decreases it again.
On some benchmarks,
 like \emph{triangular} and \emph{p1f}, translating no constraints
is beneficial.
Also, translating all constraints in \conf{T_4} performs worst of all tested configurations.
\item[Redundant Nogood Check (\confb{ST}, \confb{NR})]
To evaluate this feature,
we decided to translate all constraints (\commandfont{-{}-translate-constraints=-1}).
Since this configuration is not producing good results for a comparison
(most of the time the translation is simply too large to be finished), we additionally split the constraints
into ternary ones with option \commandfont{-{}-split-size=3}.
With this, we compare the configuration with redundancy check in column \conf{ST}
with \conf{NR} where redundancy checking is disabled (\commandfont{-{}-redundant-nogood-check=false}).
The redundant nogood check is fast and simply removes redundant nogoods
from the order encoding.
Benchmark classes like \emph{costas} and \emph{cvrp} perform
better with the reduced set of nogoods,
while redundant nogoods are beneficial for \emph{gfd-schedule} and \emph{radiation}.
\item[Don't Care Propagation (\confb{D}, \confb{ND})]
is enabled by default
and removes unnecessary implications from the problem.
Disabling this feature (\commandfont{-{}-dont-care-propagation=false}) in column \conf{ND}
decreases performance.
\item[Order Atom Generation (\confb{M_1}, \confb{D}, \confb{M_2}, \confb{M_3})]
Adding order atoms lazily is mandatory to handle large domains.
We now evaluate the effect of adding a small amount of order atoms eagerly
for every constraint variable, evenly spread among its domain values.
Therefore, we compare
column \conf{M_1} using \commandfont{-{}-min-lits-per-var=0} (adding no atoms),
with \conf{D} using \commandfont{-{}-min-lits-per-var=1000} (adding 1000 order atoms per variable),
with \conf{M_2} using \commandfont{-{}-min-lits-per-var=10000} (adding 10000),
and
\conf{M_3} using \commandfont{-{}-min-lits-per-var=-1} (adding all order atoms).
Adding no order atoms in advance drastically reduces performance of the system
while adding 1000 to 10000 order atoms achieves best performance.
When adding too many or even all order atoms before solving,
performance is again decreased,
especially on classes with large domains like \emph{zephyrus}.
Also, note that the tested benchmark classes are very sensitive to this option
as adding atoms beforehand may influence the heuristic of the search.
\item[Flattening Objective Functions (\confb{D}, \confb{NF})]
is a feature well received by this benchmark set.
All \flatzinc\ encodings contain only one variable subject to minimization.
On most benchmark classes this variable
simply represents the sum of a set of variables.
Adding this set directly to the objective function
avoids adding an unnecessary and probably large constraint
and also improves propagation strength of the learned nogoods.
Unlike \conf{D},
configuration \conf{NF} disables this feature via \commandfont{-{}-flatten-optimization=false}.
We observe that flattening the optimization statement increases the performance on many benchmark classes.
\item[Lazy Nogood Generation (\confb{P_1}, \confb{P_2}, \confb{P_3}, \confb{D})]
We now evaluate the four afore-described propagation strengths
where
\commandfont{-{}-prop-strength=1} is reflected by the results in column \conf{P_1},
\commandfont{-{}-prop-strength=2} by the ones in column \conf{P_2},
\commandfont{-{}-prop-strength=3} in column \conf{P_3},
and
\commandfont{-{}-prop-strength=4} in the default configuration \conf{D}.
We see that a high propagation strength is important.
Especially propagating changed bounds with \commandfont{-{}-prop-strength=3}
is necessary for many benchmark classes.
Interestingly, less propagation performs best for
the classes \emph{knapsack} and \emph{triangular} where constraint propagation
is not dominating the search but still takes time.
On these classes, 
configurations with propagation strength~1 or~2
spend less time on \CspPropagation{} and more on pure CDCL search,
as attested by a much higher number of choices.
\item[Explicit Binary Order Nogoods (\confb{D}, \confb{EO})]
Default configuration \conf{D} does not introduce explicit
binary order nogoods $\Phi(\mathcal{V},D)$ but uses a propagator
for capturing the corresponding inferences lazily.
The option \commandfont{-{}-explicit-binary-order=true} (reflected by column \conf{EO}) 
creates these nogoods explicitly 
for all order atoms created during preprocessing,
leaving the others subject to lazy nogood propagation.
Although, overall performance of the implicit binary order nogoods
is better, for some benchmark classes like \emph{cvrp} and \emph{spot5} using binary order nogoods explicitly
is the best choice.
This is one of the options for which it is hard to find a clear cut default setting
and that needs consideration for each benchmark class.
\item[Reduced Nogood Learning (\confb{D}, \confb{RL})]
\clingcon's default configuration \conf{D} adds all
nogoods returned by \CspPropagation\ to the set of learned nogoods (viz.\ $\nabla$ in Algorithm~\ref{algo:propagate}).
Lazily adding these nogoods when they are actually needed for conflict analysis
is achieved with \commandfont{-{}-learn-nogoods}; the results are shown in column \conf{RL}.
The average performance of adding nogoods lazily is inferior to the one obtained
by learning all nogoods.
Nevertheless, the latter setting performs best on
\emph{costas} and \emph{nmseq}, the two decision problems in our benchmark set.
Future work has to investigate which of the nogoods have to be learned
and which of them can be added lazily.
\end{description}

Configuration \conf{DT} is the configuration with the highest overall score.
Nevertheless, \clingcon's default configuration is more conservative
since it allows for using lazy variable generation in all cases.
For instance,
with configuration \conf{DT} it is impossible to run the multi-shot $n$-queens example presented in Section~\ref{sec:multi:shot},
because $2^{30}$ order atoms had to be created per queen in order to use 
cardinality constraints for the distinct constraint.

Next,
we compare \clingcon{} to state of the art CP solvers on the same set of benchmarks
with the same scoring system.%
\begin{table}[t]
\begin{center}
    \begin{tabular}{|l||r|r|r|r|r|r|r|r|r|}
      \cline{1-9}
      \input{external_table.csv}

      \cline{1-9}
    \end{tabular}
\caption{Comparing \clingcon~3.2.0 \conf{DT} with different state of the art CP solvers
on the \minizinc\ competition 2015 benchmark set.}
\label{tab:external}
\end{center}
\end{table}
The second column of Table~\ref{tab:external} shows configuration \conf{DT} of \clingcon~3.
This is the best configuration of the internal comparison in Table~\ref{tab:internal},
which is obtained using the command line option \commandfont{-{}-distinct-to-cardinality=true}.
We compare it to \gfd\ (Mercury FD Solver), 
which is the G12 FlatZinc interpreter's default solver,
taken from the \minizinc~2.0.11 package.%
\footnote{\url{http://www.minizinc.org/software.html}}
Furthermore,
we have taken \gecode~4.4.0,\footnote{\url{http://www.gecode.org}}
a well-known classical CP solver.
Also, the lazy clause generating solvers \minisatid~3.11.0~\cite{brbodede13a}%
\footnote{With some bugfixes. Special thanks to Bart Bogaerts for his great support on this work.}
as well as \chuffed,\footnote{\url{https://github.com/geoffchu/chuffed} --- SHA 5b379ed9942ee59e8684149eae3fec1af426f6ee} 
the best solver of the \minizinc\ competition 2015.%
\footnote{It did not participate in the ranking as it is was entered by the organizers.
It ran outside of competition and was faster than the winning system.}
Finally, we compare to \picatsat~2.0,\footnote{\url{http://picat-lang.org}}
a CP solver that won the second place at the \minizinc\ competition 2016
by translating constraints into SAT using a logarithmic encoding.
We ran \gfd\ and \gecode\ with \commandfont{-{}-ignore-user-search}
to disable any special heuristic given in the problem encodings for all solvers.
In the competition, this is called ``free search''.
To measure the core performance of the systems,
it is most instructive to consider 
\chuffed$'$ and \picatsat$'$ which
use the two solvers on exactly the same set of constraints as \clingcon.
Hence, all non-linear and global constraints (except distinct)
are translated using \mznfzn\ in the same way for all systems.
\footnote{%
Unfortunately,
we were unable to compare to the lazy clause generating system
\sysfont{g12lazy}, as it produced wrong results on some of the benchmarks and is no longer maintained.
We were also unable to convert the competition benchmarks to
a format readable by \sugar,
as existing converters are outdated and not compatible anymore.}

The results in Table~\ref{tab:external} show that \clingcon%
\footnote{Note that the Borda Count scores are relative to the compared systems, 
  and therefore are different for the same configuration of \clingcon{} in Table~\ref{tab:internal} and~\ref{tab:external}.}
outperforms established systems such as \gfd, \gecode, \minisatid, and even \picatsat.
There are also different benchmark classes
where solvers dominate each other and vice versa.
We point out that \gecode\ has special propagators for
many non-linear and global constraints that have been used in the benchmarks.
Also \chuffed,
as a lazy clause generating solver,
has propagators
for many other constraints and can therefore handle some of the benchmark
classes much better.
As we are building a CASP system,
we refrain from supporting a broad variety of global constraints,
as some of them can be modeled in ASP\@.
So for a better comparison on the features of \clingcon,
we translated all non-linear and global constraints except for distinct in the
columns \chuffed$'$ and \picatsat$'$ into linear ones.
Here, we see that these systems profit from the dedicated treatment of global constraints
but that the base performance of \clingcon{} is comparable.
In general,
\clingcon{} does not match the performance of the best solver of the \minizinc\ competition 2015
but on benchmark classes like \emph{freepizza}, \emph{grid-colour}, \emph{opd},
\emph{knapsack}, and \emph{spot5}, it even outperformed \chuffed.
We conclude that \clingcon{},
despite being a CASP system,
is at eye level with state of the art CP solvers
but cannot top the best lazy clause generating systems.

Finally, we compare \clingcon\ against six other CASP systems.
\begin{itemize}
\item \inca{}~\cite{drewal10a} 
with the option \commandfont{-{}-linear-bc},\footnote{This option was recommended by the authors of the system for these kind of benchmarks.}
a lazy nogood generating system not supporting lazy variable generation.
\item \clingcon~2~\cite{ostsch12a}, using \gecode~3.7.3 as a black-box CP solver.
\item \ezcsp~1.6.24~\cite{ballie13a}, also pursuing a black-box approach but using CP solver B-Prolog~7.4 with ASP solver \clasp.
\item \aspartame~\cite{bageinospescsotawe15a}, a system using an eager translation of the constraint part by means of an ASP encoding.
\item \ezsmt~1.0.0~\cite{liesus16a}, translating CASP programs to SMT, solved by SMT solver \zthree~4.2.2.
\item \clingo~5.1.0, a pure ASP solver to measure the influence of the CP part on solving.
\end{itemize}
%
The first benchmark class is the two dimensional strip packing problem~\cite{sointabana10a};
its encoding is shown in Listing~\ref{encoding:2sp}.
\begin{table}[t]
\begin{center}
    \begin{tabular}{|l|r|r|r|r|r|r|r|r|r|r|}
      \cline{1-11}
      \input{2sp_table.csv}

      \cline{1-11}
    \end{tabular}
 \caption{Comparison of different CASP systems on the two dimensional strip packing problem.}
\label{tab:2sp}
\end{center}
 \end{table}
In Table~\ref{tab:2sp},
column \clingo~5 reflects the results obtained with a highly optimized ASP encoding,
using a handcrafted order encoding.
Time is given in seconds, letting - denote a timeout of 1800 seconds.
The best objective value computed so far is given in the columns headed with opt.
For \aspartame{},
we have taken an encoding provided in~\cite{bageinospescsotawe15a}.
For the other systems such as \clingcon~2, \clingcon~3, and \inca,
we adjusted the syntax for the linear constraints.
We refrained from comparing with \ezcsp{} or \ezsmt{} as
both systems are not supporting optimization of integer variables.
The bottom row counts the number of times a system performed best.
We clearly see that \clingcon~2 is outperformed even by the manual ASP encoding.
The new \clingcon~3 system performs best.
The translational approach of \aspartame{} is close
to the \inca{} system,
and both perform better than the manual ASP approach.
According to~\cite{sointabana10a},
these results are in accord with dedicated, state of the art systems.

The next benchmark classes are incremental scheduling, weighted sequence,
and reverse folding, all stemming from the ASP competition.\footnote{\url{http://aspcomp2015.dibris.unige.it/LPNMR-comp-report.pdf}}
Encodings for \clingo, \ezcsp,%
\footnote{To be comparable, we used the encoding without \emph{cumulative} constraint.}
\ezsmt{} and \clingcon~2 have been taken from~\cite{liesus16a}
in combination with instances from the ASP competition.%
\footnote{We refrained from using the other three benchmark classes from this source as the available instances were too easy to solve to produce informative results.}
We changed the pure ASP encoding for \clingo{} slightly for a better grounding performance.
For these classes, we could not provide an encoding for \aspartame,
as its prototypical CASP support does not allow for modeling parametrized n-ary constraints.

For incremental scheduling,
\inca{} produces wrong results due to its usage of an intermediate version of \gringo, viz.\ 3.0.92.
\begin{table}[h]
\begin{center}
    \begin{tabular}{|l|r|r|r|r|r|}
      \cline{1-6}
      \input{incSched_table.csv}

      \cline{1-6}
    \end{tabular}
 \caption{Comparison of different CASP systems on the incremental scheduling problem.}
\label{tab:incSched}
\end{center}
\end{table}
The runtime in seconds for incremental scheduling is shown in Table~\ref{tab:incSched}.
We see that \clingcon~2 improves
on the dedicated ASP encoding.
In fact, incremental scheduling is a true CASP problem
where the pure ASP encoding can be improved by using CP\@.
While the black-box approach of \ezcsp{} performs worst,
\ezsmt{} and \clingcon~3 clearly dominate this comparison.%
\footnote{The time to run the completion and translation processes for \ezcsp{} and \ezsmt{} is not included in the tables.}
The enhanced preprocessing techniques and the lazy variable generation of \clingcon{}
even outperforms the industrial SMT solver \zthree{} (as used in \ezsmt{}).

\begin{table}[h]
\begin{center}
    \begin{tabular}{|l|r|r|r|r|r|r|}
      \cline{1-7}
      \input{weighted_table.csv}

      \cline{1-7}
    \end{tabular}
 \caption{Comparison of different CASP systems on the weighted sequence problem.}
\label{tab:weighted}
\end{center}
 \end{table}
For the weighted sequence problem,
we see in Table~\ref{tab:weighted} that
\inca{}, \clingo{}, \ezsmt{}, and \clingcon~3
perform well on this benchmark set,
while \clingcon~2 could not compete with the timings of the other systems
and \ezcsp{} did not solve any of them.
Again, time is shown in seconds and - denotes a timeout of 1800 seconds.
We also see that the performance of the pure ASP encoding
is in the same range as that of the winning CASP systems.
Hence, the ASP solving part clearly dominates the CSP part.
This also explains the slightly worse performance of \clingcon~3
due to its heavy preprocessing of the CSP part.

For the reverse folding problem,
we compare the same systems as before.
Table~\ref{tab:reverse} gives the running time in seconds.
While all CASP systems improve upon the pure ASP encoding,
\clingcon~2 and \clingcon~3 perform best on this benchmark class.
The preprocessing overhead of \clingcon~3 does not pay off in terms of runtime
on this benchmark class, making it perform slightly worse than \clingcon~2.
Of the two lazy nogood generating solvers \inca{} and \clingcon~3,
the latter performs better due to lazy variable generation,
as not all order atoms have to be generated before solving.
While the black-box approach of \ezcsp{} can solve the problem,
the translation to SMT by \ezsmt{} performs even better.
We conclude that this is also due to the fact that
no auxiliary atoms for an encoding of the constraints are used in \ezsmt.
A closer inspection revealed that the number of choices for \inca{} and \clingcon~2 is
below 100 on average.
For this problem, the ASP part is dominated by the CSP part.
This is also the reason why the pure ASP encoding produces a memory out
on all instances (it was not able to ground all constraints).
\begin{table}[h]
\begin{center}
    \begin{tabular}{|l|r|r|r|r|r|r|}
      \cline{1-7}
      \input{reverse_table.csv}

      \cline{1-7}
    \end{tabular}
 \caption{Comparison of different CASP systems on the reverse folding problem.}
\label{tab:reverse}
\end{center}
 \end{table}

We conclude that \clingcon~3 improves
significantly upon its predecessor \clingcon~2,
is comparable to state of the art CP systems,
and the currently fastest CASP system available.
All benchmarks, encodings, instances and results are available online.%
\footnote{\url{https://potassco.org/clingcon}}
%

%% file: internal_table.csv
instances&\conf{D}&\conf{NE}&\conf{DT}&\conf{NP}&\conf{PO}&\conf{SC}&\conf{SP}&\conf{ST}&\conf{NS}&\conf{D_1}&\conf{D_2}&\conf{D_3}&\conf{T_1}&\conf{T_2}&\conf{T_3}&\conf{T_4}&\conf{NR}&\conf{ND}&\conf{M_1}&\conf{M_2}&\conf{M_3}&\conf{NF}&\conf{P_1}&\conf{P_2}&\conf{P_3}&\conf{EO}&\conf{RL}\\\cline{1-28}
\emph{costas}*\textdagger		&	38	&	47	&	15	&	39	&	41	&	38	&	27	&	27	&	24	&	27	&	27	&	26	&	12	&	33	&	0	&	0	&	16	&	38	&	36	&	38	&	40	&	38	&	12	&	12	&	38	&	32	&\textbf{60}\\
\emph{cvrp}\textdagger			&	49	&	38	&	90	&	61	&	65	&	49	&	51	&	44	&	52	&	51	&	51	&	51	&	31	&	67	&	41	&	0	&	34	&	55	&	93	&	95	&	97	&	49	&	9	&	9	&	49	&\textbf{100}	&	66\\
\emph{freepizza}			&\textbf{121}	&	111	&\textbf{121}	&\textbf{121}	&\textbf{121}	&\textbf{121}	&	47	&	26	&	49	&	51	&	47	&	47	&	79	&\textbf{121}	&\textbf{121}	&	26	&	26	&	100	&	99	&	107	&	102	&	115	&	109	&	113	&	120	&	120	&	109\\
\emph{gfd-schedule}			&	105	&	69	&	103	&	102	&	102	&	103	&	52	&	11	&	47	&	54	&	54	&	54	&	58	&	102	&	102	&	0	&	36	&	50	&	84	&\textbf{114}	&	112	&	103	&	81	&	81	&	105	&	102	&	93\\
\emph{grid-colour}			&\textbf{130}	&\textbf{130}	&\textbf{130}	&\textbf{130}	&\textbf{130}	&\textbf{130}	&\textbf{130}	&\textbf{130}	&\textbf{130}	&\textbf{130}	&\textbf{130}	&\textbf{130}	&\textbf{130}	&\textbf{130}	&\textbf{130}	&\textbf{130}	&\textbf{130}	&\textbf{130}	&\textbf{130}	&\textbf{130}	&\textbf{130}	&\textbf{130}	&\textbf{130}	&\textbf{130}	&\textbf{130}	&\textbf{130}	&\textbf{130}\\
\emph{is}\textdagger			&	118	&	111	&	116	&	119	&	111	&	118	&	45	&	0	&	45	&	44	&	45	&	44	&	96	&	118	&	118	&	0	&	0	&	93	&\textbf{128}	&	53	&	56	&	118	&	0	&	0	&	117	&	90	&	80\\
\emph{mapping}				&	106	&	87	&	106	&	106	&	106	&	106	&	33	&	0	&	18	&	30	&	33	&	33	&	104	&	112	&	110	&	0	&	0	&\textbf{124}	&	84	&	109	&	111	&	106	&	0	&	0	&	106	&	97	&	112\\
\emph{knapsack}			&	66	&	55	&	64	&	66	&	66	&	64	&	7	&	2	&	7	&	3	&	7	&	7	&	65	&	65	&	66	&	0	&	2	&	66	&	44	&	73	&	76	&	57	&\textbf{99}	&\textbf{99}	&	64	&	40	&	34\\
\emph{nmseq}*				&	100	&	100	&	100	&	100	&	98	&	99	&	16	&	27	&	25	&	25	&	28	&	28	&	100	&	67	&	100	&	0	&	28	&	100	&	88	&	79	&	100	&	96	&	0	&	0	&	92	&	39	&\textbf{130}\\
\emph{opd}				&	89	&	82	&	88	&	89	&	89	&	89	&\textbf{125}	&\textbf{125}	&	121	&\textbf{125}	&\textbf{125}	&\textbf{125}	&	52	&	89	&	88	&	0	&\textbf{125}	&	63	&	44	&	89	&	90	&	88	&	79	&	79	&	88	&	78	&	83\\
\emph{open-stacks}\textdagger		&	96	&	73	&	116	&	107	&	77	&	95	&	57	&	54	&	71	&	65	&	58	&	59	&	33	&	96	&	77	&	21	&	55	&	95	&	79	&	95	&	103	&	92	&	61	&	61	&	94	&\textbf{117}	&	54\\
\emph{p1f}\textdagger			&	66	&	66	&\textbf{100}	&	70	&	71	&	66	&	35	&	65	&	35	&	36	&	35	&	35	&	84	&	66	&	66	&	0	&	65	&	75	&	57	&	74	&	76	&	71	&	64	&	64	&	66	&	78	&	68\\
\emph{radiation}			&	125	&	118	&	125	&	125	&	124	&	115	&	62	&	43	&	40	&	52	&	62	&	62	&	108	&	107	&	103	&	0	&	50	&	125	&	112	&	119	&	110	&	101	&	68	&	68	&	125	&\textbf{128}	&	40\\
\emph{roster}				&\textbf{130}	&\textbf{130}	&\textbf{130}	&\textbf{130}	&\textbf{130}	&\textbf{130}	&\textbf{130}	&\textbf{130}	&\textbf{130}	&\textbf{130}	&\textbf{130}	&\textbf{130}	&\textbf{130}	&\textbf{130}	&\textbf{130}	&	0	&\textbf{130}	&\textbf{130}	&\textbf{130}	&\textbf{130}	&\textbf{130}	&\textbf{130}	&\textbf{130}	&\textbf{130}	&\textbf{130}	&\textbf{130}	&\textbf{130}\\
\emph{spot5}				&	86	&	86	&	86	&	86	&	86	&	86	&	105	&	75	&	105	&	100	&	105	&	105	&	86	&	86	&	86	&	0	&	78	&	86	&	68	&	111	&	68	&	86	&	102	&	102	&	86	&\textbf{120}	&	106\\
\emph{tdtsp}				&	77	&	69	&	77	&	77	&	77	&	77	&	94	&	0	&	35	&	94	&	94	&	94	&	47	&	76	&	30	&	0	&	0	&	3	&	70	&	59	&\textbf{119}	&	77	&	44	&	12	&	84	&	46	&	75\\
\emph{triangular}			&	15	&	15	&	15	&	15	&	15	&	15	&	38	&	11	&	38	&	38	&	38	&	15	&	56	&	15	&	15	&	0	&	11	&	15	&	15	&	15	&	15	&	8	&\textbf{77}	&\textbf{77}	&	15	&	33	&	27\\
\emph{zephyrus}			&	72	&	47	&	72	&	72	&	72	&	72	&	62	&	0	&	62	&	62	&	62	&	62	&	64	&	72	&	72	&	0	&	0	&	72	&	71	&\textbf{103}	&	0	&	72	&	0	&	0	&	46	&	72	&	7\\\cline{1-28}
total				&	1589	&	1434	&\textbf{1654}	&	1615	&	1581	&	1573	&	1116	&	770	&	1034	&	1117	&	1131	&	1107	&	1335	&	1552	&	1455	&	177	&	786	&	1420	&	1432	&	1593	&	1535	&	1537	&	1065	&	1037	&	1555	&	1552	&	1404\\

%% file: external_table.csv
instances&\multicolumn{1}{c|}{\clingcon{}}&\gfd&\gecode&\minisatid&\chuffed&\chuffed$'$&\picatsat&\picatsat$'$\\\cline{1-9}
costas*\textdagger		&	6	&	0	&\textbf{19}	&	9	&	8	&	8	&	11	&	11\\
cvrp\textdagger			&	24	&	3	&	6	&	5	&\textbf{30}	&	22	&	4	&	6\\	
freepizza			&\textbf{35}	&	15	&	0	&	31	&	26	&	26	&	3	&	3\\				
gfd-schedule			&	10	&	9	&	12	&	27	&\textbf{28}	&\textbf{28}	&	20	&	14\\		
grid-colour			&\textbf{35}	&	1	&	8	&	34	&	23	&	23	&	31	&	31\\	
is\textdagger			&	17	&	15	&	2	&	17	&\textbf{35}	&	32	&	19	&	11\\		
mapping				&	17	&	14	&	0	&	17	&\textbf{29}	&\textbf{29}	&	23	&	16\\				
knapsack			&	19	&	11	&	14	&	8	&	5	&	5	&\textbf{26}	&\textbf{26}\\	
nmseq*				&	24	&	15	&\textbf{25}	&	3	&	21	&	21	&	5	&	5\\				
opd				&	25	&	6	&	4	&	15	&	21	&	17	&	31	&\textbf{32}\\						
open-stacks\textdagger          &	20	&	0	&	9	&	8	&\textbf{35}	&\textbf{35}	&	21	&	16\\
p1f\textdagger			&	21	&	0	&	22	&	3	&\textbf{34}	&	21	&	5	&	8\\	
radiation			&	28	&	7	&	7	&	12	&\textbf{33}	&\textbf{33}	&	18	&	14\\				
roster				&\textbf{35}	&\textbf{35}	&	0	&\textbf{35}	&\textbf{35}	&\textbf{35}	&	29	&	28\\
spot5				&	20	&	10	&	0	&	19	&	16	&	16	&\textbf{35}	&	31\\					
tdtsp				&	8	&\textbf{35}	&	10	&	1	&	31	&	24	&	10	&	9\\
triangular			&	15	&\textbf{29}	&	5	&	26	&	23	&	24	&	12	&	12\\			
zephyrus			&	3	&	20	&	7	&	10	&	26	&	26	&\textbf{30}	&\textbf{30}\\\cline{1-9}			
total				&	362	&	225	&	150	&	280	&\textbf{459}	&	425	&	333	&	303\\

%% file: 2sp_table.csv
&\multicolumn{2}{c|}{\clingo~5}&\multicolumn{2}{c|}{\aspartame}&\multicolumn{2}{c|}{\clingcon~2}&\multicolumn{2}{c|}{\inca}&\multicolumn{2}{c|}{\clingcon~3} \\
instances&time&opt&time&opt&time&opt&time&opt&time&opt\\\cline{1-11}
BENG01	&	9	&	30	&	20	&	30	&	-	&		&	916	&	30	&\textbf{2}	&\textbf{30}	\\
BENG02	&	-	&	58	&\textbf{1336}	&\textbf{57}	&	-	&		&	-	&	58	&	-	&	58	\\
BENG03	&	-	&	87	&	-	&	85	&	-	&		&	-	&	85	&\textbf{775}	&\textbf{84}	\\
BENG04	&	-	&	111	&	-	&\textbf{108}	&	-	&		&	-	&\textbf{108}	&	-	&\textbf{108}	\\
BENG05	&	-	&	141	&	-	&\textbf{136}	&	-	&		&	-	&\textbf{136}	&	-	&\textbf{136}	\\
BENG06	&	1226	&	36	&	32	&	36	&	-	&		&\textbf{5}	&\textbf{36}	&	23	&	36	\\
BENG07	&	-	&	69	&	-	&\textbf{68}	&	-	&		&	-	&	69	&	-	&\textbf{68}	\\
BENG08	&	-	&		&	-	&		&	-	&		&	-	&	104	&	-	&\textbf{103}	\\
BENG09	&	-	&		&	-	&		&	-	&		&	-	&		&	-	&\textbf{128}	\\
BENG10	&	-	&		&	-	&		&	-	&		&	-	&		&	-	&\textbf{158}	\\
CGCUT01	&\textbf{1}	&\textbf{23}	&\textbf{1}	&\textbf{23}	&	-	&	26	&	-	&	25	&\textbf{0}	&\textbf{23}	\\
CGCUT02	&	-	&	66	&	-	&\textbf{65}	&	-	&		&	-	&	67	&	-	&\textbf{65}	\\
CGCUT03	&	-	&		&	-	&		&	-	&		&	-	&		&	-	&		\\
GCUT01	&	-	&	1016	&	5	&	1016	&\textbf{0}	&\textbf{1016}	&\textbf{0}	&\textbf{1016}	&\textbf{0}	&\textbf{1016}	\\
GCUT02	&	-	&	1242	&	-	&	1195	&	-	&		&	-	&\textbf{1190}	&	-	&\textbf{1190}	\\
GCUT03	&	-	&		&	134	&	1803	&	-	&		&\textbf{1}	&\textbf{1803}	&	12	&	1803	\\
GCUT04	&	-	&		&	-	&		&	-	&		&	-	&		&	-	&		\\
HT01	&\textbf{1}	&\textbf{20}	&\textbf{1}	&\textbf{20}	&	-	&	22	&	346	&	20	&\textbf{0}	&\textbf{20}	\\
HT02	&	8	&	20	&	5	&	20	&	-	&	25	&	77	&	20	&\textbf{1}	&\textbf{20}	\\
HT03	&\textbf{1}	&\textbf{20}	&\textbf{1}	&\textbf{20}	&	-	&		&	10	&	20	&\textbf{0}	&\textbf{20}	\\
HT04	&	840	&	15	&	33	&	15	&	-	&		&	-	&	16	&\textbf{8}	&\textbf{15}	\\
HT05	&	12	&	15	&\textbf{9}	&\textbf{15}	&	-	&		&\textbf{8}	&\textbf{15}	&	13	&	15	\\
HT06	&	14	&	15	&	8	&	15	&	-	&		&	359	&	15	&\textbf{1}	&\textbf{15}	\\
HT07	&	-	&	31	&\textbf{175}	&\textbf{30}	&	-	&		&	-	&	31	&	-	&	31	\\
HT08	&\textbf{1284}	&\textbf{30}	&	-	&	31	&	-	&		&	-	&	31	&	-	&	31	\\
HT09	&	-	&	31	&	272	&	37	&	-	&		&	-	&	31	&\textbf{41}	&\textbf{30}	\\
NGCUT01	&\textbf{0}	&\textbf{23}	&\textbf{0}	&\textbf{23}	&\textbf{1}	&\textbf{23}	&\textbf{0}	&\textbf{23}	&\textbf{0}	&\textbf{23}	\\
NGCUT02	&\textbf{2}	&\textbf{30}	&\textbf{1}	&\textbf{30}	&	-	&	33	&	80	&	30	&\textbf{0}	&\textbf{30}	\\
NGCUT03	&\textbf{2}	&\textbf{28}	&\textbf{2}	&\textbf{28}	&	-	&		&\textbf{1}	&\textbf{28}	&\textbf{0}	&\textbf{28}	\\
NGCUT04	&\textbf{0}	&\textbf{20}	&\textbf{0}	&\textbf{20}	&\textbf{0}	&\textbf{20}	&\textbf{0}	&\textbf{20}	&\textbf{0}	&\textbf{20}	\\
NGCUT05	&\textbf{0}	&\textbf{36}	&\textbf{0}	&\textbf{36}	&	-	&		&\textbf{0}	&\textbf{36}	&\textbf{0}	&\textbf{36}	\\
NGCUT06	&	8	&	31	&\textbf{1}	&\textbf{31}	&	-	&		&\textbf{0}	&\textbf{31}	&\textbf{0}	&\textbf{31}	\\
NGCUT07	&\textbf{0}	&\textbf{20}	&\textbf{0}	&\textbf{20}	&\textbf{0}	&\textbf{20}	&\textbf{0}	&\textbf{20}	&\textbf{0}	&\textbf{20}	\\
NGCUT08	&\textbf{1}	&\textbf{33}	&\textbf{1}	&\textbf{33}	&	-	&	36	&	38	&\textbf{33}	&\textbf{0}	&\textbf{33}	\\
NGCUT09	&\textbf{87}	&\textbf{50}	&	-	&	50	&	-	&	57	&	-	&	50	&	1549	&	50	\\
NGCUT10	&	6	&	80	&\textbf{1}	&\textbf{80}	&	-	&	81	&\textbf{0}	&\textbf{80}	&\textbf{0}	&\textbf{80}	\\
NGCUT11	&	\textbf{4}	&	52	&\textbf{1}	&\textbf{52}	&	-	&	55	&\textbf{0}	&\textbf{52}	&\textbf{0}	&\textbf{52}	\\
NGCUT12	&	-	&	87	&	\textbf{3}	&	87	&	-	&		&\textbf{0}	&\textbf{87}	&\textbf{0}	&\textbf{87}	\\\cline{1-11}
\#best	&\multicolumn{2}{c|}{13}&\multicolumn{2}{c|}{21}&\multicolumn{2}{c|}{4}&\multicolumn{2}{c|}{16}&\multicolumn{2}{c|}{\textbf{28}} \\

%% file: incSched_table.csv
instances&\clingo~5&\clingcon~2&\clingcon~3&\ezcsp{}&\ezsmt\\\cline{1-6}
020-inc	&	302	&\textbf{1}	&\textbf{0}	&\textbf{0}	&\textbf{0}\\ 
028-inc	&	-	&	16	&\textbf{4}	&	-	&\textbf{5}\\
044-inc	&	-	&	518	&	149	&	-	&\textbf{116}\\
063-inc	&	335	&\textbf{0}	&\textbf{1}	&	-	&\textbf{0}\\
083-inc	&	268	&	-	&\textbf{1}	&	-	&\textbf{0}\\
096-inc	&	719	&	298	&\textbf{1}	&	-	&\textbf{0}\\
106-inc	&	470	&	-	&\textbf{2}	&	-	&\textbf{1}\\
158-inc	&	355	&	\textbf{4}	&\textbf{1}	&	\textbf{0}	&\textbf{1}\\
175-inc	&	-	&	83	&	\textbf{6}	&	-	&\textbf{4}\\
181-inc	&	-	&	\textbf{5}	&\textbf{1}	&	-	&\textbf{2}\\
184-inc	&	425	&	-	&\textbf{1}	&	-	&\textbf{1}\\
211-inc	&	-	&	-	&\textbf{194}	&	-	&	799\\
214-inc	&	-	&	-	&\textbf{7}	&	-	&	76\\
230-inc	&	-	&	77	&\textbf{9}	&	-	&	24\\
256-inc	&	-	&	-	&	-	&	-	&	-\\
257-inc	&	-	&	-	&	-	&	-	&	-\\
266-inc	&	-	&	-	&\textbf{767}	&	-	&	-\\
334-inc	&	-	&	-	&	-	&	-	&	-\\
338-inc	&	-	&	-	&	-	&	-	&	-\\
362-inc	&	-	&	-	&	-	&	-	&	-\\\cline{1-6}
\#best  & 0 & 4 & \textbf{14} & 2 & 11 \\

%% file: weighted_table.csv
instances&\clingo~5&\inca&\clingcon~2&\clingcon~3&\ezcsp&\ezsmt\\\cline{1-7}
01-tree	&	\textbf{6}	&	\textbf{5}	&	\textbf{6}	&	\textbf{1}	&	-	&	\textbf{1}	\\   
02-tree	&	\textbf{8}	&	\textbf{1}	&	\textbf{8}	&	\textbf{3}	&	-	&	\textbf{5}	\\
05-tree	&	\textbf{5}	&	\textbf{1}	&	\textbf{4}	&	\textbf{1}	&	-	&	\textbf{3}	\\
06-tree	&	\textbf{3}	&	\textbf{2}	&	19	&	\textbf{1}	&	-	&	\textbf{3}	\\
07-tree	&	\textbf{3}	&	\textbf{4}	&	\textbf{1}	&	\textbf{1}	&	-	&	\textbf{4}	\\
11-tree	&	\textbf{2}	&	\textbf{0}	&	6	&	\textbf{1}	&	-	&	\textbf{0}	\\
15-tree	&	\textbf{2}	&	\textbf{3}	&	\textbf{1}	&	\textbf{1}	&	-	&	\textbf{1}	\\
16-tree	&	\textbf{2}	&	\textbf{4}	&	15	&	\textbf{5}	&	-	&	\textbf{1}	\\
22-tree	&	\textbf{4}	&	\textbf{0}	&	\textbf{4}	&	\textbf{1}	&	-	&	\textbf{0}	\\
23-tree	&	\textbf{3}	&	\textbf{0}	&	\textbf{4}	&	\textbf{0}	&	-	&	\textbf{3}	\\
26-tree	&	\textbf{2}	&	\textbf{3}	&	51	&	\textbf{4}	&	-	&	\textbf{1}	\\
29-tree	&	7	&	\textbf{1}	&	26	&	\textbf{2}	&	-	&	\textbf{5}	\\
33-tree	&	\textbf{9}	&	\textbf{7}	&	95	&	15	&	-	&	\textbf{9}	\\
35-tree	&	24	&	\textbf{10}	&	231	&	30	&	-	&	26	\\
38-tree	&	\textbf{7}	&	17	&	30	&	\textbf{10}	&	-	&	\textbf{11}	\\
39-tree	&	\textbf{6}	&	\textbf{7}	&	330	&	46	&	-	&	23	\\
40-tree	&	\textbf{3}	&	9	&	398	&	47	&	-	&	14	\\
41-tree	&	36	&	\textbf{14}	&	\textbf{18}	&	\textbf{13}	&	-	&	34	\\
49-tree	&	\textbf{7}	&	\textbf{3}	&	220	&	22	&	-	&	14	\\
53-tree	&	\textbf{5}	&	8	&	297	&	33	&	-	&	\textbf{2}	\\\cline{1-7}
\#best & \textbf{17}  & \textbf{17} & 8 & 14 & 0 & 15 \\

%% file: reverse_table.csv
instances&\clingo~5&\inca&\clingcon~2&\clingcon~3&\ezcsp&\ezsmt\\\cline{1-7}
07-reverse	&	-	&	\textbf{1}	&	\textbf{0}	&	\textbf{1}	&	11	&	\textbf{1}	\\ 
11-reverse	&	-	&	\textbf{1}	&	\textbf{0}	&	\textbf{1}	&	9	&	\textbf{1}	\\
15-reverse	&	-	&	\textbf{1}	&	\textbf{1}	&	\textbf{1}	&	\textbf{6}	&	\textbf{1}	\\
18-reverse	&	-	&	\textbf{2}	&	\textbf{1}	&	\textbf{2}	&	27	&	\textbf{1}	\\
20-reverse	&	-	&	\textbf{6}	&	\textbf{3}	&	\textbf{4}	&	60	&	10	\\
24-reverse	&	-	&	12	&	\textbf{5}	&	\textbf{8}	&	272	&	49	\\
28-reverse	&	-	&	11	&	\textbf{5}	&	\textbf{7}	&	107	&	\textbf{8}	\\
31-reverse	&	-	&	20	&	\textbf{8}	&	36	&	128	&	73	\\
34-reverse	&	-	&	25	&	\textbf{11}	&	20	&	625	&	112	\\
35-reverse	&	-	&	35	&	\textbf{15}	&	23	&	353	&	96	\\
39-reverse	&	-	&	40	&	\textbf{18}	&	\textbf{23}	&	682	&	212	\\
44-reverse	&	-	&	-	&	\textbf{33}	&	\textbf{38}	&	-	&	339	\\
47-reverse	&	-	&	\textbf{6}	&	\textbf{4}	&	\textbf{4}	&	86	&	\textbf{4}	\\
49-reverse	&	-	&	\textbf{7}	&	\textbf{4}	&	\textbf{4}	&	67	&	\textbf{4}	\\
50-reverse	&	-	&	\textbf{2}	&	\textbf{1}	&	\textbf{2}	&	12	&	\textbf{4}	\\\cline{1-7}
\#best      & 0 & 8           & \textbf{15} & 12 & 1 & 8 \\

%% file: discussion.tex
\section{Discussion}\label{sec:discussion}

CASP combines ASP with CP, and thus brings together various techniques from both areas.
Groundbreaking work has been done with the systems \sysfont{ac}- and \adsolver~\cite{megezh08a,melgel08a}
by using an off-the-shelf CP solver.
This is called a black-box approach.
It features a very high abstraction level
and allows for great flexibility, for instance, 
for changing solvers or theories.
Unfortunately, this high abstraction hinders tight
integration techniques that are necessary to
achieve a performance suitable for real world problems.
Still using a black-box CP solver but having a tighter integration
into modern CDCL algorithms is common to systems like \ezcsp{} and its extensions~\cite{balduccini09a,ballie13a},
\dlvhex~\cite{eifikrre12a}, and \clingcon~2~\cite{geossc09a}.
These systems use a CP solver for propagation and consistency checking.
No auxiliary variables are used to represent non-Boolean variables.
This prevents these systems from producing strong reasons and conflicts,
needed for effective CDCL-based search.
The system \clingcon~2 tries to circumvent this problem.
It strengthens propagation
and integration~\cite{ostsch12a} by using filtering techniques
and special knowledge about the theory.
A different way to tackle the problem is the eager approach.
The theory part of the problem is translated to ASP, SAT, or SMT
in a preprocessing step.
\dingo~\cite{jalini11a} translates ASP enriched with difference constraints to SMT,
\ezsmt{} translates CASP to SMT,
and \aspartame~\cite{bageinospescsotawe15a} provides an ASP
encoding to translate CP (and CASP) into ASP\@.
The eager approach has the strongest integration because only one solver without dedicated propagators is used to solve the problem.
The features of modern CDCL algorithms such as
conflict driven heuristics and learning are supported
natively without any change to the ASP solver.
Nevertheless, translational methods into ASP or SAT have the drawback
of being very memory intensive
since the whole theory has to be represented using propositional variables.
Encodings that use for instance binary representations of integer variables
lack propagation strength.
To overcome these problems,
\inca~\cite{drewal12a} translates constraints on the fly,
that is, it relies upon lazy nogood generation,
which is strongly inspired by lazy clause generation~\cite{ohstco09a}.
It features a tight integration,
profits from the learning capabilities of CDCL, and 
avoids the grounding bottleneck of eager techniques
since only the currently interesting part of the theory is generated.
\inca{} concentrates on the support of various encodings
and implements a propagator for linear and distinct constraints.
Unfortunately,
the basic vocabulary of these encodings has to be provided beforehand,
not removing the grounding bottleneck for variables with large domains.
Lazy variable generation~\cite{thistu09a} overcomes this problem
and is a state of the art technique in CP\@.
Close work in the neighboring area of model expansion
was done in the \idp{} system~\cite{brbobrde14a}
using \minisatid~\cite{brbodede13a}
for combining
lazy clause generation and lazy variable generation
for handling linear constraints.
Also, the constraint solver \chuffed{},
the leading CP solver in the \minizinc{} competition 2015,
supports lazy clause and variable generation
and has propagators for a set of global constraints.

We take this up to extend ASP with CP for tackling CASP problems with modern CP techniques.
By extending the input language of \gringo{} in a modular way,
we enhance the modeling capabilities of ASP with linear constraints over integers
and handle them with advanced hybrid search and propagation techniques.
Our design goal is to have a tight integration,
overcoming grounding and memory bottlenecks of translation-based approaches,
while using the learning capabilities of CDCL algorithms.
We integrated these techniques in \clasp{} and \clingo{}
while preserving features like multi-threading,
unsatisfiable core optimization, multi-objective optimization etc.
We developed a propagator for linear constraints
and are able to translate parts of the constraints beforehand.
Furthermore, variables with huge domains are managed by introducing order atoms
on the fly.
Several dedicated preprocessing techniques 
improve our lazy nogood generation approach.
Our empirical evaluation leads to the result that some techniques
that are known to be crucial for translational approaches
using the order encoding cannot be adopted easily.
This concerns especially sorting and splitting of constraints,
which has either no or even a negative effect on the performance of
lazy nogood propagation.
Other, more general techniques
like equality processing, don't care propagation, and flattening of the objective function
improve the performance in general.
Another interesting result is that translating a subset of small constraints
is beneficial over translating none or all.
These techniques have allowed us to develop the modern CASP solver \clingcon.
It combines the first-order modeling language of ASP
with the performance of state of the art CP solvers
for handling constraints over integers.
Also, making \clingcon{} incremental,
such that multi-shot solving can be used with \clingo's API,
enables us to use CASP in reactive environments and thus opens up new application areas.
Our software is open source and freely available
as part of the potassco project.\footnote{https://potassco.org}

\paragraph{Future work.}

CASP is a useful paradigm to
solve problems with resources, capacities, and fine-grained timing information.
Its semantics has been extended in various ways, as for instance in
bound founded ASP~\cite{azchst13a}
or default reasoning with constraints~\cite{cakaossc16a}.
The latter approach already presents a translator relying upon \clingcon~3.
This indicates that related approaches can take advantage of the development
of CASP and its systems.

We plan to develop a translation option for
converting a CASP problem into an (C)ASP problem
by (partially) translating the constraints.
The output can then be handled by other solvers than \clasp.
We also preserved special functionalities of the ASP solver \clasp{} 
in order to use unsatisfiable core techniques~\cite{ankamasc12a}
and multi-criteria optimization~\cite{gekakasc11c} for integer variables.
Also, domain-specific heuristics~\cite{gekaotroscwa13a} can be
used in the encodings of CASP problems.
However, all these features still need to be evaluated in the context of CASP\@.
Furthermore,
we want to use the ability to handle constraints over large domains
to tackle complex planning problems~\cite{bamama16a}.
These often involve a fine grained handling of resources and timings
and are a perfect area of application for CASP\@.

\input{acknowledgments}

%% file: acknowledgments.tex
\paragraph{Acknowledgments.}

This work was partially funded by JSPS (KAKENHI~15K00099) and DFG (SCHA~550/9).
We are grateful to Bart Bogaerts for his help with \minisatid.
A special thanks goes to Philipp Wanko for his comments, and of course, 
to Roland Kaminski for all his support!
%

%% file: paper.bbl
\begin{thebibliography}{}

\bibitem[\protect\citeauthoryear{Andres, Kaufmann, Matheis, and Schaub}{Andres
  et~al\mbox{.}}{2012}]{ankamasc12a}
{\sc Andres, B.}, {\sc Kaufmann, B.}, {\sc Matheis, O.}, {\sc and} {\sc Schaub,
  T.} 2012.
\newblock Unsatisfiability-based optimization in clasp.
\newblock See \citeN{iclp-lipics12}, 212--221.

\bibitem[\protect\citeauthoryear{Aziz, Chu, and Stuckey}{Aziz
  et~al\mbox{.}}{2013}]{azchst13a}
{\sc Aziz, R.}, {\sc Chu, G.}, {\sc and} {\sc Stuckey, P.} 2013.
\newblock Stable model semantics for founded bounds.
\newblock {\em Theory and Practice of Logic Programming\/}~{\em 13,\/}~4-5,
  517--532.

\bibitem[\protect\citeauthoryear{Balduccini}{Balduccini}{2009}]{balduccini09a}
{\sc Balduccini, M.} 2009.
\newblock Representing constraint satisfaction problems in answer set
  programming.
\newblock In {\em Proceedings of the Second Workshop on Answer Set Programming
  and Other Computing Paradigms (ASPOCP'09)}, {W.~Faber} {and} {J.~Lee}, Eds.
  16--30.

\bibitem[\protect\citeauthoryear{Balduccini and Lierler}{Balduccini and
  Lierler}{2013}]{ballie13a}
{\sc Balduccini, M.} {\sc and} {\sc Lierler, Y.} 2013.
\newblock Integration schemas for constraint answer set programming: a case
  study.
\newblock {\em Theory and Practice of Logic Programming\/}~{\em 13}.

\bibitem[\protect\citeauthoryear{Balduccini, Magazzeni, and Maratea}{Balduccini
  et~al\mbox{.}}{2016}]{bamama16a}
{\sc Balduccini, M.}, {\sc Magazzeni, D.}, {\sc and} {\sc Maratea, M.} 2016.
\newblock {PDDL+} planning via constraint answer set programming.
\newblock {\em CoRR\/}~{\em abs/1609.00030}.

\bibitem[\protect\citeauthoryear{Banbara, Gebser, Inoue, Ostrowski, Peano,
  Schaub, Soh, Tamura, and Weise}{Banbara
  et~al\mbox{.}}{2015}]{bageinospescsotawe15a}
{\sc Banbara, M.}, {\sc Gebser, M.}, {\sc Inoue, K.}, {\sc Ostrowski, M.}, {\sc
  Peano, A.}, {\sc Schaub, T.}, {\sc Soh, T.}, {\sc Tamura, N.}, {\sc and} {\sc
  Weise, M.} 2015.
\newblock aspartame: Solving constraint satisfaction problems with answer set
  programming.
\newblock In {\em Proceedings of the Thirteenth International Conference on
  Logic Programming and Nonmonotonic Reasoning (LPNMR'15)}, {F.~Calimeri},
  {G.~Ianni}, {and} {M.~Truszczy{\'n}ski}, Eds. Lecture Notes in Artificial
  Intelligence, vol. 9345. Springer-Verlag, 112--126.

\bibitem[\protect\citeauthoryear{Barrett, Sebastiani, Seshia, and
  Tinelli}{Barrett et~al\mbox{.}}{2009}]{baseseti09a}
{\sc Barrett, C.}, {\sc Sebastiani, R.}, {\sc Seshia, S.}, {\sc and} {\sc
  Tinelli, C.} 2009.
\newblock Satisfiability modulo theories.
\newblock See \citeN{SATHandbook}, Chapter~26, 825--885.

\bibitem[\protect\citeauthoryear{Baselice, Bonatti, and Gelfond}{Baselice
  et~al\mbox{.}}{2005}]{baboge05a}
{\sc Baselice, S.}, {\sc Bonatti, P.}, {\sc and} {\sc Gelfond, M.} 2005.
\newblock Towards an integration of answer set and constraint solving.
\newblock In {\em Proceedings of the Twenty-first International Conference on
  Logic Programming (ICLP'05)}, {M.~Gabbrielli} {and} {G.~Gupta}, Eds. Lecture
  Notes in Computer Science, vol. 3668. Springer-Verlag, 52--66.

\bibitem[\protect\citeauthoryear{Biere, Heule, {van Maaren}, and Walsh}{Biere
  et~al\mbox{.}}{2009}]{SATHandbook}
{\sc Biere, A.}, {\sc Heule, M.}, {\sc {van Maaren}, H.}, {\sc and} {\sc Walsh,
  T.}, Eds. 2009.
\newblock {\em Handbook of Satisfiability}. Frontiers in Artificial
  Intelligence and Applications, vol. 185.
\newblock IOS Press.

\bibitem[\protect\citeauthoryear{Brodsky}{Brodsky}{2013}]{ictai13}
{\sc Brodsky, A.}, Ed. 2013.
\newblock {\em Proceedings of the Twenty-fifth IEEE International Conference on
  Tools with Artificial Intelligence (ICTAI'13)}. IEEE Computer Society.

\bibitem[\protect\citeauthoryear{Cabalar, Kaminski, Ostrowski, and
  Schaub}{Cabalar et~al\mbox{.}}{2016}]{cakaossc16a}
{\sc Cabalar, P.}, {\sc Kaminski, R.}, {\sc Ostrowski, M.}, {\sc and} {\sc
  Schaub, T.} 2016.
\newblock An {ASP} semantics for default reasoning with constraints.
\newblock In {\em Proceedings of the Twenty-fifth International Joint
  Conference on Artificial Intelligence (IJCAI'16)}, {R.~Kambhampati}, Ed.
  IJCAI/AAAI Press, 1015--1021.

\bibitem[\protect\citeauthoryear{Carro and King}{Carro and
  King}{2016}]{iclp-lipics16}
{\sc Carro, M.} {\sc and} {\sc King, A.}, Eds. 2016.
\newblock {\em Technical Communications of the Thirty-second International
  Conference on Logic Programming (ICLP'16)}. Vol.~52. Open Access Series in
  Informatics (OASIcs).

\bibitem[\protect\citeauthoryear{Crawford and Baker}{Crawford and
  Baker}{1994}]{crabak94a}
{\sc Crawford, J.} {\sc and} {\sc Baker, A.} 1994.
\newblock Experimental results on the application of satisfiability algorithms
  to scheduling problems.
\newblock In {\em Proceedings of the Twelfth National Conference on Artificial
  Intelligence (AAAI'94)}, {B.~Hayes-Roth} {and} {R.~Korf}, Eds. AAAI Press,
  1092--1097.

\bibitem[\protect\citeauthoryear{Davis, Logemann, and Loveland}{Davis
  et~al\mbox{.}}{1962}]{dalolo62a}
{\sc Davis, M.}, {\sc Logemann, G.}, {\sc and} {\sc Loveland, D.} 1962.
\newblock A machine program for theorem-proving.
\newblock {\em Communications of the ACM\/}~{\em 5}, 394--397.

\bibitem[\protect\citeauthoryear{Davis and Putnam}{Davis and
  Putnam}{1960}]{davput60a}
{\sc Davis, M.} {\sc and} {\sc Putnam, H.} 1960.
\newblock A computing procedure for quantification theory.
\newblock {\em Journal of the ACM\/}~{\em 7}, 201--215.

\bibitem[\protect\citeauthoryear{{De Cat}, Bogaerts, Bruynooghe, and
  Denecker}{{De Cat} et~al\mbox{.}}{2014}]{brbobrde14a}
{\sc {De Cat}, B.}, {\sc Bogaerts, B.}, {\sc Bruynooghe, M.}, {\sc and} {\sc
  Denecker, M.} 2014.
\newblock Predicate logic as a modelling language: The {IDP} system.
\newblock {\em CoRR\/}~{\em abs/1401.6312}.

\bibitem[\protect\citeauthoryear{{De Cat}, Bogaerts, Devriendt, and
  Denecker}{{De Cat} et~al\mbox{.}}{2013}]{brbodede13a}
{\sc {De Cat}, B.}, {\sc Bogaerts, B.}, {\sc Devriendt, J.}, {\sc and} {\sc
  Denecker, M.} 2013.
\newblock Model expansion in the presence of function symbols using constraint
  programming.
\newblock See \citeN{ictai13}, 1068--1075.

\bibitem[\protect\citeauthoryear{Dovier and {Santos Costa}}{Dovier and {Santos
  Costa}}{2012}]{iclp-lipics12}
{\sc Dovier, A.} {\sc and} {\sc {Santos Costa}, V.}, Eds. 2012.
\newblock {\em Technical Communications of the Twenty-eighth International
  Conference on Logic Programming (ICLP'12)}. Vol.~17. Leibniz International
  Proceedings in Informatics (LIPIcs).

\bibitem[\protect\citeauthoryear{Drescher}{Drescher}{2015}]{drescher15a}
{\sc Drescher, C.} 2015.
\newblock Conflict-driven constraint answer set solving.
\newblock Ph.D. thesis, Computer Science and Engineering, Faculty of
  Engineering, UNSW.

\bibitem[\protect\citeauthoryear{Drescher and Walsh}{Drescher and
  Walsh}{2010}]{drewal10a}
{\sc Drescher, C.} {\sc and} {\sc Walsh, T.} 2010.
\newblock A translational approach to constraint answer set solving.
\newblock {\em Theory and Practice of Logic Programming\/}~{\em 10,\/}~4-6,
  465--480.

\bibitem[\protect\citeauthoryear{Drescher and Walsh}{Drescher and
  Walsh}{2012}]{drewal12a}
{\sc Drescher, C.} {\sc and} {\sc Walsh, T.} 2012.
\newblock Answer set solving with lazy nogood generation.
\newblock See \citeN{iclp-lipics12}, 188--200.

\bibitem[\protect\citeauthoryear{Eiter, Fink, Krennwallner, and Redl}{Eiter
  et~al\mbox{.}}{2012}]{eifikrre12a}
{\sc Eiter, T.}, {\sc Fink, M.}, {\sc Krennwallner, T.}, {\sc and} {\sc Redl,
  C.} 2012.
\newblock Conflict-driven {ASP} solving with external sources.
\newblock {\em Theory and Practice of Logic Programming\/}~{\em 12,\/}~4-5,
  659--679.

\bibitem[\protect\citeauthoryear{Feydy, Somogyi, and Stuckey}{Feydy
  et~al\mbox{.}}{2011}]{fesost11a}
{\sc Feydy, T.}, {\sc Somogyi, Z.}, {\sc and} {\sc Stuckey, P.} 2011.
\newblock Half reification and flattening.
\newblock In {\em Proceedings of the Seventeenth International Conference on
  Principles and Practice of Constraint Programming (CP'11)}, {J.~Lee}, Ed.
  Lecture Notes in Computer Science, vol. 6876. Springer-Verlag, 286--301.

\bibitem[\protect\citeauthoryear{Gebser, Kaminski, Kaufmann, Ostrowski, Schaub,
  and Wanko}{Gebser et~al\mbox{.}}{2016a}]{gekakaosscwa16a}
{\sc Gebser, M.}, {\sc Kaminski, R.}, {\sc Kaufmann, B.}, {\sc Ostrowski, M.},
  {\sc Schaub, T.}, {\sc and} {\sc Wanko, P.} 2016a.
\newblock Theory solving made easy with clingo~5.
\newblock See \citeN{iclp-lipics16}, 2:1--2:15.

\bibitem[\protect\citeauthoryear{Gebser, Kaminski, Kaufmann, Ostrowski, Schaub,
  and Wanko}{Gebser et~al\mbox{.}}{2016b}]{gekakaosscwa16b}
{\sc Gebser, M.}, {\sc Kaminski, R.}, {\sc Kaufmann, B.}, {\sc Ostrowski, M.},
  {\sc Schaub, T.}, {\sc and} {\sc Wanko, P.} 2016b.
\newblock Theory solving made easy with clingo~5 (extended version).
\newblock Available at \url{http://www.cs.uni-potsdam.de/wv/publications/}.
\newblock Extended version of~\cite{gekakaosscwa16a}.

\bibitem[\protect\citeauthoryear{Gebser, Kaminski, Kaufmann, and Schaub}{Gebser
  et~al\mbox{.}}{2011}]{gekakasc11c}
{\sc Gebser, M.}, {\sc Kaminski, R.}, {\sc Kaufmann, B.}, {\sc and} {\sc
  Schaub, T.} 2011.
\newblock Multi-criteria optimization in answer set programming.
\newblock In {\em Technical Communications of the Twenty-seventh International
  Conference on Logic Programming (ICLP'11)}, {J.~Gallagher} {and}
  {M.~Gelfond}, Eds. Vol.~11. Leibniz International Proceedings in Informatics
  (LIPIcs), 1--10.

\bibitem[\protect\citeauthoryear{Gebser, Kaminski, Kaufmann, and Schaub}{Gebser
  et~al\mbox{.}}{2012}]{gekakasc12a}
{\sc Gebser, M.}, {\sc Kaminski, R.}, {\sc Kaufmann, B.}, {\sc and} {\sc
  Schaub, T.} 2012.
\newblock {\em Answer Set Solving in Practice}.
\newblock Synthesis Lectures on Artificial Intelligence and Machine Learning.
  Morgan and Claypool Publishers.

\bibitem[\protect\citeauthoryear{Gebser, Kaminski, Kaufmann, and Schaub}{Gebser
  et~al\mbox{.}}{2014}]{gekakasc14b}
{\sc Gebser, M.}, {\sc Kaminski, R.}, {\sc Kaufmann, B.}, {\sc and} {\sc
  Schaub, T.} 2014.
\newblock \textit{Clingo} = {ASP} + control: Preliminary report.
\newblock In {\em Technical Communications of the Thirtieth International
  Conference on Logic Programming (ICLP'14)}, {M.~Leuschel} {and}
  {T.~Schrijvers}, Eds. Theory and Practice of Logic Programming, Online
  Supplement, vol. arXiv:1405.3694v1.
\newblock Available at \url{http://arxiv.org/abs/1405.3694v1}.

\bibitem[\protect\citeauthoryear{Gebser, Kaminski, Obermeier, and
  Schaub}{Gebser et~al\mbox{.}}{2015}]{gekaobsc15a}
{\sc Gebser, M.}, {\sc Kaminski, R.}, {\sc Obermeier, P.}, {\sc and} {\sc
  Schaub, T.} 2015.
\newblock Ricochet robots reloaded: A case-study in multi-shot {ASP} solving.
\newblock In {\em Advances in Knowledge Representation, Logic Programming, and
  Abstract Argumentation: Essays Dedicated to {G}erhard {B}rewka on the
  Occasion of His 60th Birthday}, {T.~Eiter}, {H.~Strass},
  {M.~Truszczy{\'n}ski}, {and} {S.~Woltran}, Eds. Lecture Notes in Artificial
  Intelligence, vol. 9060. Springer-Verlag, 17--32.

\bibitem[\protect\citeauthoryear{Gebser, Kaufmann, Neumann, and Schaub}{Gebser
  et~al\mbox{.}}{2007}]{gekanesc07a}
{\sc Gebser, M.}, {\sc Kaufmann, B.}, {\sc Neumann, A.}, {\sc and} {\sc Schaub,
  T.} 2007.
\newblock Conflict-driven answer set solving.
\newblock In {\em Proceedings of the Twentieth International Joint Conference
  on Artificial Intelligence (IJCAI'07)}, {M.~Veloso}, Ed. AAAI/MIT Press,
  386--392.

\bibitem[\protect\citeauthoryear{Gebser, Kaufmann, Otero, Romero, Schaub, and
  Wanko}{Gebser et~al\mbox{.}}{2013}]{gekaotroscwa13a}
{\sc Gebser, M.}, {\sc Kaufmann, B.}, {\sc Otero, R.}, {\sc Romero, J.}, {\sc
  Schaub, T.}, {\sc and} {\sc Wanko, P.} 2013.
\newblock Domain-specific heuristics in answer set programming.
\newblock In {\em Proceedings of the Twenty-Seventh National Conference on
  Artificial Intelligence (AAAI'13)}, {M.~{desJardins}} {and} {M.~Littman},
  Eds. AAAI Press, 350--356.

\bibitem[\protect\citeauthoryear{Gebser, Kaufmann, and Schaub}{Gebser
  et~al\mbox{.}}{2012}]{gekasc12b}
{\sc Gebser, M.}, {\sc Kaufmann, B.}, {\sc and} {\sc Schaub, T.} 2012.
\newblock Multi-threaded {ASP} solving with clasp.
\newblock {\em Theory and Practice of Logic Programming\/}~{\em 12,\/}~4-5,
  525--545.

\bibitem[\protect\citeauthoryear{Gebser, Ostrowski, and Schaub}{Gebser
  et~al\mbox{.}}{2009}]{geossc09a}
{\sc Gebser, M.}, {\sc Ostrowski, M.}, {\sc and} {\sc Schaub, T.} 2009.
\newblock Constraint answer set solving.
\newblock In {\em Proceedings of the Twenty-fifth International Conference on
  Logic Programming (ICLP'09)}, {P.~Hill} {and} {D.~Warren}, Eds. Lecture Notes
  in Computer Science, vol. 5649. Springer-Verlag, 235--249.

\bibitem[\protect\citeauthoryear{{Gecode Team}}{{Gecode Team}}{2006}]{gecode}
{\sc {Gecode Team}}. 2006.
\newblock Gecode: Generic constraint development environment.
\newblock Available from \url{http://www.gecode.org}.

\bibitem[\protect\citeauthoryear{Gelfond and Lifschitz}{Gelfond and
  Lifschitz}{1988}]{gellif88b}
{\sc Gelfond, M.} {\sc and} {\sc Lifschitz, V.} 1988.
\newblock The stable model semantics for logic programming.
\newblock In {\em Proceedings of the Fifth International Conference and
  Symposium of Logic Programming (ICLP'88)}, {R.~Kowalski} {and} {K.~Bowen},
  Eds. MIT Press, 1070--1080.

\bibitem[\protect\citeauthoryear{Janhunen, Liu, and Niemelä}{Janhunen
  et~al\mbox{.}}{2011}]{jalini11a}
{\sc Janhunen, T.}, {\sc Liu, G.}, {\sc and} {\sc Niemelä, I.} 2011.
\newblock Tight integration of non-ground answer set programming and
  satisfiability modulo theories.
\newblock In {\em Proceedings of the First Workshop on Grounding and
  Transformation for Theories with Variables (GTTV'11)}, {P.~Cabalar},
  {D.~Mitchell}, {D.~Pearce}, {and} {E.~Ternovska}, Eds. 1--13.

\bibitem[\protect\citeauthoryear{Lierler and Susman}{Lierler and
  Susman}{2016}]{liesus16a}
{\sc Lierler, Y.} {\sc and} {\sc Susman, B.} 2016.
\newblock {SMT}-based constraint answer set solver {EZSMT} (system
  description).
\newblock See \citeN{iclp-lipics16}, 1:1--1:15.

\bibitem[\protect\citeauthoryear{Lifschitz}{Lifschitz}{2008}]{lifschitz08b}
{\sc Lifschitz, V.} 2008.
\newblock What is answer set programming?
\newblock In {\em Proceedings of the Twenty-third National Conference on
  Artificial Intelligence (AAAI'08)}, {D.~Fox} {and} {C.~Gomes}, Eds. AAAI
  Press, 1594--1597.

\bibitem[\protect\citeauthoryear{Marques-Silva and Sakallah}{Marques-Silva and
  Sakallah}{1999}]{marsak99a}
{\sc Marques-Silva, J.} {\sc and} {\sc Sakallah, K.} 1999.
\newblock {GRASP}: A search algorithm for propositional satisfiability.
\newblock {\em IEEE Transactions on Computers\/}~{\em 48,\/}~5, 506--521.

\bibitem[\protect\citeauthoryear{Mellarkod and Gelfond}{Mellarkod and
  Gelfond}{2008}]{melgel08a}
{\sc Mellarkod, V.} {\sc and} {\sc Gelfond, M.} 2008.
\newblock Integrating answer set reasoning with constraint solving techniques.
\newblock In {\em Proceedings of the Ninth International Symposium on
  Functional and Logic Programming (FLOPS'08)}, {J.~Garrigue} {and}
  {M.~Hermenegildo}, Eds. Lecture Notes in Computer Science, vol. 4989.
  Springer-Verlag, 15--31.

\bibitem[\protect\citeauthoryear{Mellarkod, Gelfond, and Zhang}{Mellarkod
  et~al\mbox{.}}{2008}]{megezh08a}
{\sc Mellarkod, V.}, {\sc Gelfond, M.}, {\sc and} {\sc Zhang, Y.} 2008.
\newblock Integrating answer set programming and constraint logic programming.
\newblock {\em Annals of Mathematics and Artificial Intelligence\/}~{\em
  53,\/}~1-4, 251--287.

\bibitem[\protect\citeauthoryear{Metodi, Codish, and Stuckey}{Metodi
  et~al\mbox{.}}{2013}]{mecost13a}
{\sc Metodi, A.}, {\sc Codish, M.}, {\sc and} {\sc Stuckey, P.} 2013.
\newblock Boolean equi-propagation for concise and efficient {SAT} encodings of
  combinatorial problems.
\newblock {\em Journal of Artificial Intelligence Research\/}~{\em 46},
  303--341.

\bibitem[\protect\citeauthoryear{Ohrimenko, Stuckey, and Codish}{Ohrimenko
  et~al\mbox{.}}{2009}]{ohstco09a}
{\sc Ohrimenko, O.}, {\sc Stuckey, P.}, {\sc and} {\sc Codish, M.} 2009.
\newblock Propagation via lazy clause generation.
\newblock {\em Constraints\/}~{\em 14,\/}~3, 357--391.

\bibitem[\protect\citeauthoryear{Ostrowski}{Ostrowski}{2017}]{ostrowski17a}
{\sc Ostrowski, M.} 2017.
\newblock Modern constraint answer set solving.
\newblock Ph.D. thesis, University of Potsdam.

\bibitem[\protect\citeauthoryear{Ostrowski and Schaub}{Ostrowski and
  Schaub}{2012}]{ostsch12a}
{\sc Ostrowski, M.} {\sc and} {\sc Schaub, T.} 2012.
\newblock {ASP} modulo {CSP}: The clingcon system.
\newblock {\em Theory and Practice of Logic Programming\/}~{\em 12,\/}~4-5,
  485--503.

\bibitem[\protect\citeauthoryear{Rossi, {van Beek}, and Walsh}{Rossi
  et~al\mbox{.}}{2006}]{CPHandbook}
{\sc Rossi, F.}, {\sc {van Beek}, P.}, {\sc and} {\sc Walsh, T.}, Eds. 2006.
\newblock {\em Handbook of Constraint Programming}.
\newblock Elsevier Science.

\bibitem[\protect\citeauthoryear{Roussel and Lecoutre}{Roussel and
  Lecoutre}{2009}]{roulec09a}
{\sc Roussel, O.} {\sc and} {\sc Lecoutre, C.} 2009.
\newblock {XML} representation of constraint networks: Format {XCSP} 2.1.
\newblock {\em CoRR\/}~{\em abs/0902.2362}.

\bibitem[\protect\citeauthoryear{Schulte and Tack}{Schulte and
  Tack}{2005}]{schtac06a}
{\sc Schulte, C.} {\sc and} {\sc Tack, G.} 2005.
\newblock Views and iterators for generic constraint implementations.
\newblock In {\em Proceedings of the Eleventh International Conference on
  Principles and Practice of Constraint Programming (CP'05)}, {P.~{van Beek}},
  Ed. Lecture Notes in Computer Science, vol. 3709. Springer-Verlag, 118--132.

\bibitem[\protect\citeauthoryear{Simons, Niemelä, and Soininen}{Simons
  et~al\mbox{.}}{2002}]{siniso02a}
{\sc Simons, P.}, {\sc Niemelä, I.}, {\sc and} {\sc Soininen, T.} 2002.
\newblock Extending and implementing the stable model semantics.
\newblock {\em Artificial Intelligence\/}~{\em 138,\/}~1-2, 181--234.

\bibitem[\protect\citeauthoryear{Soh, Inoue, Tamura, Banbara, and
  Nabeshima}{Soh et~al\mbox{.}}{2010}]{sointabana10a}
{\sc Soh, T.}, {\sc Inoue, K.}, {\sc Tamura, N.}, {\sc Banbara, M.}, {\sc and}
  {\sc Nabeshima, H.} 2010.
\newblock A {SAT}-based method for solving the two-dimensional strip packing
  problem.
\newblock {\em Fundamenta Informaticae\/}~{\em 102,\/}~3-4, 467--487.

\bibitem[\protect\citeauthoryear{Tamura, Banbara, and Soh}{Tamura
  et~al\mbox{.}}{2013}]{basota13a}
{\sc Tamura, N.}, {\sc Banbara, M.}, {\sc and} {\sc Soh, T.} 2013.
\newblock Compiling pseudo-boolean constraints to {SAT} with order encoding.
\newblock See \citeN{ictai13}, 1020--1027.

\bibitem[\protect\citeauthoryear{Tamura, Taga, Kitagawa, and Banbara}{Tamura
  et~al\mbox{.}}{2009}]{tatakiba09a}
{\sc Tamura, N.}, {\sc Taga, A.}, {\sc Kitagawa, S.}, {\sc and} {\sc Banbara,
  M.} 2009.
\newblock Compiling finite linear {CSP} into {SAT}.
\newblock {\em Constraints\/}~{\em 14,\/}~2, 254--272.

\bibitem[\protect\citeauthoryear{Thibaut and Stuckey}{Thibaut and
  Stuckey}{2009}]{thistu09a}
{\sc Thibaut, F.} {\sc and} {\sc Stuckey, P.} 2009.
\newblock Lazy clause generation reengineered.
\newblock In {\em Proceedings of the Fifteenth International Conference on
  Principles and Practice of Constraint Programming (CP'09)}, {I.~Gent}, Ed.
  Lecture Notes in Computer Science, vol. 5732. Springer-Verlag, 352--366.

\bibitem[\protect\citeauthoryear{Thiffault, Bacchus, and Walsh}{Thiffault
  et~al\mbox{.}}{2004}]{thbawa04a}
{\sc Thiffault, C.}, {\sc Bacchus, F.}, {\sc and} {\sc Walsh, T.} 2004.
\newblock Solving non-clausal formulas with {DPLL} search.
\newblock In {\em Proceedings of the Tenth International Conference on
  Principles and Practice of Constraint Programming (CP'04)}, {M.~Wallace}, Ed.
  Lecture Notes in Computer Science, vol. 3258. Springer-Verlag, 663--678.

\bibitem[\protect\citeauthoryear{Walsh}{Walsh}{2000}]{walsh00a}
{\sc Walsh, T.} 2000.
\newblock {SAT} versus {CSP}.
\newblock In {\em Proceedings of the Sixth International Conference on
  Principles and Practice of Constraint Programming (CP'00)}, {R.~Dechter}, Ed.
  Lecture Notes in Computer Science, vol. 1894. Springer-Verlag, 441--456.

\bibitem[\protect\citeauthoryear{Zhang, Madigan, Moskewicz, and Malik}{Zhang
  et~al\mbox{.}}{2001}]{zamamoma01a}
{\sc Zhang, L.}, {\sc Madigan, C.}, {\sc Moskewicz, M.}, {\sc and} {\sc Malik,
  S.} 2001.
\newblock Efficient conflict driven learning in a {B}oolean satisfiability
  solver.
\newblock In {\em Proceedings of the International Conference on Computer-Aided
  Design (ICCAD'01)}. ACM Press, 279--285.

\end{thebibliography}
